\documentclass[10pt,twocolumn,letterpaper]{article}

\usepackage[pagenumbers]{cvpr} 

\usepackage[utf8]{inputenc} 
\usepackage[T1]{fontenc}    
\usepackage{url}            
\usepackage{booktabs}       
\usepackage{amsfonts}       
\usepackage{nicefrac}       
\usepackage{microtype}      
\usepackage{xcolor}         

\usepackage{graphicx}
\usepackage{graphbox}
\usepackage{floatrow}
\usepackage{amsmath}
\usepackage{amssymb}
\usepackage{paralist}

\usepackage{float}
\usepackage{longtable}

\usepackage{xcolor}
\usepackage{multirow}
\usepackage{array}
\usepackage{tabularx}
\usepackage{color}
\usepackage{colortbl}
\usepackage{paralist}
\usepackage{dblfloatfix}
\newcolumntype{K}[1]{>{\centering\arraybackslash}p{#1}}
\newcommand{\band}{\rowcolor{gray!20}}

\newcolumntype{?}{!{\vrule width 1.5pt}}

\newcommand*{\mySpecialfootnotes}[1]{%
  \patchcmd{\@footnotetext}{\floatingpenalty\@MM}{\floatingpenalty#1\relax}%
           {}{\errmessage{Couldn't patch \string\@footnotetext}}%
}

\usepackage[pagebackref,breaklinks,colorlinks]{hyperref}

\usepackage[capitalize]{cleveref}
\crefname{section}{Sec.}{Secs.}
\Crefname{section}{Section}{Sections}
\Crefname{table}{Table}{Tables}
\crefname{table}{Tab.}{Tabs.}

\begin{document}

\title{On the Transferability of Visual Features in Generalized Zero-Shot Learning}

\author{Paola Cascante-Bonilla\textsuperscript{$\dagger$} \quad Leonid Karlinsky\textsuperscript{$\ddagger$} \quad James Seale Smith\textsuperscript{$\mathsection$} \quad Yanjun Qi\textsuperscript{$\natural$} \quad Vicente Ordonez\textsuperscript{$\dagger$}\\
\textsuperscript{$\dagger$}Rice University \quad \textsuperscript{$\ddagger$}MIT-IBM Watson AI Lab \quad \textsuperscript{$\mathsection$}Georgia Institute of Technology \quad \textsuperscript{$\natural$}University of Virginia\\
{\tt\small \{pc51, vicenteor\}@rice.com, leonidka@ibm.com, jamessealesmith@gatech.edu, yq2h@virginia.edu}
}
\maketitle

\begin{abstract}

Generalized Zero-Shot Learning (GZSL) aims to train a classifier that can generalize to unseen classes, using a set of attributes as auxiliary information, and the visual features extracted from a pre-trained convolutional neural network.
While recent GZSL methods have explored various techniques to leverage the capacity of these features, there has been an extensive growth of representation learning techniques that remain under-explored. 
In this work, we investigate the utility of different GZSL methods when using different feature extractors, and examine how these models' pre-training objectives, datasets, and architecture design affect their feature representation ability.
Our results indicate that 1) methods using generative components for GZSL provide more advantages when using recent feature extractors;
2) feature extractors pre-trained using self-supervised learning objectives combined with cross-entropy and knowledge distillation provide better feature 
representations, increasing up to 15\% performance when used with recent GZSL techniques; 3) specific feature extractors pre-trained with larger datasets do not necessarily boost the performance of GZSL methods. 
In addition, we investigate how GZSL methods fare against CLIP, a more recent multi-modal pre-trained model with strong zero-shot performance. We found that GZSL tasks still benefit from generative-based GZSL methods along with CLIP's internet-scale pre-training to achieve state-of-the-art performance in fine-grained datasets.
We release a modular framework for analyzing representation learning issues in GZSL here: \href{https://github.com/uvavision/TV-GZSL}{https://github.com/uvavision/TV-GZSL}.

\end{abstract}

\begin{figure}[t!]
    \centering
    \includegraphics[width=\linewidth]{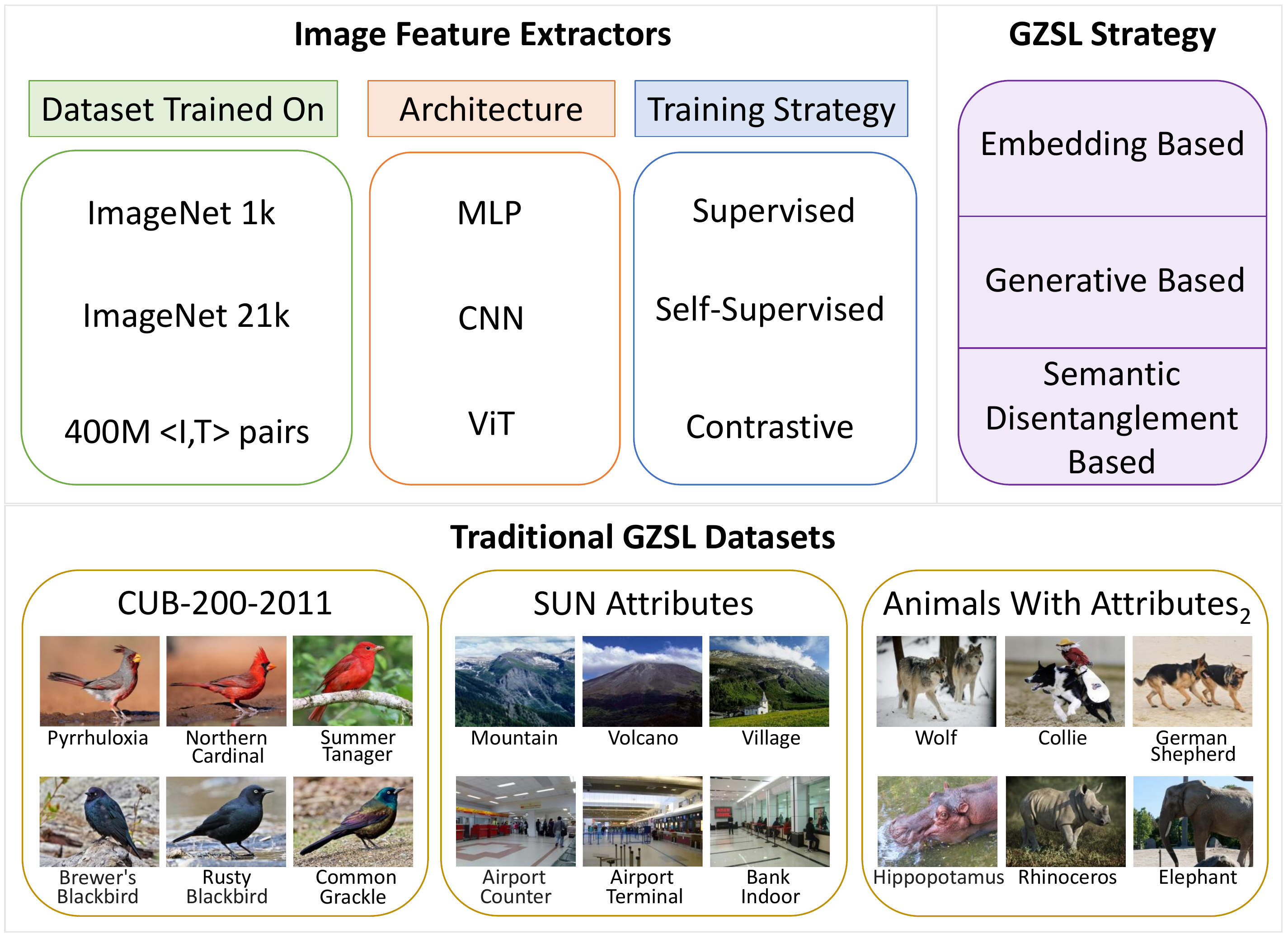}
    \caption{\textbf{Overview of our large-scale analysis.} We explore a diverse set of visual feature backbones that were trained using different objectives. We further use them to extract visual features and train a diverse set of classic and modern Generalized Zero-shot Learning (GZSL) methods. We use three standard datasets to measure the methods' performance: CUB-200-2011 and SUN Attributes, which are fine-grained, and Animales with Attributes 2 (AWA2) which is a coarse-grained dataset. Best viewed in color.}
    \label{fig:figure1}
    \vspace{-0.1in}
\end{figure}

\section{Introduction}

Deep learning models have achieved remarkable accuracy in many computer vision classification tasks when labeled data is available, and the data distribution is consistent during training and test time~\cite{effectivenessData, revisitEffectivenessData, Ren2015FasterRT}. It is now possible to train image classifiers that can distinguish with high accuracy thousands of image categories~\cite{Russakovsky2015ImageNetLS}. 
However, in order to enable a model to recognize novel categories, it is still necessary to collect a dataset with representative human-labeled examples.
For this reason, literature has proposed zero-shot learning to help a model recognize novel {\em unseen} categories without needing any images of the new category. Zero-shot learning relies on auxiliary information such as textual descriptions or category attributes~\cite{5206772,elhoseiny2013write,lampert2013attribute}. 
The general idea is to leverage the auxiliary information to transfer visual knowledge from images in {\em seen} categories to a set of images from {\em unseen} categories. 
Once this function is learned, it can be used to categorize {\em unseen} samples into novel classes.

Early work on zero-shot learning focused on obtaining a good accuracy just on a set of {\em unseen} categories at inference time. In a more challenging scenario known as Generalized Zero-shot Learning (GZSL), both {\em seen} and {\em unseen} categories are considered at test time~\cite{Chao2016AnES, Pourpanah2020ARO}. This setup is more realistic and has been adopted in all the recent works in zero-shot learning; therefore, we focus exclusively on this setting.

First-generation methods are coined as embedding-based techniques, which focus on learning a function to align the seen images and unseen attributes and further measure the similarity between the mapped and predicted representations of the data samples in the embedding space~\cite{ESZSL, ALE, Zhang2017LearningAD, Zhang2020TowardsED}.
Further GZSL methods have shown significant improvement over this early work by teaching a model to generate the visual features of the {\em unseen} classes based on the visual features of the {\em seen} classes and the semantic representations of both {\em seen} and {\em unseen} categories. 
More recent works have instead explored rich image feature representations and their ability to provide enough information for a mapping function to generalize to new classes~\cite{Tong2019HierarchicalDO, SDGZSL}. 
These methods propose to learn discriminative representations from image features through disentanglement over feature groups by factorizing the useful dimensions to avoid bias towards the seen categories when trying to learn an attribute-visual alignment.

While disentanglement-based methods have significantly improved over prior work,
the representation capabilities of image features have mostly been tested under a ResNet101 model~\cite{RNs} pre-trained on Imagenet~\cite{AWA2, Liu2018GeneralizedZL}. 
In this work, we propose to explore different architectures trained with different objectives to be used as feature extractors. Given the wide availability of pre-trained model parameters~\cite{Gavrikov2022CNNFD, timm, Croce2021RobustBenchAS}, we perform a large-scale analysis to assess the impact of modern visual features backbones, our results are summarized in Figure~\ref{fig:figure1}. 
In our analysis, we consider visual features extracted from both uni-modal and multi-modal architectures pre-trained on standard settings, i.e., models pre-trained with ImageNet-1K; to models pretrained with even larger amounts of data, i.e., models pre-trained with ImageNet-21K and models pre-trained with 400 million Image/Text pairs.

Our results show that ResNet101 and similar feature extractors may not provide enough information, given the nature of the backbone architecture limitations and the training objective. 
Moreover, when examining feature extractors pre-trained with larger datasets, one would assume that the features extracted would contain more similarities to the data present in the GSZL test splits, causing all methods to perform better due to data leakage. However, we found that using features extracted from large-scale pre-trained {\em uni-modal networks} does not significantly impact the zero-shot performance.
Our experiments also reveal that with feature representations extracted from Visual Transformer~\cite{ViT} based architectures, no semantic disentanglement module is necessary to achieve state-of-the-art performance, and generative-based methods are superior in all benchmarks by a large margin ($\approx23\%$) when using features extracted from Transformer based architectures~\cite{ViT} trained with a contrastive objective.
Furthermore, given the capabilities of new models~\cite{CLIP, ALIGN, Singh2022FLAVAAF} to generalize to new tasks, we investigate how CLIP~\cite{CLIP} performs against GZSL methods and reveal that generative-based GZSL techniques are still necessary to achieve state-of-the-art results on fine-grained datasets.

Our primary contributions can be summarized as follows:

\begin{compactitem}
\item A large-scale study of different methods and features extracted from a diverse set of architectures and training approaches, applied to state-of-the-art methods from different GZSL families (i.e., embedding based~\cite{ALE, ESZSL, DeViSE}, generative based~\cite{tfvaegan, CADA_VAE, CE}, semantic disentanglement based~\cite{Tong2019HierarchicalDO, Chen2021FREE, SDGZSL}).

\item A library containing the revisited GSZL methods we chose to explore, allowing a unified codebase for reproducibility and further analysis based on the findings shown in this work. Texts generated to finetune the models based on the attribute vectors provided in each dataset, along with the weights and the extracted features we use in our experiments.

\item Updates and key insights which we hope will reshape the Generalized Zero-Shot Learning research track in favor of leveraging richer feature representations. 

\end{compactitem}

We expect our work to motivate further feature representation explorations to the GZSL task in a more realistic, practical, and challenging scenario, given the recent advances with large pre-trained models. All the resources used, including GPU information and computing infrastructure are detailed in the Appendix, along with hyperparameter selections and data splits we use for all our experiments.

\section{Related Work}

The goal of GZSL is to classify images from both seen and unseen categories by transferring knowledge from seen to unseen classes using a set of attributes as auxiliary information.
Since every attribute vector entry represents a class description, the assumption is that the classes with similar descriptions contain a similar attribute vector in the semantic space. Therefore, the general idea is to learn a function that allows this mapping between modalities. 
The general idea is to align visual features from seen classes with their corresponding attribute vectors. 
Current methods can be broadly categorized into:

\vspace{0.05in}
\textbf{Embedding-based methods}, which aim to learn a mapping function or a projection between visual features and their attributes or descriptions~\cite{DeViSE, ESZSL, ALE}. This mapping function is used to project image features into a semantic space so that it is possible to classify seen and unseen classes by estimating how close these features are to a class embedding vector~\cite{Rahman2018AUA, Chen2018ZeroShotVR, Shen2020InvertibleZR, 9724125}.

\vspace{0.05in}
\textbf{Generative-based methods}, which synthesize an unlimited number of visual features using the auxiliary information from the unseen classes, and compensate for the imbalance classification problem that poses the GZSL task~\cite{CADA_VAE, tfvaegan, CE}. A generative model is a probabilistic model that is representative of the conditional probability of the observable input $X$, given a target $y$~\cite{VAEs, GANs}. Recent advances in generative modeling have gained a significant amount of attention. In the GZSL setting, generative models are leveraged to learn to generate visual features or images for the unseen classes\cite{Su_2022_CVPR, Kong_2022_CVPR}. This is achieved using samples from the seen classes and semantic representations of both seen and unseen classes. Generative-based methods convert the GZSL problem into a supervised learning problem by generating samples for both seen and unseen classes.

\vspace{0.05in}
\textbf{Semantic disentanglement-based methods}, which aim to factorize the useful dimensions of a given visual feature to learn the attribute-visual alignment. The assumption is that the image features extracted from a ResNet101 that was pre-trained on Imagenet, broadly used in all traditional GSZL datasets~\cite{CUB, SUN, AWA2}, are not ideal for the zero-shot learning task~\cite{Tong2019HierarchicalDO, Chen2021FREE, SDGZSL}. Since these features are not representative with respect to the specific attributes that describe the image parts/composition, not all the dimensions of these features are semantically related to the given attributes; thus, it is necessary to factorize or disentangle the useful dimensions to avoid bias when trying to learn the attribute-visual alignment.
In this context, entanglement refers to the property of not having independence among attributes of one representation. In an entangled representation, all the factors of variation are mixed, and there is no explicit separation that represents the important characteristics in the images~\cite{Bengio2013RepresentationLA}. On the other hand, given an image dataset of birds such as the CUB dataset~\cite{CUB}, a disentangled representation may consist of separate dimensions for wing color, breast color, bill shape, tail pattern, crown color, wing pattern, etc~\cite{Eastwood2018AFF}.

\vspace{-0.05in}
\section{Generalized Zero-Shot Learning}

In the GZSL setting, we define $S$ as the set of seen classes within $N_s$ categories and $Y_s$ as their corresponding labels. We also define $U$ as the set of unseen classes with $N_u$ categories and $Y_u$ as their corresponding labels. Here, $S$ and $U$ have no intersection. 
Each set consists of image-features $x$, class labels $y$ available during training and class-embeddings $c(y)$. 
Thus, $S = \{ {x_{i}^{s}, y_{i}^{s}) \} }^{N_{s}}_{i=1}$ and $U = \{ {x_{i}^{u}, y_{i}^{u}) \} }^{N_{u}}_{i=1}$. 
Additionally, we have access to a set of semantic descriptions of the seen and unseen classes $A = \{ {a_{i}^{s} \} }^{N_{s} + N_{u} }_{i=1}$, which are typically class-embeddings vectors of hand-annotated  attributes. The image features are typically extracted from a feature backbone (i.e. ResNet101 pretrained on ImageNet-1K).
When training, we use $A$ along with the visual features and labels from the seen set (i.e., $ \{ {x_{i}^{s}, y_{i}^{s} \} }^{N_{s}}_{i=1}$) and only the labels from the unseen set (i.e., $ \{ y_{i}^{u} \} ^{N_{u}}_{i=1}$). Finally, the test set contains image-features from both $S$ and $U$, and their corresponding labels.
This section presents and describes all the methods we include in this study grouped by their corresponding family of methods. These can be characterized as: embedding-based, generative-based and disentanglement-based. We also present all the datasets we use in our study. Additional training and computational details are included in the Appendix.

\subsection{Embedding-based Methods}

\textit{\textbf{DeViSE}: A Deep Visual-Semantic Embedding Model}~\cite{DeViSE} (2013) propose to learn a linear mapping between the image and the semantic space and introduce the use of a ranking loss instead of an L$_2$ loss. This avoids making the vectors become closer to one another without taking into account the incorrect labels that are closer to the target image.

\textit{An embarrassingly simple approach to zero-shot learning (\textbf{ESZSL})}~\cite{ESZSL} (2015) use a linear model to create relationships between features, attributes and classes. This linear model has two layers, the first layer  helps in defining the relationship between features and attributes, and the second layer deals with modeling the relationship between attributes and classes where the prescribed attribute signatures are fixed. This method uses a square loss to learn the bilinear compatibility between attributes and their corresponding classes. 

\textit{Label-Embedding for Image Classification} present an \textit{Attribute Label Embedding approach (\textbf{ALE})}~\cite{ALE} (2016), they propose to embed each class in the space of attribute vectors, and further propose to learn a bilinear compatibility function between the image and a label embedding using a ranking objective function.

\subsection{Generative-based Methods}

\textit{Generalized Zero- and Few-Shot Learning via Aligned Variational Autoencoders (\textbf{CADA-VAE})}~\cite{CADA_VAE} (2019) use two variational autoencoders (VAEs~\cite{VAEs}) to align the visual and semantic features by learning a shared latent space between both modalities. Then, these VAEs are used to synthesize a large number of seen and unseen features, which are then used to train a classifier.

\textit{Latent Embedding Feedback and Discriminative
Features for Zero-Shot Classification (\textbf{tfVAEGAN})}~\cite{tfvaegan} (2020) propose to use a feedback module in a model that combines a variational autoencoder (VAE~\cite{VAEs}) and a generative adversarial network (GAN~\cite{GANs}) to modulate the latent representation of the generator. They propose to enforce semantic consistency by introducing a feedback loop from the semantic embedding decoder. They propose to use both synthesized features and latent embeddings during classification.


More recently, hybrid methods such as \textit{Contrastive Embedding for Generalized Zero-Shot Learning (\textbf{CE})}~\cite{CE} (2021) have emerged. This method proposes to integrate the generation model with the embedding model to map 
the real and the synthetic samples produced by the generation model into an embedding space. To do this, they leverage a contrastive loss that learns to discriminate between one
positive sample and a large number of negative samples from different semantic descriptor classes. They claim that the original visual feature space is suboptimal for GZSL classification since it lacks discriminative information, which they aim to learn using the contrastive objective.

\subsection{Semantic Disentanglement-based Methods}

\textit{\textbf{FREE}: Feature Refinement for Generalized Zero-Shot Learning}~\cite{Chen2021FREE} (2021) use a feature refinement module to jointly map the semantic and visual modalities, and refine the visual features of seen and unseen class samples -- dropping unnecessary information from the visual features. They use the class label supervision and a semantic cycle-consistency
constraint to guide the proposed module to learn relevant feature representations that are also semantically relevant with respect to their corresponding classes.

\textit{Semantics Disentangling for Generalized Zero-Shot Learning (\textbf{SDGZSL})}~\cite{SDGZSL} (2021) introduce a total correlation penalty that is applied to the visual features of unseen classes that were generated by a conditional VAE~\cite{VAEs}. This work uses a Relation Network~\cite{RelNetwork} to correlate the factorized estimated features -- the \textit{semantic-consistent} and the \textit{semantic-unrelated} latent features.

\begin{figure*}[tp]
\begin{minipage}[t]{0.32\textwidth}
  \includegraphics[align=t,width=\linewidth]{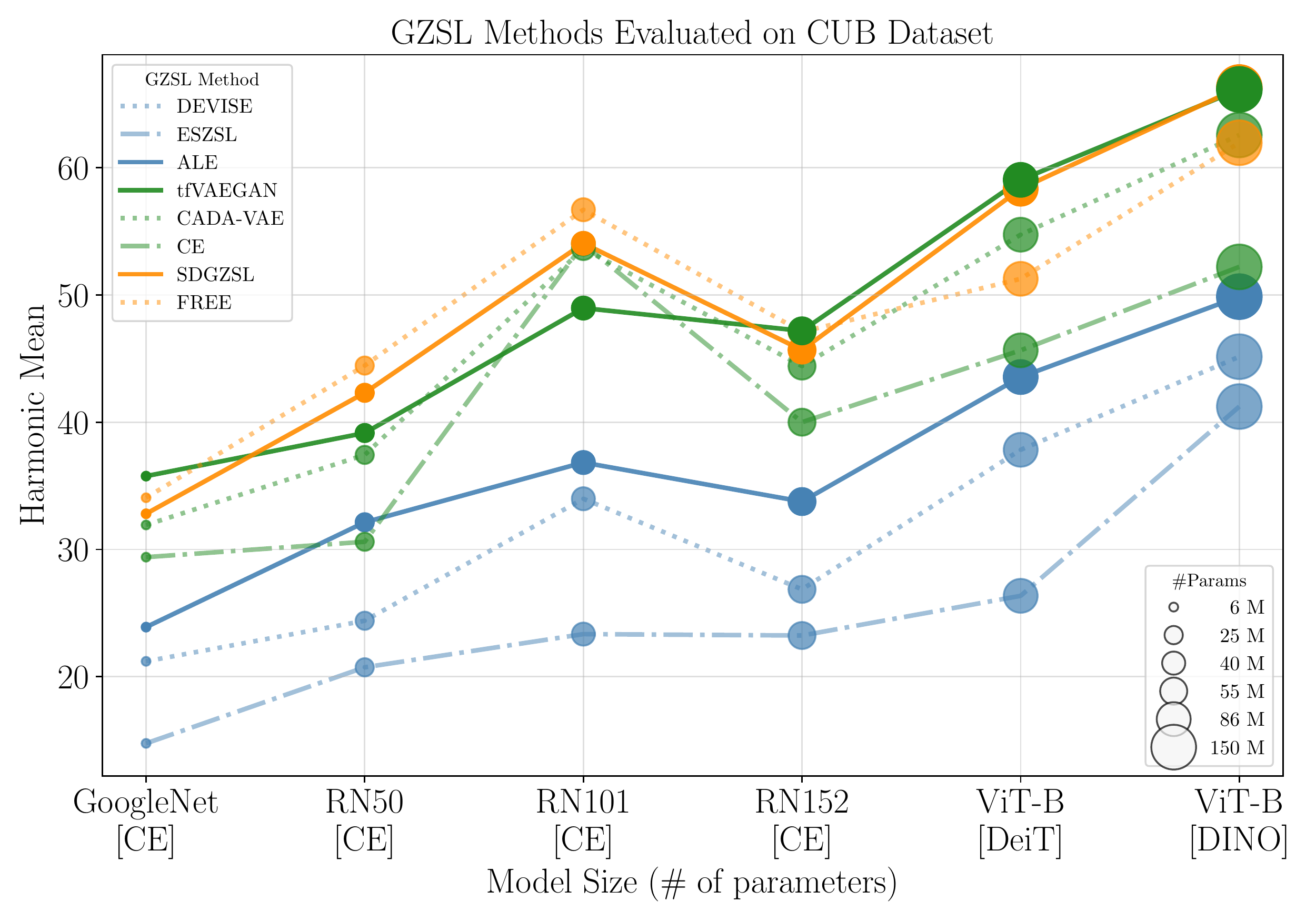}
\end{minipage}%
\hfill 
\begin{minipage}[t]{0.32\textwidth}
  \includegraphics[align=t,width=\linewidth]{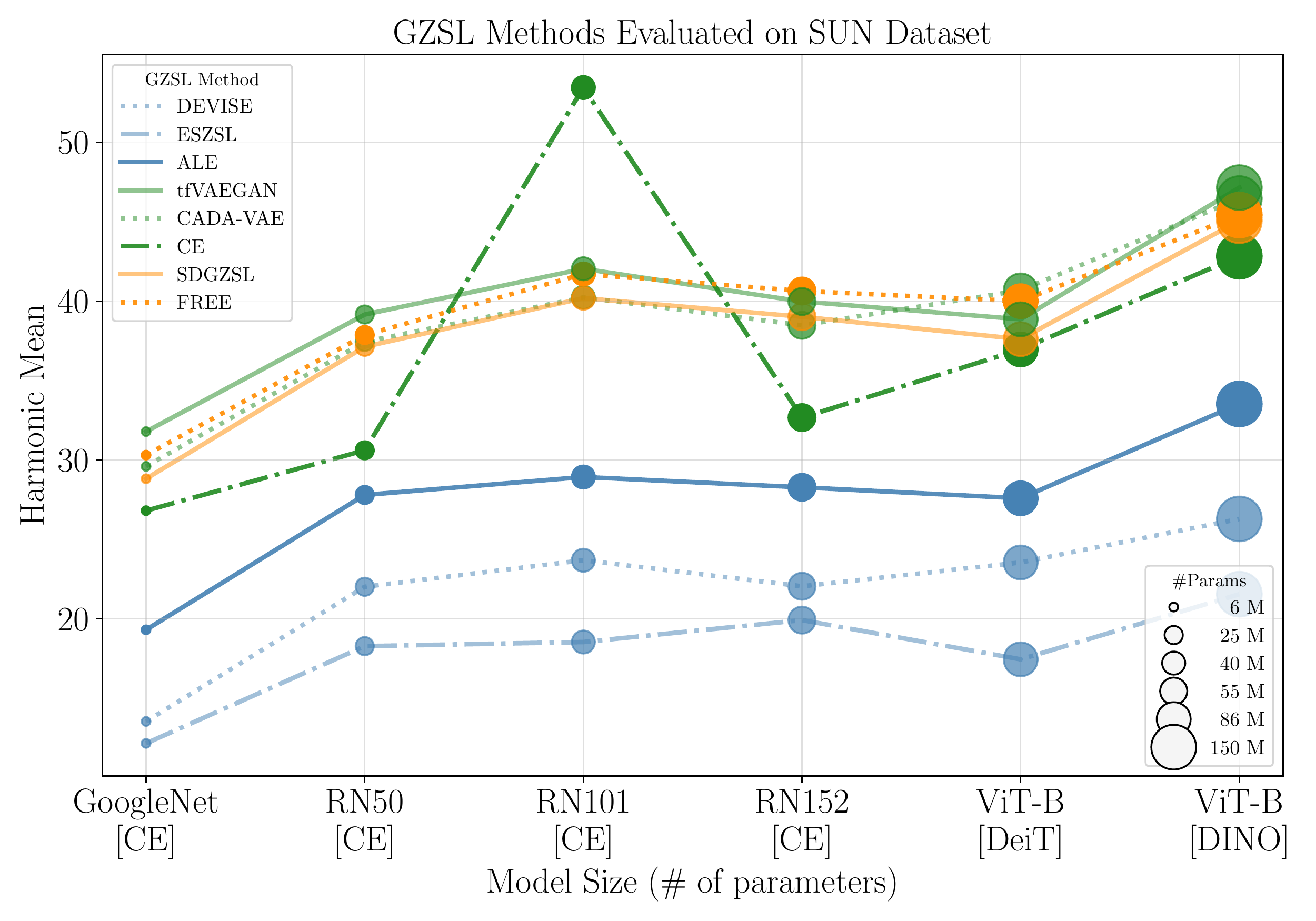}
\end{minipage}%
\hfill
\begin{minipage}[t]{0.35\textwidth}
  \includegraphics[align=t,width=\linewidth]{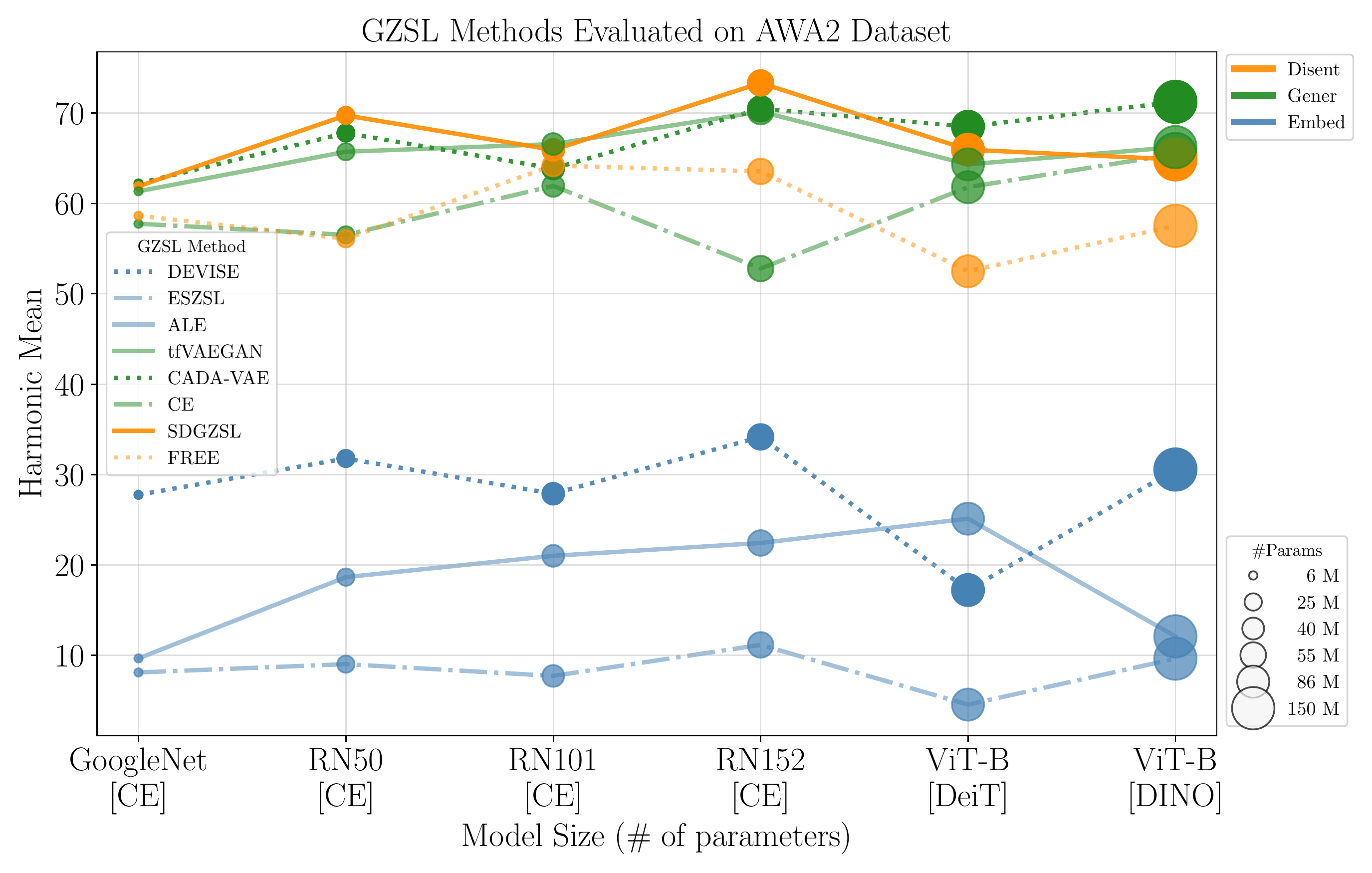}
\end{minipage}%
 \caption{\textbf{Impact of Model Parameter Size.} We show the Harmonic Mean performance of different methods when using a specific model to extract the features of image samples from CUB, SUN, and AWA2 datasets. All models were pre-trained using ImageNet-1k. The blue lines correspond to the embedding-based methods, the green lines correspond to the generative-based methods, and the orange lines correspond to the semantic disentanglement-based methods. Best viewed in color.}
 \label{fig:model_backbone_size_vs_method}
\end{figure*}

\begin{figure*}[bp]
\begin{minipage}[t]{0.33\textwidth}
  \includegraphics[align=t,width=\linewidth]{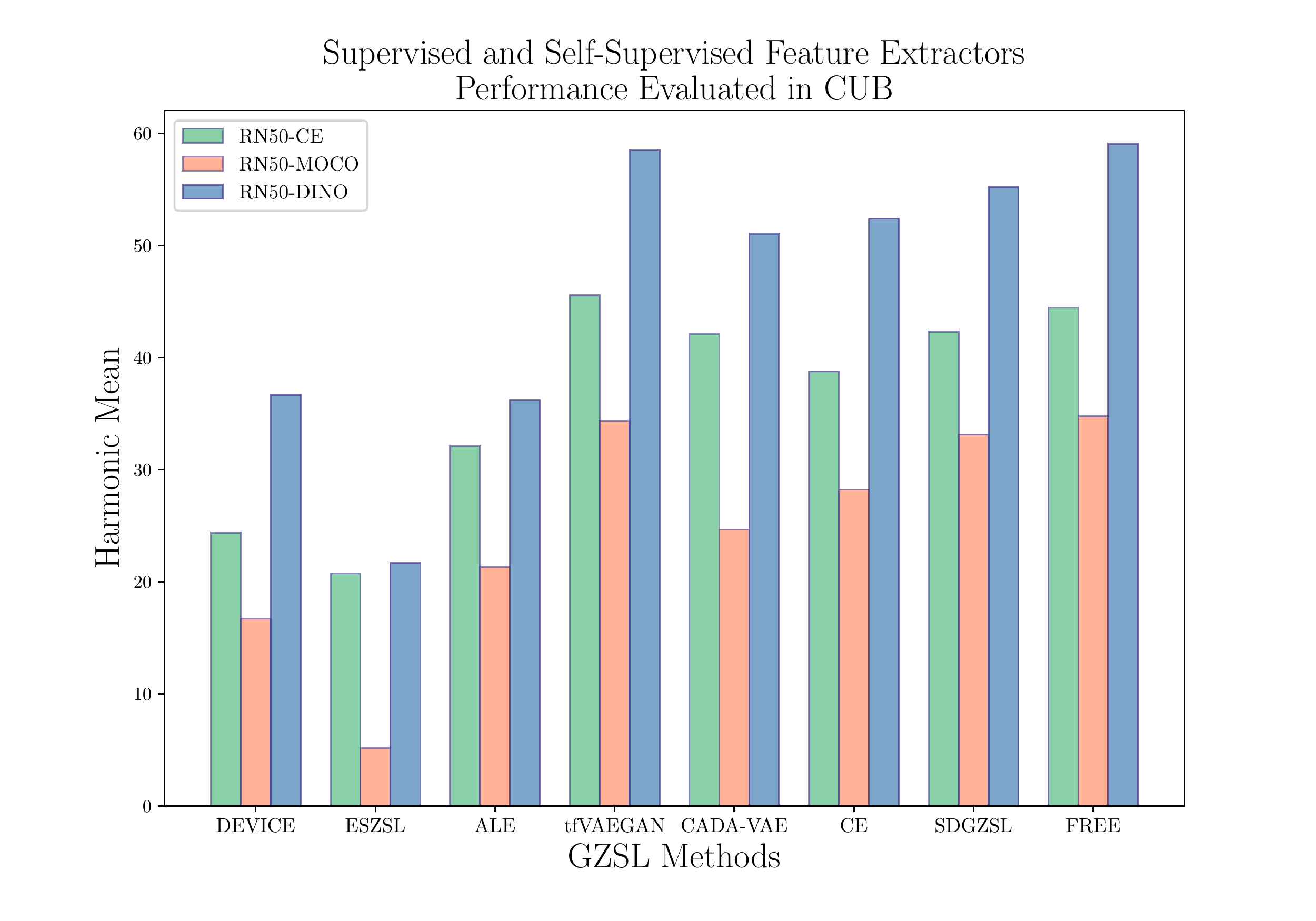}
\end{minipage}%
\hfill 
\begin{minipage}[t]{0.33\textwidth}
  \includegraphics[align=t,width=\linewidth]{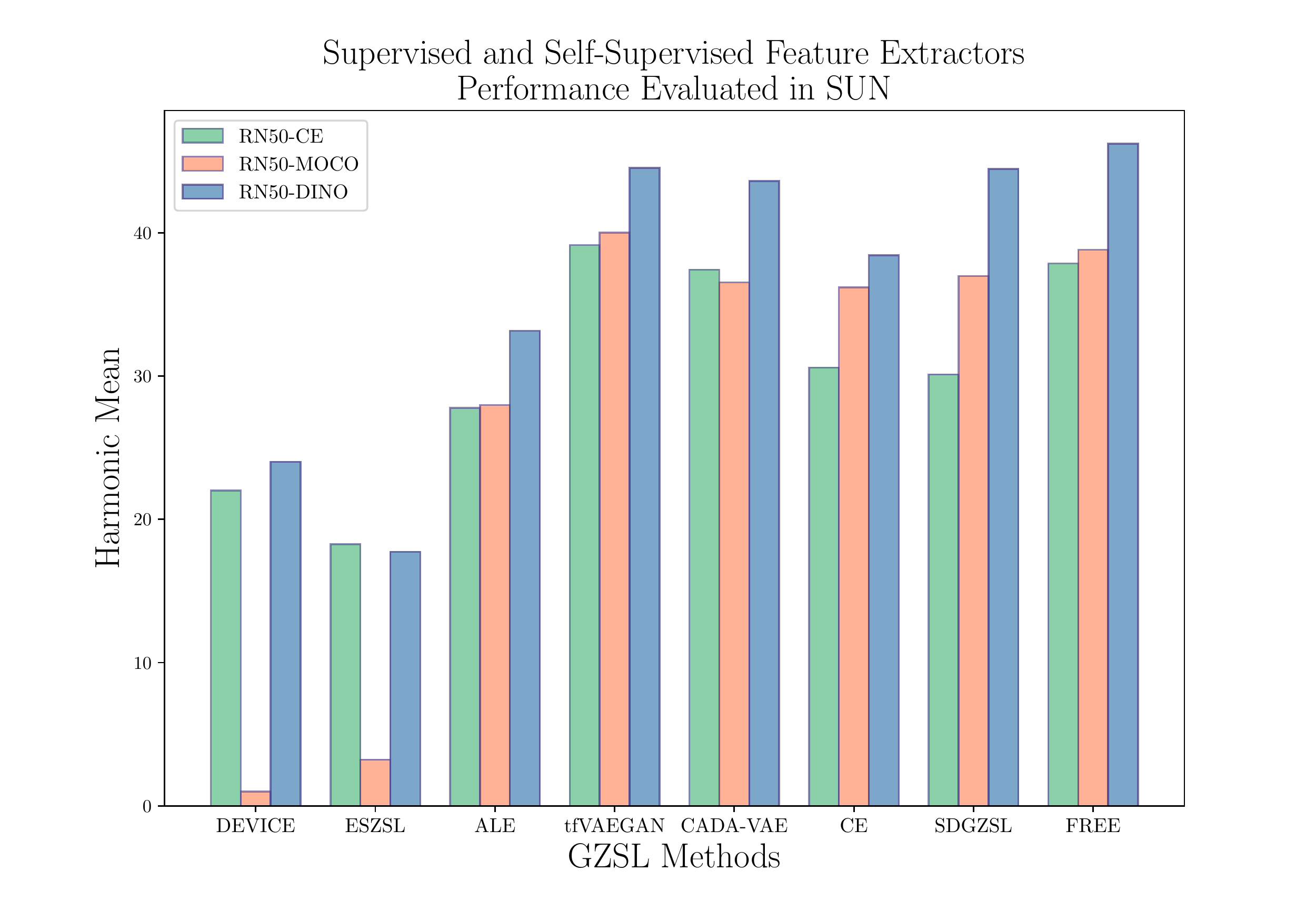}
\end{minipage}%
\hfill
\begin{minipage}[t]{0.33\textwidth}
  \includegraphics[align=t,width=\linewidth]{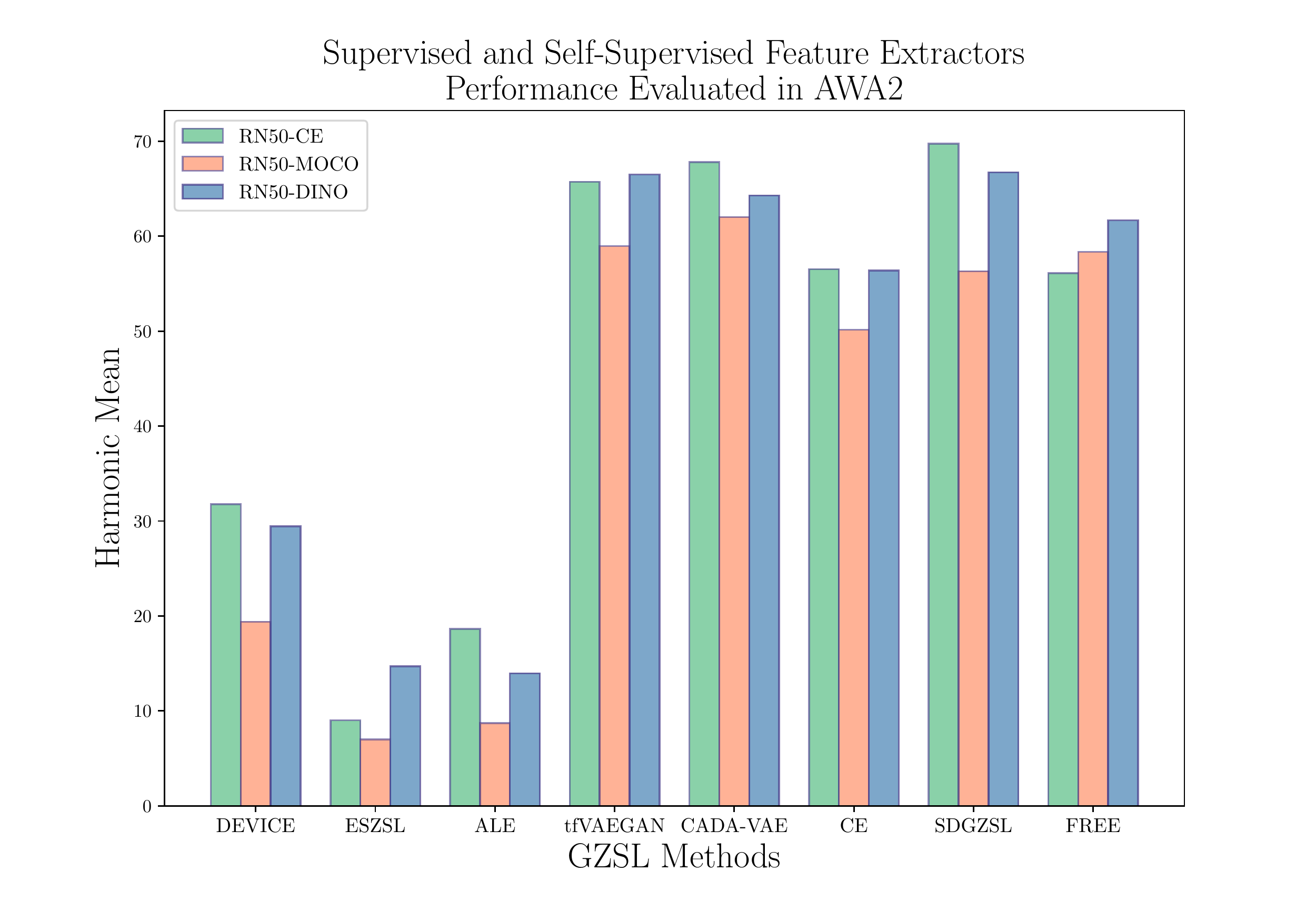}
\end{minipage}%
 \caption{\textbf{Same Feature Extractor Pretrained with Different Learning Objectives.} We show the Harmonic Mean performance of different GZSL methods when using a RN50 model to extract the features of image samples from all datasets. RN50-\underline{CE} indicates that the model was pretrained using categorical cross-entropy in a supervised way, otherwise it was a model pretrained in a self-supervised way using a contrastive loss (RN50-\underline{MOCO}~\cite{MoCo} and RN50-\underline{DINO}~\cite{DINO}). All these models were pretrained using ImageNet-1k. Best viewed in color.}
 \label{fig:model_backbone_diff_objectives}
\end{figure*}
\subsection{GZSL Datasets}

We use three datasets. \textit{Caltech-UCSD Birds-200-2011 (\textbf{CUB})} \cite{CUB}, a fine-grained dataset with $11,788$ images from $200$ different types of birds annotated corresponding to $150$ seen and $50$ unseen classes, with $312$ attributes. \textit{SUN Attribute (\textbf{SUN})} \cite{SUN}, a fine-grained dataset with $14,340$ images from $717$ types of scenes corresponding to $645$ seen and $72$ unseen classes, annotated with $102$ attributes. \textit{Animals with Attributes2 (\textbf{AWA2})} \cite{AWA2}, a dataset with $37,322$ images from $50$ animal classes corresponding to $40$ seen and $10$ unseen classes, annotated with $85$ attributes. We report the \textit{seen} and \textit{unseen} accuracy, and their harmonic mean.

\begin{figure*}[h!]
\scalebox{0.90}{
\begin{minipage}[t]{0.33\textwidth}
  \includegraphics[align=t,width=\linewidth]{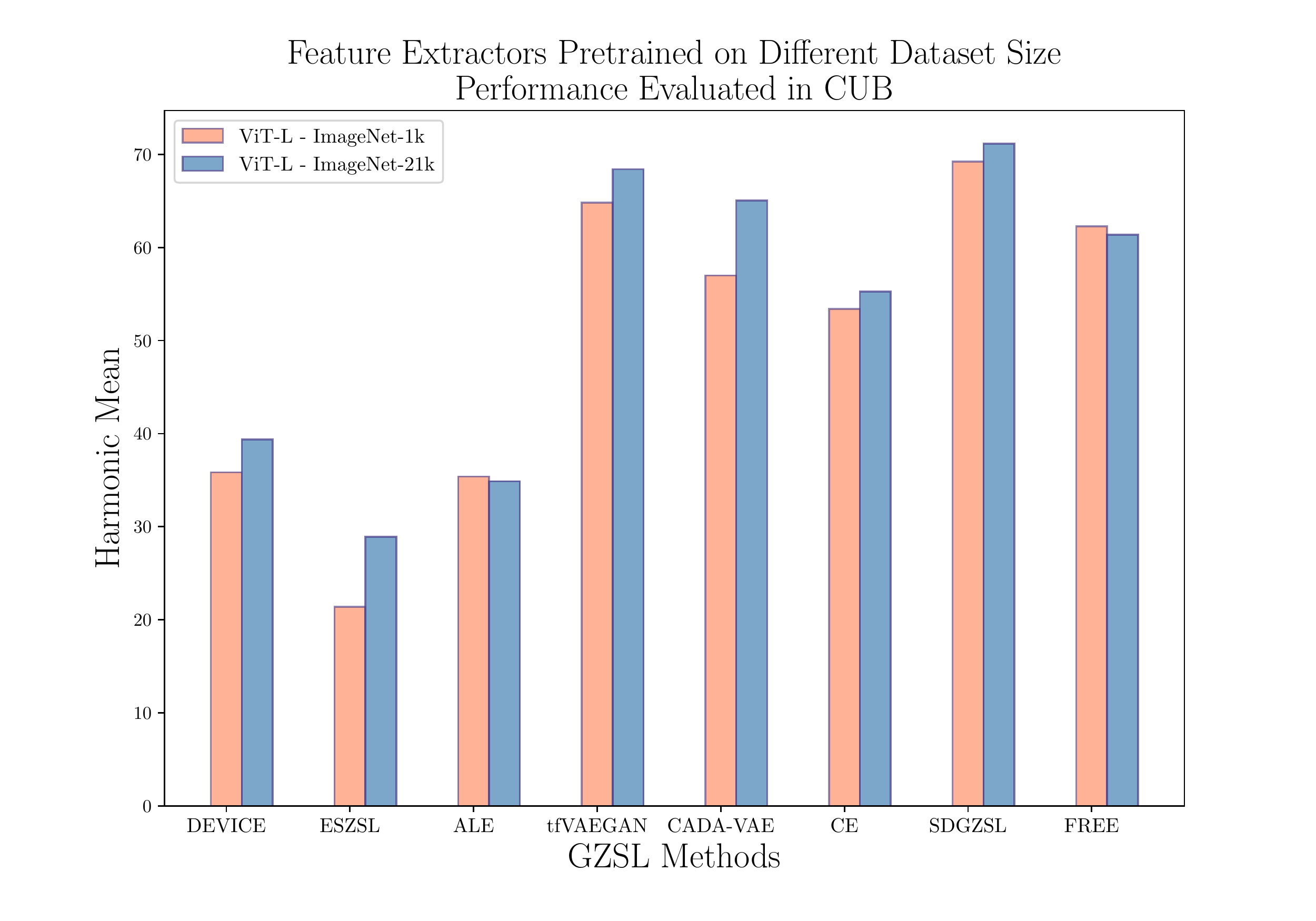}
\end{minipage}%
\hfill 
\begin{minipage}[t]{0.33\textwidth}
  \includegraphics[align=t,width=\linewidth]{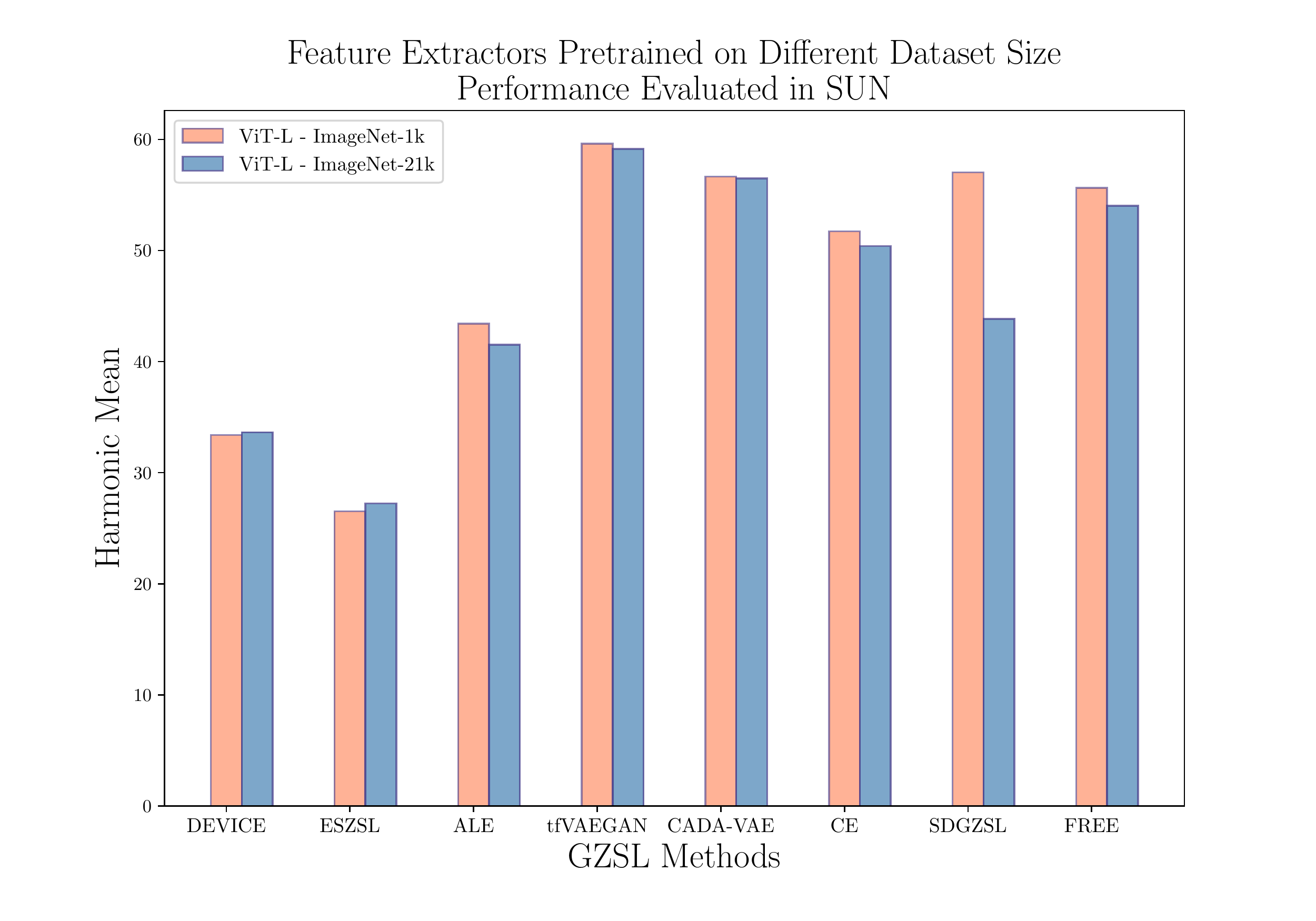}
\end{minipage}%
\hfill
\begin{minipage}[t]{0.33\textwidth}
  \includegraphics[align=t,width=\linewidth]{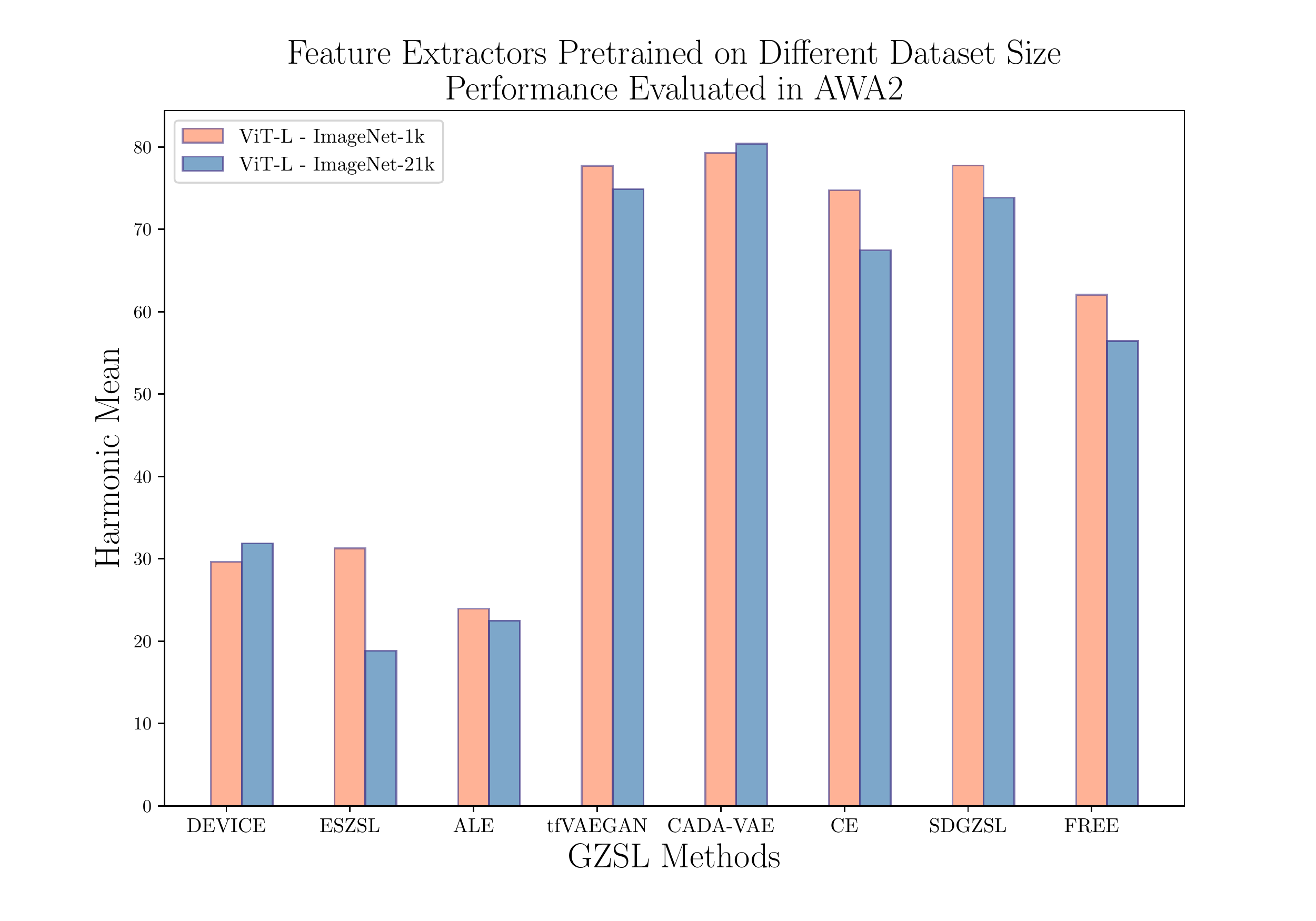}
\end{minipage}%
}
 \vspace{-0.05in}
 \caption{\textbf{Same Feature Extractor Pretrained with Different Datasets.} We show the Harmonic Mean performance of different GZSL methods when using a ViT$_{\text{Large}}$ model to extract the features of image samples from all datasets. [ViT-L - ImageNet-1k] and [ViT-L - ImageNet-21k] indicates the dataset the model was pretrained with. Best viewed in color.}
 \label{fig:model_backbone_data_size}
\end{figure*}

\section{Pre-Trained Image Feature Extractors}

We begin our analysis by extracting the corresponding image features for all our samples using a diverse set of popular models pre-trained on Imagenet-1k and Imagenet-21k -- whose weights are publicly available. Prior work~\cite{AWA2} mention that using GoogLeNet features is not as effective as using Resnet101 features; thus, we want to further investigate: does feature extractor size matters for GZSL? This work aims to evaluate if pre-training on a bigger set impacts the outcomes, and if similarly robust features extracted from different model architectures -- other than Resnet101 -- makes a significant difference. 
For the backbones that were pre-trained using Imagenet-1k, we make sure that we use the same splits proposed in~\cite{AWA2} so that pre-trained features do not violate the zero-shot principle. 
We also want to measure the impact of using visual features extracted from networks that were pre-trained with data present in the test set of our settings. Thus, we investigate how the selected GZSL methods leverage the information extracted from larger backbones pre-trained with bigger and more diverse datasets (i.e., Imagenet-21k).

\subsection{Unimodal Feature Backbones}

Overall, the pool of networks in this study have been trained using two learning objectives: 
\textit{Supervised Learning} (SL), which aims to learn a function that maps an input to a \textit{known output}, and is typically trained using a \textit{cross-entropy} loss (e.g., Resnet101~\cite{RNs}) or \textit{distillation} (e.g., DeiT~\cite{DeiT}). On the other hand, \textit{Self-Supervised Learning} (SSL), aims to learn a function that maps an input to a \textit{unknown output}; SSL can be accomplished using a \textit{contrastive loss} such as InfoNCE (e.g., MoCo~\cite{MoCo}) or adding \textit{distillation} along with a similarity metric measured with a cross-entropy loss applied over the features of two different random transformations of an input image (e.g., DINO~\cite{DINO}). 
We perform experiments on three broad types of unimodal architectures trained on images only (Imagenet-1k or Imagenet-21k):
\textit{Convolutional Neural Networks} (CNN)~\cite{Alexnet} trained using the images as a whole under SL or SSL, 
\textit{Vision Transformers} (ViT)~\cite{ViT} which takes an image, transforms it in a sequence of image patches and is trained using either SL or SSL; and
\textit{Multi-Layer Perceptron Mixers} (MLP)~\cite{MLPMixer} which also exploit image patches and are trained with supervision.

We chose the most performant models and fine-tuned them using only the training samples from the seen classes. Since the visual features are extracted from a diverse set of architectures trained in the wild, we focus on the \textit{inductive} setting, where there is no access to the unlabeled visual data from the unseen classes. In this way, we mitigate any bias reinforced by additional training of the visual representations. 
This practice has been followed to achieve better results in prior literature (which only uses Resnet101 features); however, there are no available reports for all the methods using these features; thus, we run all methods and report our findings in Section~\ref{sec:results_unimodal}.

\subsection{Multimodal Feature Backbones}

In addition, we study CLIP~\cite{CLIP}, a multi-modal model with a visual encoder and a textual encoder trained with 400 million image/text pairs in a contrastive way. Traditionally, all GZSL methods disregard the attribute values as they are given for granted, and no additional analysis is performed. We instead take a closer look at these semantic features and their corresponding attributes (e.g., color, shape, type of animal, type of place, etc)  and use them to fine-tune several pre-trained CLIP models. 

We perform three experiments with CLIP: 
(A) We directly evaluate the model using the images and class names without any further pre-training or post-processing by directly looking at the ranking the model yields for each seen and unseen samples,
(B) We fine-tune CLIP using the class names, the attribute values and a combination of class names + attribute values, and evaluate its performance, 
(C) We extract the visual features from the CNN and ViT based visual backbones available in their public repository, and use the features extracted from (A) and (B) to train the GZSL methods.
We go over our findings in Section~\ref{sec:results_multimodal}.



\begin{table*}[bp]
\newcolumntype{Y}{>{\raggedright\arraybackslash}X}
\newcolumntype{Z}{>{\centering\arraybackslash}X}
\centering
\footnotesize
\setlength\tabcolsep{1pt}
\renewcommand{\arraystretch}{1.2}

\begin{tabularx}{\textwidth}{l c YYY c YYY c YYY}
\toprule


{\multirow{2}{*}{\bf Backbone}}~~~ &~~~~&
\multicolumn{3}{@{\hskip 0.11in}c}{\bf CUB} &~~~~& 
\multicolumn{3}{@{\hskip 0.11in}c}{\bf SUN} &~~~~& 
\multicolumn{3}{@{\hskip 0.11in}c}{\bf AWA2} \\

\cmidrule{3-5}\cmidrule{7-9}\cmidrule{11-13}

&& \textit{Seen} & \textit{Novel} & \textit{Harm.} 
&& \textit{Seen} & \textit{Novel} & \textit{Harm.} 
&& \textit{Seen} & \textit{Novel} & \textit{Harm.} \\

\midrule

RN50 &  & 
45.90 & 45.44 & 45.67 &&
44.61 & 48.96 & 46.68 &&
87.86 & 82.48 & 85.08 \\ 

RN101 &  & 
48.86 & 49.44 & 49.15 && 
45.16 & 49.24 & 47.11 && 
88.66 & 84.79 & 86.67 \\ 

RN50x4 &  & 
51.89 & 55.27 & 53.53 && 
48.53 & 50.56 & 49.52 && 
92.09 & 86.52 & 89.22 \\ 

RN50x16 &  & 
56.82 & 55.19 & 55.99 && 
49.07 & 54.44 & 51.62 &&
94.36 & 89.11 & 91.65 \\ 

RN50x64 &  & 
63.81 & 57.19 & 60.32 && 
55.28 & 51.59 & 53.37 && 
95.11 & 90.11 & 92.54 \\ 

ViTB32 &  & 
51.35 & 49.65 & 50.49 && 
48.26 & 50.97 & 49.58 && 
90.99 & 85.69 & 88.26 \\ 

ViTB32$^{\dag}$ &   & 
61.46 & 58.09 & 59.73 && 
50.66 & 53.75 & 52.16 && 
87.84 & 82.12 & 84.88 \\

ViTB16 &  & 
55.76 & 56.23 & 55.99 &&
50.97 & 56.11 & 53.42 && 
93.68 & 87.19 & 90.32 \\ 

ViTL14 &  & 
62.62 & 63.19 & 62.90 &&
55.97 & 58.06 & 56.99 && 
95.80 & 89.39 & 92.48 \\


\midrule
\band
ViTL14$^{\Uparrow}$ & & 
\textbf{64.45} & \textbf{62.69} & \textbf{63.56} && 
\textbf{57.79} & \textbf{62.64} & \textbf{60.12} && 
\textbf{96.06} & \textbf{89.91} & \textbf{92.88} \\

\bottomrule
\end{tabularx}
\caption{Results of using publicly available pre-trained CLIP\cite{CLIP} models with different backbones to evaluate three standard GSZL datasets. ${\Uparrow}$ indicates we used a set of captions similar to the proposed by OpenAI to test on Imagenet\cite{Imagenet}, ${\dag}$ indicates the model used was trained on the Laion400M\cite{LAION400M} dataset.
}
\label{tab:all_datasets_clip_only}
\vspace{-0.0in}
\end{table*}

\begin{table*}[bp] 
\newcolumntype{Y}{>{\raggedright\arraybackslash}X}
\newcolumntype{Z}{>{\centering\arraybackslash}X}
\centering
\footnotesize
\setlength\tabcolsep{1pt}
\renewcommand{\arraystretch}{1.2}

\begin{tabularx}{\textwidth}{cl c YYY c YYY c YYY c YYY c YYY}
\toprule

\multicolumn{22}{@{\hskip 0.11in}c}{\bf GZSL Methods using CLIP Visual Features} \\
\midrule

{\multirow{2}{*}{\bf \shortstack{Data \\set }}} &
{\multirow{2}{*}{\bf \shortstack{Back \\bone }}} &
\multicolumn{3}{@{\hskip 0.11in}c}{\bf tfVAEGAN} &~~& 
\multicolumn{3}{@{\hskip 0.11in}c}{\bf CADA-VAE} &~~& 
\multicolumn{3}{@{\hskip 0.11in}c}{\bf SDGZSL} &~~& 
\multicolumn{3}{@{\hskip 0.11in}c}{\bf FREE} &~~& 
\multicolumn{3}{@{\hskip 0.11in}c}{\bf CE} \\

\cmidrule{4-6}\cmidrule{8-10}\cmidrule{12-14}\cmidrule{16-18}\cmidrule{20-22}

&&& \textit{Seen} & \textit{Novel} & \textit{Harm.} 
&& \textit{Seen} & \textit{Novel} & \textit{Harm.} 
&& \textit{Seen} & \textit{Novel} & \textit{Harm.} 
&& \textit{Seen} & \textit{Novel} & \textit{Harm.} 
&& \textit{Seen} & \textit{Novel} & \textit{Harm.} \\

\midrule





\parbox[t]{1mm}{\multirow{5}{*}{\rotatebox[origin=c]{90}{CUB}}}
& {R50$_{\text{x64}}$} &  & 
82.09 & 67.93 & 74.34  & ~ &
75.91 & 70.84 & 73.29  & ~ & 
74.77 & 73.70 & 74.23  & ~ &
68.53 & \underline{73.47} & 70.91  & ~ &
59.06 & 49.52 & 53.87  \\



&{ViT$_{\text{L14}}$} &  & 
\underline{77.67} & \underline{72.36} & \underline{74.92}  & ~ &
\underline{77.90} & \underline{72.98} & \underline{75.36}  & ~ &
\underline{79.50} & 73.49 & \underline{76.38}  & ~ &
\underline{79.05} & 65.38 & \underline{71.56}  & ~ &
\underline{74.04} & 46.38 & 57.03  \\

\arrayrulecolor{gray!48} \cmidrule{2-22} 
\arrayrulecolor{black}

&{ViT$_{\text{L14}}^\dag$} &  & 
80.39 & 72.86 & 76.44  & ~ &
\textbf{81.96} & 71.24 & 76.22  & ~ & 
79.19 & 74.70 & 76.88  & ~ &
78.87 & 65.89 & 71.80  & ~ &
71.49 & 48.62 & 57.88  \\

&{ViT$_{\text{L14}}^\ddag$} &  & 
80.73 & \textbf{73.59} & 76.99  & ~ &
79.51 & 74.68 & 77.01  & ~ &
\textbf{80.40} & 74.33 & 77.25  & ~ &
79.47 & \textbf{68.25} & 73.44  & ~ &
\textbf{75.12} & 51.14 & 60.86 \\

&\cellcolor{gray!18} {ViT$_{\text{L14}}^\S$} & \cellcolor{gray!18}  & 
\cellcolor{gray!18} \textbf{82.82} & \cellcolor{gray!18} 72.27 & \cellcolor{gray!18} \textbf{77.18}  & \cellcolor{gray!18} ~ &
\cellcolor{gray!18} 81.47 & \cellcolor{gray!18} \textbf{75.05} & \cellcolor{gray!18} \textbf{78.13}  & \cellcolor{gray!18} ~ &
\cellcolor{gray!18} 80.38 & \cellcolor{gray!18} \textbf{75.67} & \cellcolor{gray!18} \textbf{77.96}  & \cellcolor{gray!18} ~ &
\cellcolor{gray!18} \textbf{80.48} & \cellcolor{gray!18} 67.54 & \cellcolor{gray!18} \textbf{73.45}  & \cellcolor{gray!18} ~ &
\cellcolor{gray!18} 71.15 & \cellcolor{gray!18} \textbf{56.82} & \cellcolor{gray!18} \textbf{63.18} \\


\midrule





\parbox[t]{1mm}{\multirow{5}{*}{\rotatebox[origin=c]{90}{SUN}}}
&{R50$_{\text{x64}}$} &  & 
57.21 & \underline{69.79} & 62.88  & ~ &
59.46 & 65.63 & 62.39  & ~ &
57.17 & 66.60 & 61.52  & ~ &
\underline{60.04} & 57.78 & 58.89  & ~ &
50.04 & 59.10 & 54.19  \\



&{ViT$_{\text{L14}}$} &  & 
\underline{59.84} & 68.89 & \underline{64.05}  & ~ &
\underline{62.11} & \underline{63.40} & \underline{63.18}  & ~ &
\underline{63.57} & 62.71 & \underline{63.13}  & ~ &
58.18 & \underline{62.71} & \underline{60.36}  & ~ &
\underline{60.16} & 57.85 & \underline{58.98}  \\

\arrayrulecolor{gray!48} \cmidrule{2-22}
\arrayrulecolor{black}

&{ViT$_{\text{L14}}^\dag$} &  & 
61.75 & \textbf{70.49} & 65.83  &\cellcolor{gray!18}  ~ &
\cellcolor{gray!18} \textbf{61.47} &\cellcolor{gray!18}  65.76 &\cellcolor{gray!18}  \textbf{63.54}  & ~ &
\textbf{62.56} & 64.44 & 63.49  & ~ &
\textbf{58.95} & 62.99 & 60.90  & ~ &
52.79 & \textbf{64.37} & 58.01  \\

&{ViT$_{\text{L14}}^\ddag$} &  & 
60.93 & 67.08 & 63.86  & ~ &
57.95 & \textbf{68.19} & 62.65  & ~ &
60.08 & 67.15 & 63.42  & ~ &
58.37 & 62.43 & 60.33  & ~ &
56.78 & 64.03 & 60.19  \\

&{ViT$_{\text{L14}}^\S$} &\cellcolor{gray!18}  & 
\cellcolor{gray!18} \textbf{62.25} &\cellcolor{gray!18}  70.03 &\cellcolor{gray!18}  \textbf{65.91}  & ~ &
58.91 & 66.74 & 62.58  &\cellcolor{gray!18}  ~ &
\cellcolor{gray!18} 60.70 &\cellcolor{gray!18}  \textbf{67.50} &\cellcolor{gray!18}  \textbf{63.92}  &\cellcolor{gray!18}  ~ &
\cellcolor{gray!18} 58.72 &\cellcolor{gray!18}  \textbf{63.61} &\cellcolor{gray!18}  \textbf{61.07}  &\cellcolor{gray!18}  ~ &
\cellcolor{gray!18} \textbf{59.65} &\cellcolor{gray!18}  60.90 &\cellcolor{gray!18}  \textbf{60.27}  \\


\midrule





\parbox[t]{1mm}{\multirow{5}{*}{\rotatebox[origin=c]{90}{AWA2}}}
&{R50$_{\text{x64}}$} &  & 
89.69 & 68.21 & 77.49  & ~ &
90.54 & 63.75 & 75.81  & ~ &
78.67 & 69.07 & 73.56  & ~ &
\underline{87.60} & 61.04 & 71.95  & ~ &
89.95 & 63.33 & 74.33  \\



&{ViT$_{\text{L14}}$} &  & 
\underline{90.95} & 69.50 & \underline{78.70}  & ~ &
\underline{92.68} & 68.14 & \underline{77.99}  & ~ &
\underline{89.99} & 68.76 & 77.91  & ~ &
81.10 & \underline{65.99} & \underline{72.77}  & ~ &
\underline{85.63} & \underline{71.96} & \underline{78.20}  \\

\arrayrulecolor{gray!48} \cmidrule{2-22}
\arrayrulecolor{black}

&{ViT$_{\text{L14}}^\dag$} &  & 
91.25 & 67.77 & 77.77  & ~ &
90.48 & 70.07 & 78.98  & ~ &
89.18 & 71.40 & 79.30  &\cellcolor{gray!18} ~ &
\cellcolor{gray!18}81.27 &\cellcolor{gray!18} \textbf{71.56} &\cellcolor{gray!18} \textbf{75.94}  &\cellcolor{gray!18} ~ &
\cellcolor{gray!18}85.12 &\cellcolor{gray!18} \textbf{78.02} &\cellcolor{gray!18} \textbf{81.41}  \\

&{ViT$_{\text{L14}}^\ddag$} &\cellcolor{gray!18}  & 
\cellcolor{gray!18}\textbf{93.40} &\cellcolor{gray!18} 72.61 &\cellcolor{gray!18} \textbf{81.70}  &\cellcolor{gray!18} ~ &
\cellcolor{gray!18}\textbf{93.51} &\cellcolor{gray!18} \textbf{73.85} &\cellcolor{gray!18} \textbf{82.53}  &\cellcolor{gray!18} ~ &
\cellcolor{gray!18}85.77 &\cellcolor{gray!18} \textbf{74.16} &\cellcolor{gray!18} \textbf{79.54}  & ~ &
81.29 & 65.62 & 72.62  & ~ &
89.85 & 64.12 & 74.84  \\

&{ViT$_{\text{L14}}^\S$} &  & 
89.62 & \textbf{73.56} & 80.80  & ~ &
93.35 & 68.14 & 78.78  &\cellcolor{gray!18} ~ &
\cellcolor{gray!18}\textbf{91.50} &\cellcolor{gray!18} 70.33 &\cellcolor{gray!18} \textbf{79.54}  & ~ &
\textbf{81.53} & 69.84 & 75.23  & ~ &
\textbf{90.78} & 67.96 & 77.73  \\

\bottomrule
\end{tabularx}
\caption{Results of Generative and Disentanglement Based Methods for the CUB, SUN and AWA2 datasets using different features extracted from different size and architecture of the visual head from diverse CLIP models (i.e., Resnet50 (R50) and Vision Transformer (ViT)). The three bottom rows per section correspond to fine-tuned features using sentences with: $\dag$ the class names, $\ddag$ the attributes, and $\S$ both class names and attributes. The bold numbers correspond to the highest scores per column, the underline numbers correspond to the to the highest scores using features not fine-tuned, and the shaded rows correspond to the most performant image feature per method over all.}
\label{tab:gzsl_using_clip_features}
\end{table*}

\section{Results using Unimodal Feature Extractors}
\label{sec:results_unimodal}

We evaluate all GZSL methods trained with feature extractors pre-trained using ImageNet-1k. In Figure~\ref{fig:model_backbone_size_vs_method}, we show the Harmonic Mean performance of different methods when using a specific model to extract the features of image samples. Surprisingly, CADA-VAE~\cite{CADA_VAE} and tfVAEGAN~\cite{tfvaegan} consistently outperform all methods. While current disentanglement methods show significant improvements when using features from transformer-based architectures, they are outperformed by the generative-based methods.
We can also observe that DINO~\cite{DINO} provides better
feature representations for all methods across all datasets.

We also want to evaluate the impact of using a feature extractor with the \textit{same architecture type but trained with different learning objectives}. In Figure~\ref{fig:model_backbone_diff_objectives}, we show the Harmonic Mean performance of
different GZSL methods when using a Resnet model architecture pre-trained on Imagenet-1k as the image feature extractor. Surprisingly, the features extracted from DINO~\cite{DINO} increase the Harmonic Mean performance up to 15\% in both fine-grained datasets (i.e., CUB and SUN datasets). More surprisingly, the feature vectors extracted from MOCO perform worse than traditional supervised learning models trained with a cross-entropy objective function. MOCO's training objective is formulated by the InfoNCE loss, which encourages the model to maximize the Mutual Information (MI) between $N$ random samples containing one positive sample, and minimize the Mutual information between the anchor sample and $N-1$ negative samples. 
On the other hand, in DINO, a teacher and a student model are trained by feeding two different random transformations of an input image to each network; the objective is to maximize the similarity between both outputs, which is encouraged and measured with a cross-entropy loss. 
Thus, the generalization capabilities shown with DINO support prior observations when using its image features in classification tasks with respect to other self-supervised techniques~\cite{DINO}.

We then evaluate the impact of using feature extractors that were pre-trained with more data variety and size and summarize our findings in Figure~\ref{fig:model_backbone_data_size}. Surprisingly, features extracted from backbones trained with more data (Imagenet-21k) do not always perform better, but features extracted from bigger architectures seem consistently better.

We show more detailed results on CUB, SUN and AWA2 using GZSL methods grouped in their corresponding families in the following subsections.
Please refer to the Appendix to check the full list of numerical results for all uni-modal backbones and GZSL methods. 

\subsection{Results of \textbf{Embedding}-based Methods}

The ViT$_{\text{huge}}$ features pre-trained on ImageNet-21k seem to be the best for all the methods using ALE. 
However, for the AWA2 dataset, all methods perform better using the features extracted from a network pre-trained using ImageNet-1k.
For CUB and SUN datasets, the performance gap against the features extracted from a network trained using ImageNet-1k and ImageNet-21k is not significant for all methods.
More detailed results are available in Tables~\ref{tab:cub_embedding_CNN}, \ref{tab:sun_embedding_CNN} and \ref{tab:awa2_embedding_CNN} from Section~\ref{sec:all_tables} in the Appendix.

\subsection{Results of \textbf{Generative}-based Methods} 

The most performant visual features are extracted from a ViT$_{\text{huge}}$ pretrained on ImageNet-21k and fine-tuned with the seen classes, using the CADA-VAE method. More interestingly, the features from a ViT$_{\text{large}}$ pretrained on ImageNet-1k seem competitive with the features from a ViT$_{\text{large}}$ pretrained on ImageNet-21k for the CE and tfVAEGAN methods respectively.
More detailed results are available in Tables~\ref{tab:cub_generative_CNN}, \ref{tab:sun_generative_CNN} and \ref{tab:awa2_generative_CNN} from Section~\ref{sec:all_tables} in the Appendix. 

\subsection{Results of \textbf{Disentanglement}-based Methods} 

The most performant visual features are extracted from a ViT$_{\text{large}}$ pre-trained on ImageNet-1k using the SDGZSL method. 
Here, the features extracted from architectures pre-trained using ImageNet-21k perform worse than the ones pre-trained using ImageNet-1k, except for the CUB dataset.
More detailed results are available in Tables~\ref{tab:cub_disentanglement_CNN}, \ref{tab:sun_disentanglement_CNN} and \ref{tab:awa2_disentanglement_CNN} from Section~\ref{sec:all_tables} in the Appendix.

\begin{figure*}[hbt!]
\centering
\scalebox{0.90}{
\includegraphics[width=.3\textwidth]{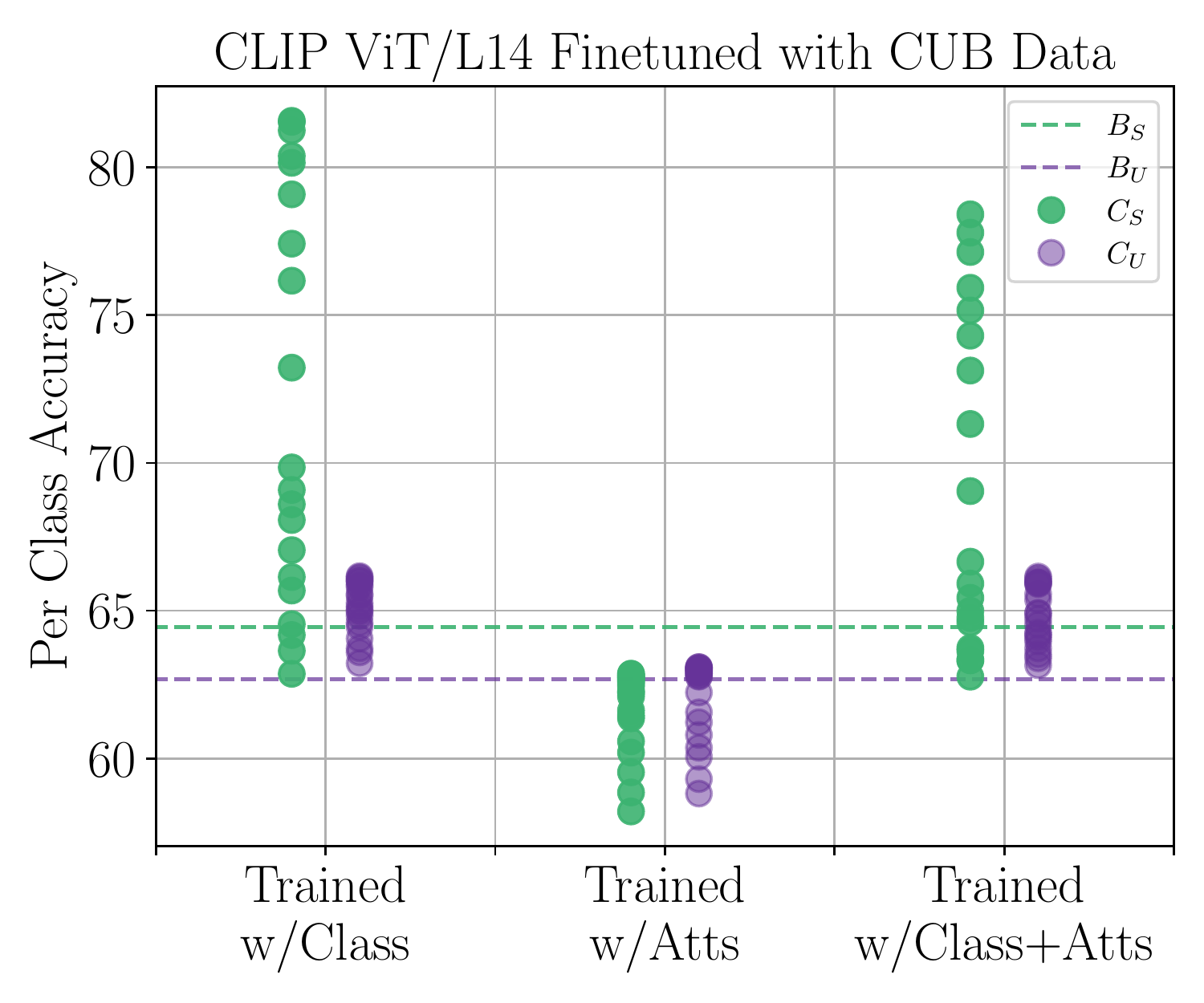}\hfill
\hspace{1cm}
\includegraphics[width=.3\textwidth]{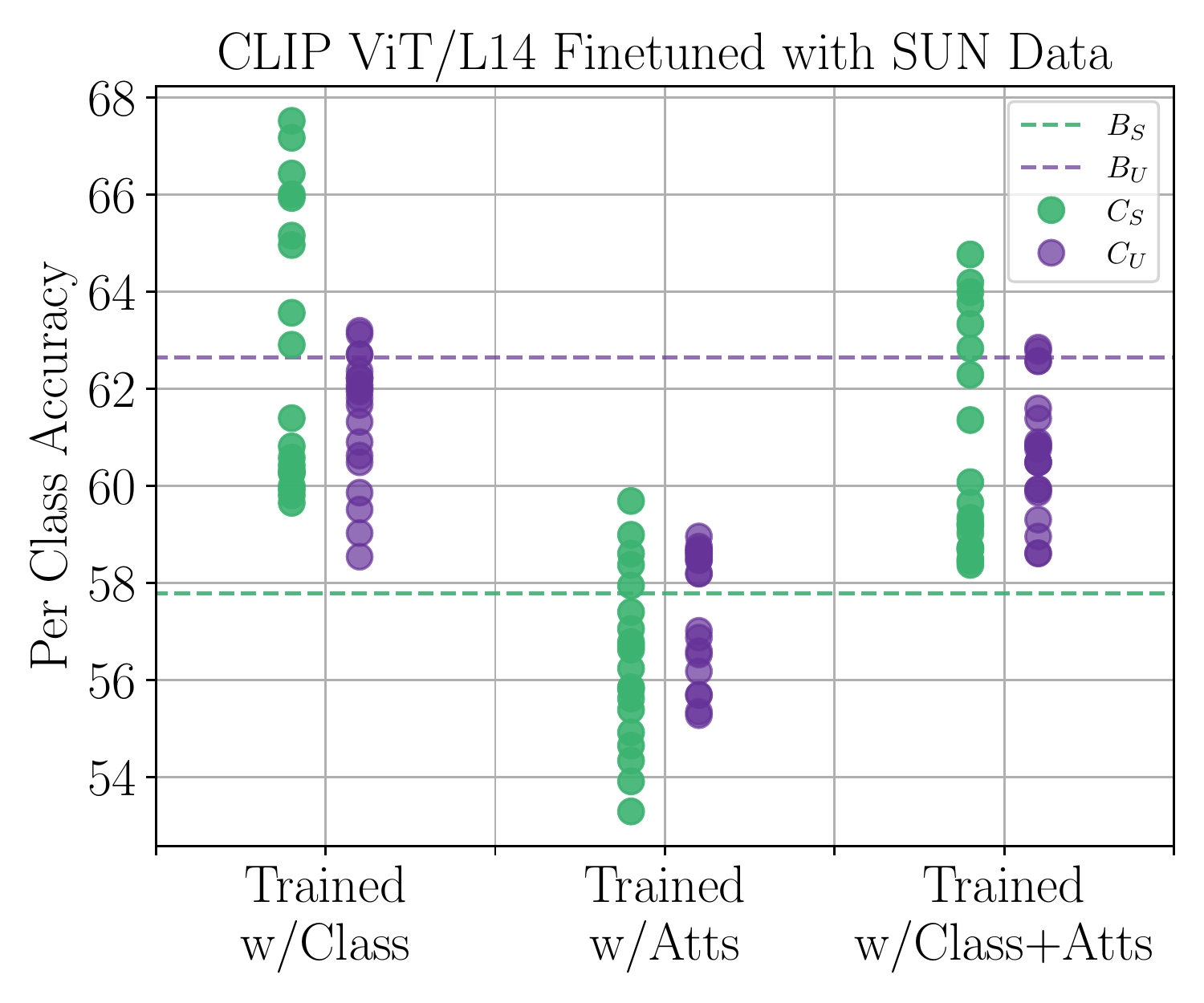}\hfill
\hspace{1cm}
\includegraphics[width=.3\textwidth]{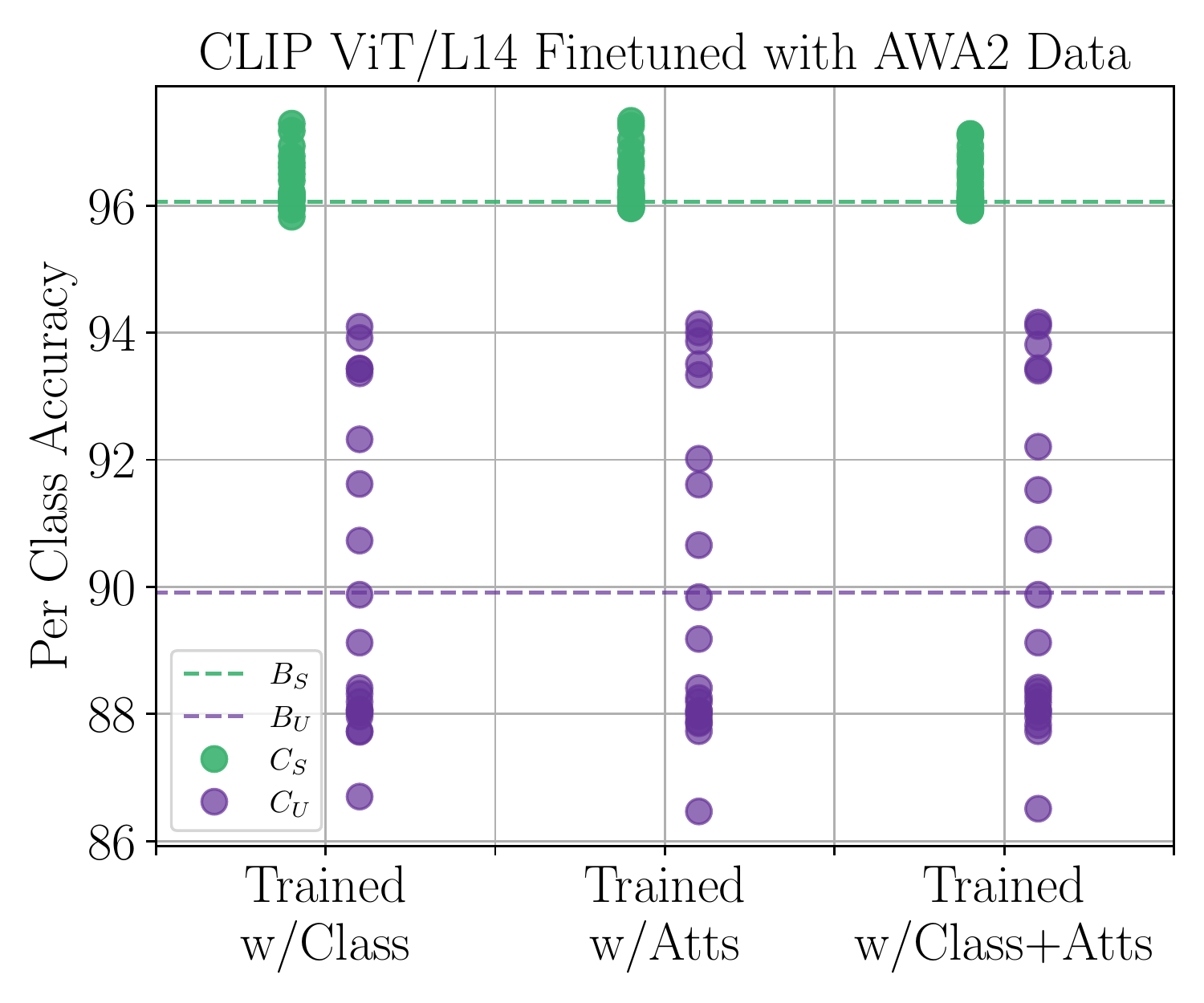}
}
\caption{\textbf{Fine-tuning CLIP with seen samples}. Results of fine-tuning a CLIP model using the images and different text descriptions, including the class name, the attribute text, and combining both text captions. We show accuracy variations after 110k training iterations for CUB\cite{CUB}, SUN\cite{SUN} and AWA2\cite{AWA2}, where B$_S$ and B$_S$ indicate the base performance of seen and unseen classes without fine-tuning. C$_S$ refers to the seen classes, and C$_U$ refers to the unseen (novel) classes after fine-tuning.} 
\label{fig:finetuned_clip_accs}
\vspace{-0.1in}
\end{figure*}

\section{Results using CLIP as a Multimodal Feature Extractor} 
\label{sec:results_multimodal}
\textbf{Direct evaluation of CLIP} using the images and class names without any further pre-training or post-processing: we use different template captions to generate the textual descriptions (e.g., \textit{"An image of a [class name]"}) and chose to evaluate the best-performing model with the template captions proposed by OpenAI to test on Imagenet~\cite{openai_2022}.
We also evaluate a CLIP model trained with the publicly available LAION-400M dataset~\cite{LAION400M} in the publicly available visual transformer backbone (i.e.~ViTB32). We show results in Table~\ref{tab:all_datasets_clip_only}. Overall, we expected CLIP to perform well in all the selected datasets, even with CUB, whose class names are particularly specific; similar results on this dataset have been reported in concurrent work~\cite{Vogel2022VLTabooAA}. Moreover, our results show that generative-based models work on par and even outperform CLIP when using the Resnet101 fine-tuned features, indicating that there might be room for improvement.

\textbf{Evaluation of CLIP performance after fine-tuning} using the class
names, the attribute values and a combination of class names
$\&$ attribute values: Figure~\ref{fig:finetuned_clip_accs} shows the accuracy
variations after fine-tuning CLIP for 110k iterations using different types of text prompts, using only the seen training set for each dataset. Interestingly, we observed that fine-tuning enhances both the seen and unseen accuracy.

\begin{figure*}[hbt!]
\centering
\scalebox{0.94}{
\includegraphics[width=.33\textwidth]{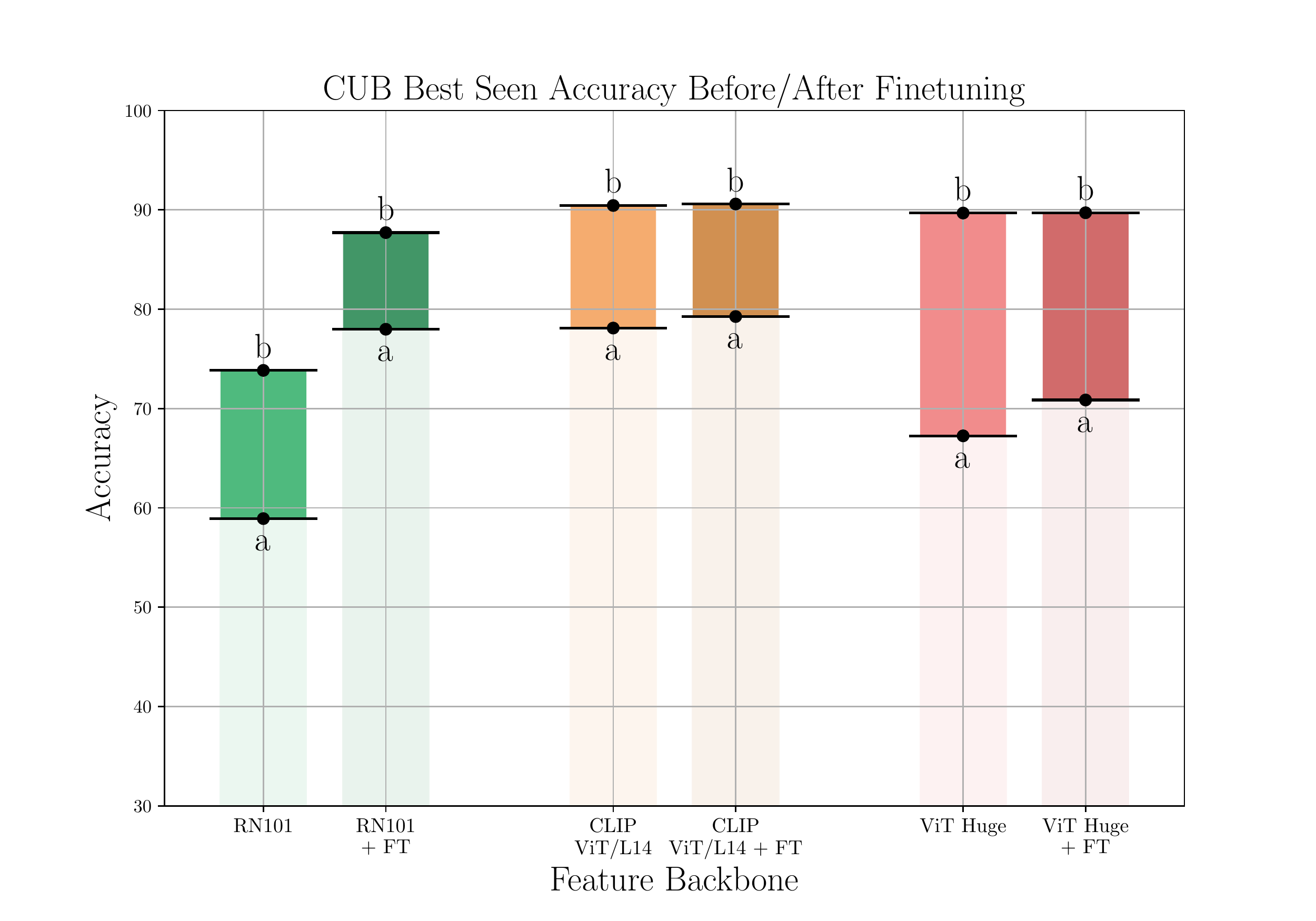}\hfill
\hspace{0.5cm}
\includegraphics[width=.33\textwidth]{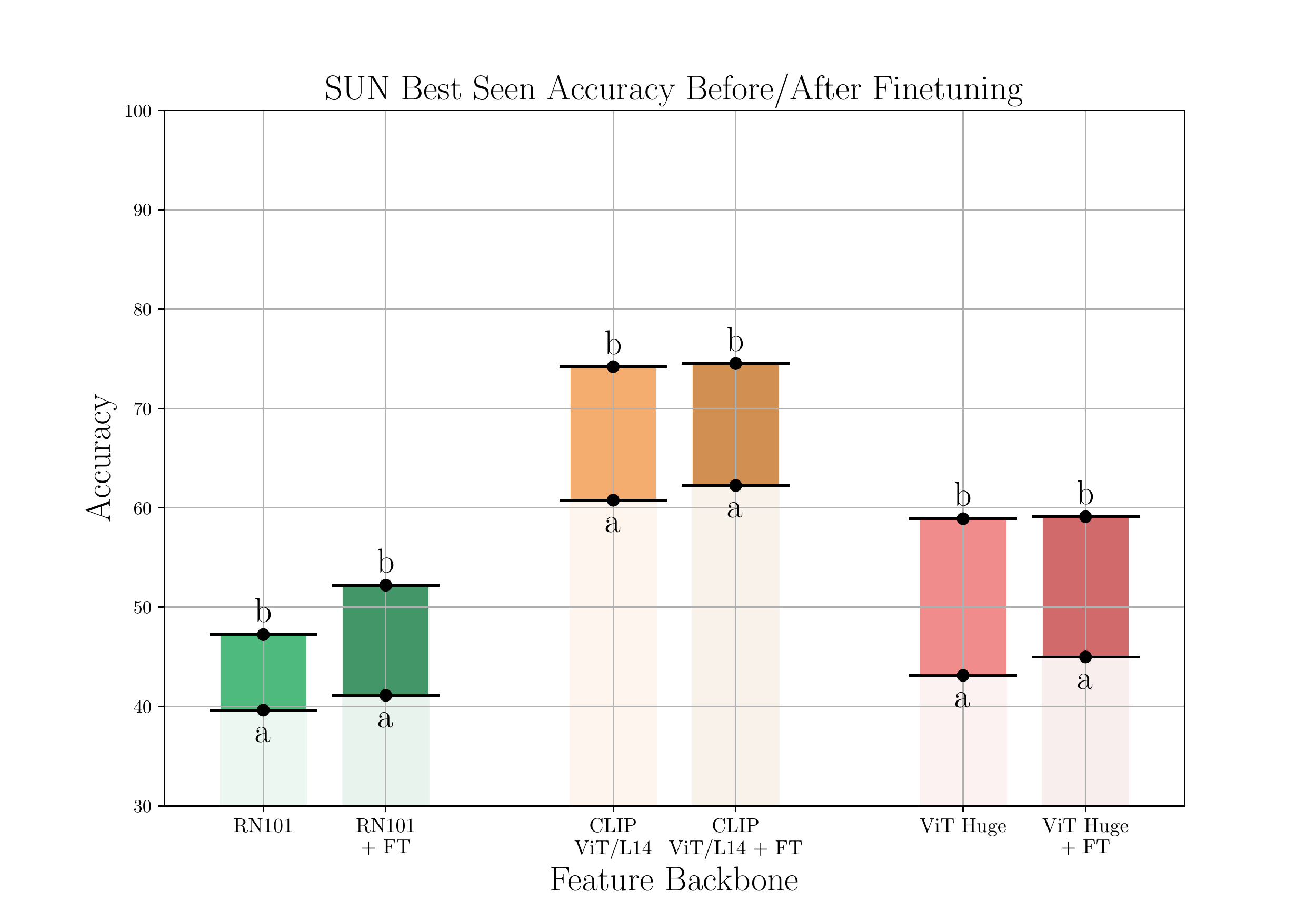}\hfill
\hspace{0.5cm}
\includegraphics[width=.33\textwidth]{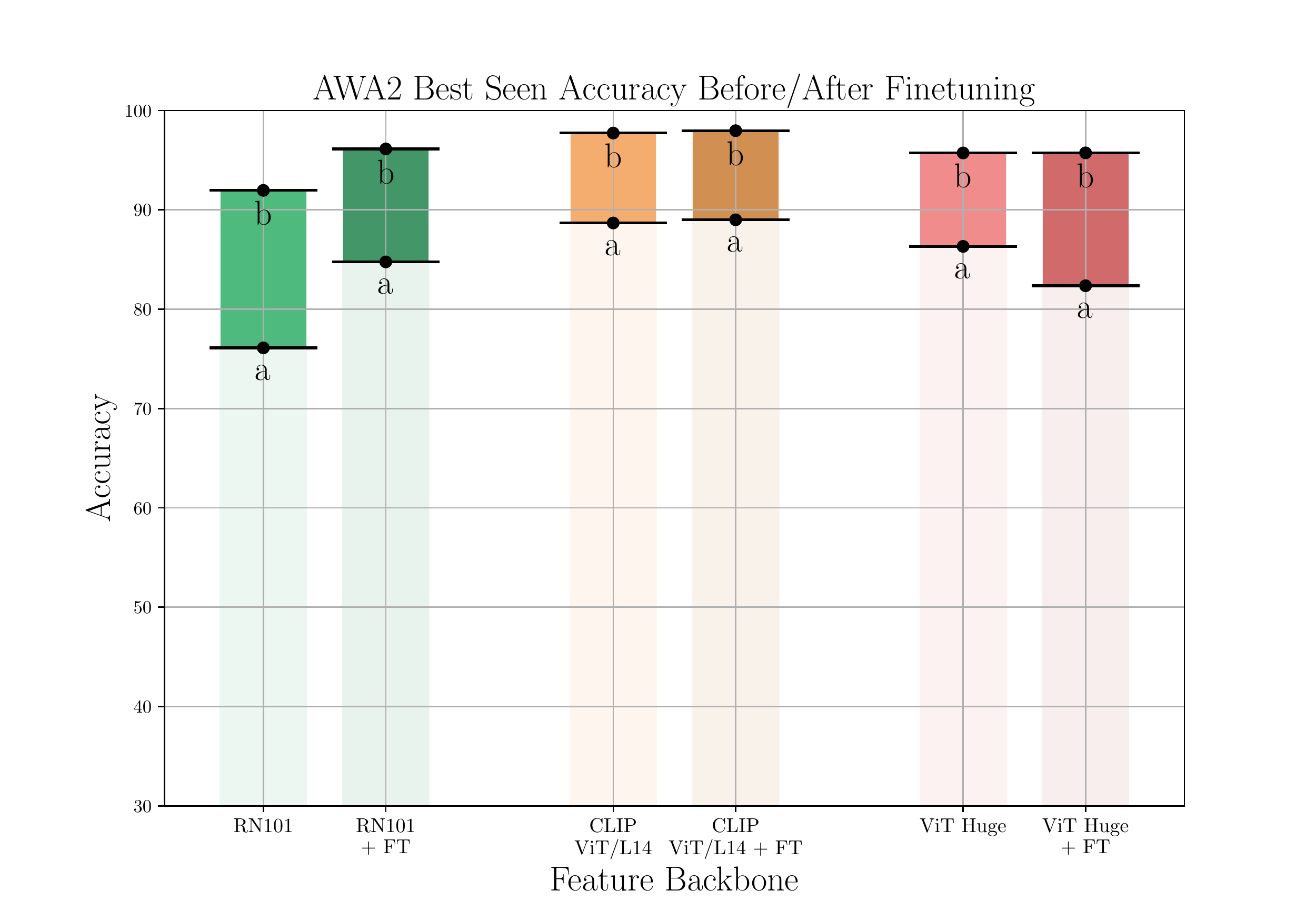}
}
\vspace{-0.25in}
\caption{\textbf{Fine-tuning diverse feature extractors with seen samples}. We show the difference between: a) the best-seen accuracy achieved among all GZSL methods, and b) the best-seen accuracy after training and evaluating the seen classes using a linear classifier probe. We use different visual features extracted from different backbones and training strategies. The classification upper-bound only increases significantly when fine-tuning the RN101\cite{RNs} features but remains the same with the ViT\cite{ViT} features regardless of the classification objective (i.e., cross-entropy vs. contrastive). }.
\label{fig:seen_acc_difference}
\vspace{-0.1in}
\end{figure*}


\textbf{Evaluation of the GZSL methods using the visual features extracted from the visual encoder of CLIP}
to train all GZSL methods: We show in Table~\ref{tab:gzsl_using_clip_features} the effect of using CLIP features and which fine-tuned model performs the best. Unsurprisingly, using both class name and class attributes outperforms other backbones. Similarly to uni-modal backbones, the generative-based models (e.g., CADA-VAE and tfVAEGAN) outperform other methods on all datasets. More surprisingly, both methods also outperform the CLIP results in CUB and SUN, but CLIP alone outperforms all methods in the AWA2 dataset. 
We run experiments using all eight visual encoders available from CLIP model, but show the most relevant results in Table~\ref{tab:gzsl_using_clip_features}. Please refer to the Appendix to check the full list of results.

\section{Fine-tuning Feature Extractors}

We observe that in the generative and disentanglement-based methods, there seems to be a trade-off in the multi-modal latent space, where some features from the seen set are distilled to the projected features of the unseen set. Typically, these methods augment the training set and convert the problem into a classification task, thus, the final seen accuracy is penalized by the classifier. 
For this reason, we also investigate how much these models penalize the final accuracy of the seen sets by measuring the difference of a classifier trained only using the seen set versus the best-seen accuracy achieved among all GZSL generative and disentanglement-based methods. 
The results are shown in Figure~\ref{fig:seen_acc_difference}. 
We observe that while fine-tuning a Resnet101 model increment the seen accuracy of both the classifier and the GZSL method, it does not increase the seen accuracy of recent models such as vision transformers. 
We also observe that the seen accuracy of the GZSL method does not improve substantially when using ViT features, 
and fine-tuning on AWA2 hurts the best seen accuracy on the GZSL result while not improving the classifier seen accuracy.

\section{Conclusion}

In this paper, we provide strong empirical evidence that indicates that:

\begin{compactitem}
\item Using Transformer based architectures provides superior feature representation capabilities while not violating the zero-shot principle of being pre-trained on unseen classes.

\item The feature representations extracted from unimodal architectures that were pre-trained on larger datasets (e.g., ImageNet-21k) do not necessarily boost GZSL performance.

\item Using Convolutional based architectures pre-trained without labels, using contrastive learning and self-distillation, provides better feature representations for GZSL than models trained using supervised learning, with known labels and cross-entropy loss alone.

\item Fine-tuning does not significantly impact the performance on Transformer based unimodal backbones but may boost the performence on multimodal backbones.

\item Multimodal architectures trained on internet-scale large data (CLIP) still benefit from generative based GZSL methods to achieve state-of-the art performance in CUB and SUN, which are fine-grained datasets. This may indicate that feature representations from CLIP are more suitable for GZSL when there is less inter-class correlation among data samples.

\item Fine-tuning a CLIP model using prompts including the class names and attributes 
from the seen categories also boosts the ranking performance of the unseen classes.
\end{compactitem}

In summary, our work provides an update on GZSL methods in the era of large-scale multi-modal pre-training, and re-evaluates in this context the progress that has been made so far in this area.
We release a well-documented codebase that both replicates our findings and provides a modular framework for further feature representation explorations to the GZSL task with recent large pre-trained models.

{\setlength{\parindent}{0cm}
\textbf{Acknowledgements} This material is based upon work supported by the National Science Foundation under Grant No. 2221943 and Grant No. 2201710.
}

{\small
\bibliographystyle{ieee_fullname}
\bibliography{egbib}
}


\clearpage

\clearpage
\appendix

\section{Appendix}

In Section~\ref{sec:implementation_details} we show implementation details with hyperparameter selections for all GZSL methods, and detail the computing infrastructure we use while conducting all of our experiments. 
Then we show in fine-tuning information for the multi-modal backbone in Section~\ref{sec:prompt_engineering_text}.
We also show extended results for all of our experiments using all features from different backbones and methods in Section~\ref{sec:all_tables}.
Lastly, we discuss about some ethical considerations and the importance of Generalized Zero-Shot Learning in Section~\ref{ethical}.

\section{Implementation Details}
\label{sec:implementation_details}

We follow the original code and recommended hyperparameters from the existing implementations provided by their corresponding authors for all the GZSL methods in this study. In Table~\ref{tab:hyperparamters_gzslmethods} we detail all values for all methods.

\subsection{Computing Infrastructure}

We performed all the GZSL methods experiments using features extracted from unimodal backbones in 3 servers with 4 NVIDIA TITAN RTX GPUs each. 
We performed all the GZSL methods experiments using features extracted from multimodal backbones in a single server with 8 NVIDIA A40 GPUs. 
Feature extraction and fine-tuning were done in a single server with 8 NVIDIA A40 GPUs.
All experiments were run on a single GPU.


\begin{table*}[!h]
\vspace{-0.05in}
\setlength{\tabcolsep}{5pt}
\begin{center}
\scalebox{0.6}{
\begin{tabular}{?l|l|l?l|l|l?l|l|l?}
\multicolumn{3}{c}{DEVISE\cite{DeViSE}} & \multicolumn{3}{c}{ESZSL\cite{ESZSL}} & \multicolumn{3}{c}{ALE\cite{ALE}} \\
\hline
CUB     & SUN        &   AWA2   &   CUB   & SUN   & AWA2  & CUB   & SUN   & AWA2 \\
\hline
norm = L2 & norm = None  & norm = STD & $\alpha = 3$ & $\alpha = 3$ & $\alpha = 3$ & norm = L2 & norm = L2 & norm = L2 \\
lr = 1.0  & lr = 0.01    & lr = 0.001 & $\gamma = 0$ & $\gamma = 2$ & $\gamma = 0$ & lr = 0.3 & lr = 0.1 & lr = 0.01 \\
mr = 1.0  &  mr = 3.0    & mr = 150   & & & & & & \\
\midrule
\multicolumn{3}{c}{tfVAEGAN\cite{tfvaegan}} & \multicolumn{3}{c}{CADA-VAE\cite{CADA_VAE}} & \multicolumn{3}{c}{CE\cite{CE}} \\
\hline
CUB     & SUN        &   AWA2   &                                   CUB   & SUN   & AWA2  & CUB   & SUN   & AWA2 \\
\hline
$\gamma_{D} = 10$ &	$\gamma_{D} = 1$&	$\gamma_{D} = 10$& \multicolumn{3}{c?}{lr = 0.00015}  &syn num = 100 &	syn num = 100 &	syn num = 1800 \\
$\gamma_{G} = 10$&	$\gamma_{G} = 1$&	$\gamma_{G} = 10$& \multicolumn{3}{c?}{cls lr = 0.001} &batch size = 2048 &	batch size = 2048 &	batch size = 4096 \\
nepoch = 300 &	nepoch = 401& 	nepoch = 300            & \multicolumn{3}{c?}{cls loss = L1} &attSize = 312 &	attSize = 312 &	attSize = 312 \\
ngh = 4096 &	ngh = 4096 &	ngh = 4096              & \multicolumn{3}{c?}{$\beta$ factor = 0.25} &nz = 1024 &	nz = 1024 &	nz = 1024 \\
ndh = 4096 &	ndh = 4096 &	ndh = 4096              & \multicolumn{3}{c?}{cross reconst factor = 2.37} &embedSize = 2048 &	embedSize = 2048 &	embedSize = 2048 \\
lr = 0.0001 &	lr = 0.001 &	lr = 0.00001            & \multicolumn{3}{c?}{distance factor = 8.13} &outzSize = 512 &	outzSize = 512 &	outzSize = 512 \\
classifier lr = 0.001 &	classifier lr = 0.0005&	classifier lr = 0.001&  \multicolumn{3}{c?}{nepoch = 200} &nhF = 2048 &	nhF = 2048& 	nhF = 2048 \\
$\lambda = 10$ &	$\lambda = 10$ &	$\lambda = 10$ & \multicolumn{3}{c?}{batch size = 100} &ins weight = 0.001 &	ins weight = 0.001 &	ins weight = 0.001 \\
critic iter = 5&	critic iter = 5&	critic ite =r 5& \multicolumn{3}{c?}{lattent size = 64} &cls weight = 0.001 &	cls weight = 0.001 &	cls weight = 0.001 \\
batch size = 64 &	batch size = 64 &	batch size = 64 & \multicolumn{3}{c?}{~} &ins temp = 0.1 &	ins temp = 0.1 	&ins temp = 10.0 \\
nz = 312& 	nz = 102 &	nz = 85 & \multicolumn{3}{c?}{~} &cls temp = 0.1 &	cls temp = 0.1 &	cls temp = 1.0 \\
latent size = 312 &	latent size = 102	&latent size = 85 & \multicolumn{3}{c?}{~} &nepoch = 401&	nepoch = 401&	nepoch = 131 \\
syn num = 300 &	syn num = 400 &	syn num = 1800 & \multicolumn{3}{c?}{~} &manualSeed = 3483 &	manualSeed = 4115 &	manualSeed = 9182 \\
recons weight = 0.01 &	recons weight = 0.01 	&recons weight = 0.1 & \multicolumn{3}{c?}{~} &&& \\
feed lr = 0.00001 &	feed lr = 0.0001&	feed lr = 0.0001 & \multicolumn{3}{c?}{~} &&& \\
dec lr = 0.0001 &	dec lr = 0.0001 &	dec lr = 0.0001 & \multicolumn{3}{c?}{~} &&& \\
feedback loop = 2&	feedback loop = 2	&feedback loop = 2& \multicolumn{3}{c?}{~} &&& \\
manualSeed = 3483&	manualSeed = 4115&	manualSeed = 9182& \multicolumn{3}{c?}{~} &&& \\

\midrule
\multicolumn{3}{c}{SDGZSL\cite{SDGZSL}} & \multicolumn{6}{c}{FREE\cite{Chen2021FREE}}  \\
\hline
CUB     & SUN        &   AWA2   &   \multicolumn{2}{c|}{CUB}   & \multicolumn{2}{c|}{SUN}   & \multicolumn{2}{c?}{AWA2}  \\
\hline
$\gamma = 5$ 	&	$\gamma = 30 $	&	$\gamma = 0.5 $	&	\multicolumn{2}{l|}{	nepoch = 201 	}	&	\multicolumn{2}{l|}{	nepoch = 601 	}	&	\multicolumn{2}{l?}{	nepoch = 401 	}	 \\
$\beta = 0.003$	&	$\beta = 0.3 $	&	$\beta = 1 $	&	\multicolumn{2}{l|}{	ngh = 4096 	}	&	\multicolumn{2}{l|}{	ngh = 4096 	}	&	\multicolumn{2}{l?}{	ngh = 4096 	}	 \\
dis = 0.3 	&	dis = 0.5 	&	dis = 0.3 	&	\multicolumn{2}{l|}{	lr = 0.0001 	}	&	\multicolumn{2}{l|}{	loop = 2 	}	&	\multicolumn{2}{l?}{	$\lambda = 10$ 	}	 \\
nSample = 1000 	&	nSample = 400 	&	nSample = 5000 	&	\multicolumn{2}{l|}{	classifier lr = 0.001 	}	&	\multicolumn{2}{l|}{	feed lr = 0.0001 	}	&	\multicolumn{2}{l?}{	critic iter = 1 	}	 \\
lr = 0.0001 	&	lr = 0.0003 	&	lr = 0.00003 	&	\multicolumn{2}{l|}{	$\lambda = 10$	}	&	\multicolumn{2}{l|}{	ndh = 4096 	}	&	\multicolumn{2}{l?}{	feed lr = 0.0001 	}	 \\
classifier lr = 0.002 	&	kl warmup = 0.001 	&	classifier lr = 0.003 	&	\multicolumn{2}{l|}{	critic iter = 1 	}	&	\multicolumn{2}{l|}{	$\lambda = 10$ 	}	&	\multicolumn{2}{l?}{	dec lr = 0.0001 	}	 \\
nepoch = 600 	&	tc warmup = 0.0003 	&	kl warmup = 0.01 	&	\multicolumn{2}{l|}{	feed lr = 0.00001 	}	&	\multicolumn{2}{l|}{	critic iter = 1 	}	&	\multicolumn{2}{l?}{	loop = 2 	}	 \\
kl warmup = 0.001 	&	vae dec drop = 0.2 	&	tc warmup = 0.001 	&	\multicolumn{2}{l|}{	dec lr = 0.0001 	}	&	\multicolumn{2}{l|}{	batch size = 512 	}	&	\multicolumn{2}{l?}{	batch size = 64 	}	 \\
tc warmup = 0.0001 	&	dis step = 3 	&	vae dec drop = 0.5 	&	\multicolumn{2}{l|}{	loop = 2 	}	&	\multicolumn{2}{l|}{	nz = 102 	}	&	\multicolumn{2}{l?}{	nz = 85 	}	 \\
weight decay = 1e-8 	&	ae drop = 0.4	&	vae enc drop = 0.4 	&	\multicolumn{2}{l|}{	batch size = 64 	}	&	\multicolumn{2}{l|}{	latent size = 102 	}	&	\multicolumn{2}{l?}{	latent size = 85 	}	 \\
vae enc drop = 0.1 	&		&	dis step = 2  	&	\multicolumn{2}{l|}{	nz = 312 	}	&	\multicolumn{2}{l|}{	lr = 0.0002 	}	&	\multicolumn{2}{l?}{	lr = 0.00001 	}	 \\
vae dec drop = 0.1 	&		&	ae drop = 0.2 	&	\multicolumn{2}{l|}{	latent size = 312 	}	&	\multicolumn{2}{l|}{	classifier lr = 0.0005 	}	&	\multicolumn{2}{l?}{	classifier lr = 0.001 	}	 \\
dis step = 3 	&		&	gen nepoch = 220 	&	\multicolumn{2}{l|}{	syn num = 700 	}	&	\multicolumn{2}{l|}{	syn num = 300 	}	&	\multicolumn{2}{l?}{	syn num = 4600 	}	 \\
ae drop = 0.0	&		&	evl start = 40000 	&	\multicolumn{2}{l|}{	center margin = 200 	}	&	\multicolumn{2}{l|}{	center margin = 120 	}	&	\multicolumn{2}{l?}{	center margin = 50 	}	 \\
	&		&	evl interval = 400 	&	\multicolumn{2}{l|}{	center weight = 0.5 	}	&	\multicolumn{2}{l|}{	incenter weight = 0.8 	}	&	\multicolumn{2}{l?}{	center weight = 0.5 	}	 \\
	&		&	manualSeed = 6152	&	\multicolumn{2}{l|}{	recons weight = 0.001 	}	&	\multicolumn{2}{l|}{	center weight = 0.5 	}	&	\multicolumn{2}{l?}{	recons weight = 0.001 	}	 \\
	&		&		&	\multicolumn{2}{l|}{	incenter weight = 0.8	}	&	\multicolumn{2}{l|}{	recons weight = 0.1	}	&	\multicolumn{2}{l?}{	incenter weight = 0.5	}	 \\
	&		&		&	\multicolumn{2}{l|}{	manualSeed = 3483 	}	&	\multicolumn{2}{l|}{	manualSeed = 4115	}	&	\multicolumn{2}{l?}{	manualSeed = 9182	}	 \\
\bottomrule

\end{tabular}
}
\end{center}
\caption{Hyper-parameter selection details for all methods.} 
\label{tab:hyperparamters_gzslmethods}
\end{table*}

\section{Prompt Engineering for CLIP Fine-Tuning}
\label{sec:prompt_engineering_text}

Prompt for all classes adding class name in the sentence: 'This is a photo of a $\{\}$', shown in Table~\ref{tab:dataset_class_names}.
\begin{table*}[!h]
\vspace{-0.05in}
\setlength{\tabcolsep}{5pt}
\begin{center}
\scalebox{0.7}{
\begin{tabular}{?lllll?}
\multicolumn{5}{c}{CUB\cite{CUB}} \\
\hline
Black footed Albatross & 
Laysan Albatross & 
Sooty Albatross & 
Groove billed Ani & 
Crested Auklet  \\
Least Auklet & 
Parakeet Auklet & 
Rhinoceros Auklet & 
Brewer Blackbird & 
Red winged Blackbird  \\
Rusty Blackbird & 
Yellow headed Blackbird & 
Bobolink & 
Indigo Bunting & 
Lazuli Bunting  \\
Painted Bunting & 
Cardinal & 
Spotted Catbird & 
Gray Catbird & 
Yellow breasted Chat \\
Eastern Towhee & 
Chuck will Widow & 
Brandt Cormorant & 
Red faced Cormorant & 
Pelagic Cormorant  \\
Bronzed Cowbird & 
Shiny Cowbird & 
Brown Creeper & 
American Crow & 
Fish Crow  \\
Black billed Cuckoo & 
Mangrove Cuckoo & 
Yellow billed Cuckoo & 
Gray crowned Rosy Finch & 
Purple Finch  \\
Northern Flicker & 
Acadian Flycatcher & 
Great Crested Flycatcher & 
Least Flycatcher & 
Olive sided Flycatcher \\
Scissor tailed Flycatcher & 
Vermilion Flycatcher & 
Yellow bellied Flycatcher & 
Frigatebird & 
Northern Fulmar  \\
Gadwall & 
American Goldfinch & 
European Goldfinch & 
Boat tailed Grackle & 
Eared Grebe  \\ 
Horned Grebe & 
Pied billed Grebe & 
Western Grebe & 
Blue Grosbeak & 
Evening Grosbeak  \\
Pine Grosbeak & 
Rose breasted Grosbeak & 
Pigeon Guillemot & 
California Gull & 
Glaucous winged Gull \\
Heermann Gull & 
Herring Gull & 
Ivory Gull & 
Ring billed Gull & 
Slaty backed Gull  \\
Western Gull & 
Anna Hummingbird & 
Ruby throated Hummingbird & 
Rufous Hummingbird & 
Green Violetear  \\
Long tailed Jaeger & 
Pomarine Jaeger & 
Blue Jay & 
Florida Jay & 
Green Jay  \\
Dark eyed Junco & 
Tropical Kingbird & 
Gray Kingbird & 
Belted Kingfisher & 
Green Kingfisher \\ 
Pied Kingfisher & 
Ringed Kingfisher & 
White breasted Kingfisher & 
Red legged Kittiwake & 
Horned Lark  \\
Pacific Loon & 
Mallard & 
Western Meadowlark & 
Hooded Merganser & 
Red breasted Merganser  \\
Mockingbird & 
Nighthawk & 
Clark Nutcracker & 
White breasted Nuthatch & 
Baltimore Oriole  \\
Hooded Oriole & 
Orchard Oriole & 
Scott Oriole & 
Ovenbird & 
Brown Pelican \\ 
White Pelican & 
Western Wood Pewee & 
Sayornis & 
American Pipit & 
Whip poor Will \\
Horned Puffin & 
Common Raven & 
White necked Raven & 
American Redstart & 
Geococcyx  \\
Loggerhead Shrike & 
Great Grey Shrike & 
Baird Sparrow & 
Black throated Sparrow & 
Brewer Sparrow  \\
Chipping Sparrow & 
Clay colored Sparrow & 
House Sparrow & 
Field Sparrow & 
Fox Sparrow \\
Grasshopper Sparrow & 
Harris Sparrow & 
Henslow Sparrow & 
Le Conte Sparrow & 
Lincoln Sparrow  \\
Nelson Sharp tailed Sparrow & 
Savannah Sparrow & 
Seaside Sparrow & 
Song Sparrow & 
Tree Sparrow  \\
Vesper Sparrow & 
White crowned Sparrow & 
White throated Sparrow & 
Cape Glossy Starling & 
Bank Swallow  \\
Barn Swallow & 
Cliff Swallow & 
Tree Swallow & 
Scarlet Tanager & 
Summer Tanager \\ 
Artic Tern & 
Black Tern & 
Caspian Tern & 
Common Tern & 
... \\
\midrule
\multicolumn{5}{c}{SUN\cite{SUN}} \\
\hline
abbey &
access road &
airfield &
airlock &
airplane cabin \\
airport airport &
airport entrance &
airport terminal &
airport ticket counter &
alcove \\
alley &
amphitheater &
amusement arcade &
amusement park &
anechoic chamber \\
apartment building outdoor &
apse indoor &
apse outdoor &
aquarium &
aquatic theater \\
aqueduct &
arch &
archaelogical excavation &
archive &
arena basketball \\
arena hockey &
arena performance &
armory &
arrival gate outdoor &
art gallery \\
art school &
art studio &
artists loft &
assembly line &
athletic field outdoor \\
atrium home &
atrium public &
attic &
auditorium &
auto factory \\
auto mechanics indoor &
auto racing paddock &
auto showroom &
backstage &
badlands \\
badminton court indoor &
badminton court outdoor &
baggage claim &
bakery kitchen &
bakery shop \\
balcony exterior &
balcony interior &
ball pit &
ballroom &
bamboo forest \\
bank indoor &
bank outdoor &
bank vault &
banquet hall &
baptistry indoor \\
baptistry outdoor &
bar &
barn &
barndoor &
baseball field \\
basement &
basilica &
basketball court indoor &
basketball court outdoor &
bathroom \\
batters box &
batting cage indoor &
batting cage outdoor &
bayou &
bazaar indoor \\
bazaar outdoor &
beach &
beach house &
beauty salon &
bedchamber \\
bedroom &
beer garden &
beer hall &
bell foundry &
berth \\
betting shop &
bicycle racks &
bindery &
biology laboratory &
bistro indoor \\
bistro outdoor &
bleachers outdoor &
boardwalk &
boat deck &
boathouse \\
bog &
bookstore &
booth indoor &
botanical garden &
bow window indoor \\
bow window outdoor &
bowling alley &
boxing ring &
brewery indoor &
brewery outdoor \\
brickyard outdoor &
bridge &
building complex &
building facade &
bullpen \\
bullring &
burial chamber &
bus depot outdoor &
bus interior &
bus shelter \\
bus station outdoor &
butchers shop &
butte &
cabana &
cabin outdoor \\
cafeteria &
call center &
campsite &
campus &
canal natural \\
canal urban &
candy store &
canteen &
canyon &
car interior backseat \\
car interior frontseat &
caravansary &
cardroom &
cargo deck airplane &
carport freestanding \\
carport outdoor &
carrousel &
casino indoor &
casino outdoor &
castle \\
catacomb &
cathedral indoor &
cathedral outdoor &
catwalk &
cavern indoor \\
cemetery &
chalet &
chaparral &
chapel &
checkout counter \\
cheese factory &
chemical plant &
chemistry lab &
chicken coop indoor &
... \\
\midrule
\multicolumn{5}{c}{AWA2\cite{AWA2}} \\
\hline
antelope &
grizzly bear &
killer whale &
beaver &
dalmatian \\
persian cat &
horse &
german shepherd &
blue whale &
siamese cat \\
skunk &
mole &
tiger &
hippopotamus &
leopard \\
moose &
spider monkey &
humpback whale &
elephant &
gorilla \\
ox &
fox &
sheep &
seal &
chimpanzee \\
hamster &
squirrel &
rhinoceros &
rabbit &
bat \\
giraffe &
wolf &
chihuahua &
rat &
weasel \\
otter &
buffalo &
zebra &
giant panda &
deer \\
bobcat &
pig &
lion &
mouse &
polar bear \\
collie &
walrus &
raccoon &
cow &
dolphin \\
\bottomrule
\end{tabular}
}
\end{center}
\caption{Some class names per dataset.} 
\label{tab:dataset_class_names}
\end{table*}

We also finetune taking into account the attribute \textbf{labels} and scores per dataset, shown in Table~\ref{tab:dataset_attribute_names}.

\begin{table*}[!h]
\vspace{-0.05in}
\setlength{\tabcolsep}{5pt}
\begin{center}
\scalebox{0.6}{
\begin{tabular}{?lllll?}
\multicolumn{5}{c}{CUB\cite{CUB}} \\
\hline
\multicolumn{5}{?l?}{Prompt: 'Image of a bird with $\{\}$'} \\
\hline
curved (up or down) bill shape &
dagger bill shape &
hooked bill shape &
needle bill shape &
hooked seabird bill shape \\
spatulate bill shape &
all-purpose bill shape &
cone bill shape &
specialized bill shape &
blue wings \\
brown wings &
iridescent wings &
purple wings &
rufous wings &
grey wings \\
yellow wings &
olive wings &
green wings &
pink wings &
orange wings \\
black wings &
white wings &
red wings &
buff wings &
blue upperparts \\
brown upperparts &
iridescent upperparts &
purple upperparts &
rufous upperparts &
grey upperparts \\
yellow upperparts &
olive upperparts &
green upperparts &
pink upperparts &
orange upperparts \\
black upperparts &
white upperparts &
red upperparts &
buff upperparts &
blue underparts \\
brown underparts &
iridescent underparts &
purple underparts &
rufous underparts &
grey underparts \\
yellow underparts &
olive underparts &
green underparts &
pink underparts &
orange underparts \\
black underparts &
white underparts &
red underparts &
buff underparts &
solid breast pattern \\
spotted breast pattern &
striped breast pattern &
multi-ed breast pattern &
blue back &
brown back \\
iridescent back &
purple back &
rufous back &
grey back &
yellow back \\
olive back &
green back &
pink back &
orange back &
black back \\
white back &
red back &
buff back &
forked tail tail shape &
rounded tail tail shape \\
notched tail tail shape &
fan-shaped tail tail shape &
pointed tail tail shape &
squared tail tail shape &
blue upper tail \\
brown upper tail &
iridescent upper tail &
purple upper tail &
rufous upper tail &
grey upper tail \\
yellow upper tail &
olive upper tail &
green upper tail &
pink upper tail &
orange upper tail \\
black upper tail &
white upper tail &
red upper tail &
buff upper tail &
spotted head pattern \\
malar head pattern &
crested head pattern &
masked head pattern &
unique pattern head pattern &
eyebrow head pattern \\
eyering head pattern &
plain head pattern &
eyeline head pattern &
striped head pattern &
capped head pattern \\
blue breast &
brown breast &
iridescent breast &
purple breast &
rufous breast \\
grey breast &
yellow breast &
olive breast &
green breast &
pink breast \\
orange breast &
black breast &
white breast &
red breast &
buff breast \\
blue throat &
brown throat &
iridescent throat &
purple throat &
rufous throat \\
grey throat &
yellow throat &
olive throat &
green throat &
pink throat \\
orange throat &
black throat &
white throat &
red throat &
buff throat \\
blue eye &
brown eye &
purple eye &
rufous eye &
grey eye \\
yellow eye &
olive eye &
green eye &
pink eye &
orange eye \\
black eye &
white eye &
red eye &
buff eye &
broad-wings wing shape \\
tapered-wings wing shape &
long-wings wing shape &

blue forehead &
brown forehead &
iridescent forehead \\
purple forehead &
rufous forehead &
grey forehead &
yellow forehead &
olive forehead \\
green forehead &
pink forehead &
orange forehead &
black forehead &
white forehead \\
red forehead &
buff forehead &
blue under tail &
brown under tail &
iridescent under tail \\
purple under tail &
rufous under tail &
grey under tail &
yellow under tail &
olive under tail \\
green under tail &
pink under tail &
orange under tail &
black under tail &
white under tail \\
red under tail &
buff under tail &
blue nape &
brown nape &
iridescent nape \\
purple nape &
rufous nape &
grey nape &
yellow nape &
olive nape \\
green nape &
pink nape &
orange nape &
black nape &
white nape \\
red nape &
buff nape &
blue belly &
brown belly &
iridescent belly \\
purple belly &
rufous belly &
grey belly &
yellow belly &
olive belly \\
green belly &
pink belly &
orange belly &
black belly &
white belly \\
red belly &
buff belly &
rounded-wings wing shape &
pointed-wings wing shape &
broad-wings wing shape \\
iridescent primary &
purple primary &
rufous primary &
grey primary &
yellow primary \\
olive primary &
green primary &
pink primary &
orange primary &
black primary \\
white primary &
red primary &
buff primary &
blue leg &
brown leg \\
iridescent leg &
purple leg &
rufous leg &
grey leg &
yellow leg \\
olive leg &
green leg &
pink leg &
orange leg &
black leg \\
white leg &
red leg &
buff leg &
blue bill &
... \\
\midrule
\multicolumn{5}{c}{SUN\cite{SUN}} \\
\hline
\multicolumn{5}{?l?}{Prompt: 'Image of a place for $\{\}$'} \\
\multicolumn{5}{?l?}{Prompt: 'Image of a place with $\{\}$'} \\
\hline
sailing or boating &
driving &
biking &
transporting things or people &
sunbathing \\
vacationing or touring &
hiking &
climbing &
camping &
reading \\
studying or learning &
teaching or training &
research &
diving &
swimming \\
bathing &
eating &
cleaning &
socializing &
congregating \\
waiting in line or queuing &
competing &
sports &
exercise &
playing \\
gaming &
spectating or being in an audience &
farming &
constructing or building &
shopping \\
medical activity &
working &
using tools &
digging &
conducting business \\
praying &
fencing &
railing &
wire &
railroad \\
trees &
grass &
vegetation &
shrubbery &
foliage \\
leaves &
flowers &
asphalt &
pavement &
shingles \\
carpet &
brick &
tiles &
concrete &
metal \\
paper &
wood (not part of a tree) &
vinyl or linoleum &
rubber or plastic &
cloth \\
sand &
rock orstone &
dirt orsoil &
marble &
glass \\
waves or surf &
ocean &
running water &
still water &
ice \\
snow &
clouds &
smoke &
fire &
... \\
\midrule
\multicolumn{5}{c}{AWA2\cite{AWA2}} \\
\hline
\multicolumn{5}{?l?}{Prompt: 'Image of an animal with $\{\}$'} \\
\multicolumn{5}{?l?}{Prompt: 'Image of an animal that $\{\}$'} \\
\multicolumn{5}{?l?}{Prompt: 'Image of an animal that lives in the $\{\}$'} \\
Prompt: 'Image of a $\{\}$ animal' & & & & \\
\hline
black &
white &
blue &
brown &
gray \\
orange &
red &
yellow &
patches &
spots \\
stripes &
furry &
hairless &
toughskin &
big \\
small &
bulbous &
lean &
flippers &
hands \\
hooves &
pads &
paws &
longleg &
longneck \\
tail &
chewteeth &
meatteeth &
buckteeth &
strainteeth \\
horns &
claws &
tusks &
smelly &
flys \\
hops &
swims &
tunnels &
walks &
fast \\
slow &
strong &
weak &
muscle &
bipedal \\
quadrapedal &
active &
inactive &
nocturnal &
hibernate \\
agility &
fish &
meat &
plankton &
vegetation \\
insects &
forager &
grazer &
hunter &
scavenger \\
skimmer &
stalker &
newworld &
oldworld &
arctic \\
coastal &
desert &
bush &
plains &
forest \\
fields &
jungle &
mountains &
ocean &
ground \\
water &
tree &
cave &
fierce &
timid \\
smart &
group &
solitary &
nestspot &
domestic \\
\bottomrule
\end{tabular}
}
\end{center}
\caption{Prompt and some class attributes labels per dataset.} 
\label{tab:dataset_attribute_names}
\end{table*}
\section{Full experimental results for our large scale analysis}
\label{sec:all_tables}

In this section we show all the results obtained using a large variety of visual backbones from different architecture types. We showcase the performance of all methods, grouped by their corresponding GZSL families and datasets as follows: 

\vfill\eject
\begin{compactitem}
  \item Embedding-based methods \\DEVISE~\cite{DeViSE}, ESZSL~\cite{ESZSL} and ALE\cite{ALE}:
  \begin{itemize}
    \item CUB results in Table \ref{tab:cub_embedding_CNN}. 
    \item SUN results in table \ref{tab:sun_embedding_CNN}.
    \item AWA2 results in table \ref{tab:awa2_embedding_CNN}. 
  \end{itemize}
  \item Generative-based methods \\TF-VAEGAN~\cite{tfvaegan}, CADA-VAE~\cite{CADA_VAE}, CE~\cite{CE}:
    \begin{itemize}
    \item CUB results in Table \ref{tab:cub_generative_CNN}. 
    \item SUN results in table \ref{tab:sun_generative_CNN}. 
    \item AWA2 results in table \ref{tab:awa2_generative_CNN}. 
  \end{itemize}
  \item Semantic disentanglement-based methods \\SDGZSL~\cite{SDGZSL}, FREE~\cite{Chen2021FREE}:
    \begin{itemize}
        \item CUB results in Table \ref{tab:cub_disentanglement_CNN}. 
        \item SUN results in table \ref{tab:sun_disentanglement_CNN}. 
        \item AWA2 results in table \ref{tab:awa2_disentanglement_CNN}. 
  \end{itemize}
\end{compactitem}

\begin{table*}[!htbp]
\newcolumntype{Y}{>{\raggedright\arraybackslash}X}
\newcolumntype{Z}{>{\centering\arraybackslash}X}
\centering
\footnotesize
\setlength\tabcolsep{1pt}
\renewcommand{\arraystretch}{1.2}

\begin{tabularx}{\textwidth}{l c l c l c YYY c YYY c YYY}
\toprule

\multicolumn{17}{@{\hskip 0.11in}c}{\bf \shortstack{Embedding Based GZSL Methods\\ CUB\cite{CUB} Dataset}}  \\
\midrule

{\multirow{2}{*}{\bf \shortstack{Dataset\\Pret. on}}}~ &~~~&
{\multirow{2}{*}{\bf \shortstack{Arch\\ Type}}}~ &~~&
{\multirow{2}{*}{\bf Backbone}}~ &~~&
\multicolumn{3}{@{\hskip 0.11in}c}{\bf DEVISE} &~~~~& 
\multicolumn{3}{@{\hskip 0.11in}c}{\bf ESZSL} &~~~~& 
\multicolumn{3}{@{\hskip 0.11in}c}{\bf ALE} \\

\cmidrule{7-9}\cmidrule{11-13}\cmidrule{15-17}

&& && && \textit{Seen} & \textit{Novel} & \textit{Harm.} 
&& \textit{Seen} & \textit{Novel} & \textit{Harm.} 
&& \textit{Seen} & \textit{Novel} & \textit{Harm.} \\

\midrule

\multirow{17}{*}{I-1k} & &
\multirow{12}{*}{CNN} & &
RN101 &  & 
61.96 & 23.41 & 33.98 &&
56.53 & 14.70 & 23.34 &&
62.74 & 26.07 & 36.83 \\ 

&& && RN101+FT &  & 
83.42 & 28.32 & 42.28 &&
81.45 & 16.14 & 26.94 &&
83.04 & 30.36 & 44.46 \\ 

&& && RN50 &  & 
41.32 & 17.31 & 24.39 &&
41.83 & 13.78 & 20.73 &&
45.96 & 24.70 & 32.13 \\ 

&& && RN152 &  &  
45.73 & 19.01 & 26.86 &&
46.39 & 15.49 & 23.23 &&
52.69 & 24.84 & 33.76  \\ 

&& && GoogleNet &  &  
35.57 & 15.10 & 21.20 && 
30.79 & 9.70 & 14.75 && 
36.86 & 17.67 & 23.89  \\ 

&& && VGG16 &  &  
44.11 & 14.93 & 22.30 && 
46.74 & 7.29 & 12.61 && 
45.81 & 18.26 & 26.11  \\ 

&& && Alexnet &  &  
33.57 & 13.04 & 18.78 && 
33.65 & 5.62 & 9.64 && 
33.15 & 14.82 & 20.48  \\ 

&& && Shufflenet &  &  
47.24 & 20.05 & 28.15 && 
22.16 & 12.13 & 15.67 && 
48.15 & 22.18 & 30.37  \\ 

&& && Inceptionv3 &  &  
58.09 & 21.81 & 31.71 && 
48.30 & 12.91 & 20.38 && 
53.11 & 21.52 & 30.63  \\ 

&& && Inceptionv3$_{\text{adv}}$ &  &  
54.84 & 20.86 & 30.22 && 
47.70 & 11.09 & 18.00 && 
59.37 & 20.24 & 30.18  \\ 

\cmidrule{5-17}
&& && RN50-MOCO$^{\dag}$ &  &  
31.26 & 11.39 & 16.70 && 
7.69 & 3.87 & 5.15 && 
32.73 & 15.77 & 21.28  \\ 

&& && RN50-DINO$^{\dag}$ &  &  
63.20 & 25.84 & 36.68 &&
41.65 & 14.64 & 21.67 && 
60.69 & 25.79 & 36.19  \\

\cmidrule{3-17}

&& MLP&&MLP-Mixer && 
19.17 & 7.76 & 11.05 & &
20.50 & 5.53 & 8.71 & &
19.82 & 8.11 & 11.51  \\

\cmidrule{3-17}

&& \multirow{3}{*}{ViT} &&ViT$_{\text{large}}$&&
71.85 & 23.88 & 35.85 & &
\textbf{83.64} & 12.26 & 21.39  & &
72.20 & 23.44 & 35.39  \\

&& &&DeiT$_{\text{base}}$ && 
70.45 & 25.86 & 37.83 & &
60.88 & 16.81 & 26.35  & &
73.88 & 30.88 & 43.55  \\  

&& &&\cellcolor{gray!18}ViTB16-DINO$^{\dag}$&\cellcolor{gray!18}& 
75.50 & \textbf{32.20} & 45.15 &\cellcolor{gray!18} &
\cellcolor{gray!18}70.89 & \cellcolor{gray!18}\textbf{29.06} & \cellcolor{gray!18}\textbf{41.23}  & &
77.21 & \textbf{36.83} & 49.87  \\ 

\midrule

\multirow{5}{*}{I-21k}
&& MLP && 
MLP-Mixer$_{\text{L16}}$ & &
38.46 & 12.02 & 18.32 & &
34.62 & 5.82 & 9.97 & &
41.66 & 10.68 & 17.00  \\

\cmidrule{3-17}

&& \multirow{3}{*}{ViT} && ViT$_{\text{base}}$ & &
84.52 & 26.75 & 40.64 & &
81.33 & 19.57 & 31.55 & &
83.39 & 29.59 & 43.68  \\ 

&& && ViT$_{\text{large}}$ & &
\textbf{85.24} & 25.61 & 39.38 & &
82.47 & 17.53 & 28.92 & &
83.09 & 22.08 & 34.89  \\  

&& && \cellcolor{gray!18}ViT$_{\text{huge}}$ &\cellcolor{gray!18} &
\cellcolor{gray!18}82.78 & \cellcolor{gray!18}31.31 & \cellcolor{gray!18}\textbf{45.44} & &
61.88 & 18.77 & 28.80 &\cellcolor{gray!18} &
\cellcolor{gray!18}\textbf{84.00} & \cellcolor{gray!18}36.05 & \cellcolor{gray!18}\textbf{50.45}  \\ 

\bottomrule
\end{tabularx}
\caption{Results of Embedding Based Methods for the CUB\cite{CUB} dataset using different features extracted from a diverse set of architecture types pretrained on ImageNet-1k (I-1k) and ImageNet-21k (I-21k)\cite{Imagenet}. These backbones were trained via: supervised and self-supervised (${\dag}$) learning. The bold numbers correspond to the highest scores per column, and the shaded rows correspond to the most performant image feature per method. +FT indicates the features were fine-tuned with the seen classes from the training set. The ViT$_{\text{huge}}$ features pretrained on ImageNet-21k are the best for all the methods using ALE.  
}
\label{tab:cub_embedding_CNN}
\end{table*}

\begin{table*}[!htbp]
\newcolumntype{Y}{>{\raggedright\arraybackslash}X}
\newcolumntype{Z}{>{\centering\arraybackslash}X}
\centering
\footnotesize
\setlength\tabcolsep{1pt}
\renewcommand{\arraystretch}{1.2}

\begin{tabularx}{\textwidth}{l c l c l c YYY c YYY c YYY}
\toprule

\multicolumn{17}{@{\hskip 0.11in}c}{\bf \shortstack{Embedding Based GZSL Methods\\ SUN Dataset}}  \\
\midrule

{\multirow{2}{*}{\bf \shortstack{Dataset\\Pret. on}}}~ &~~~~&
{\multirow{2}{*}{\bf \shortstack{Arch\\ Type}}}~ &~~~~&
{\multirow{2}{*}{\bf Backbone}}~ &~~~~&
\multicolumn{3}{@{\hskip 0.11in}c}{\bf DEVISE} &~~~~& 
\multicolumn{3}{@{\hskip 0.11in}c}{\bf ESZSL} &~~~~& 
\multicolumn{3}{@{\hskip 0.11in}c}{\bf ALE} \\

\cmidrule{7-9}\cmidrule{11-13}\cmidrule{15-17}

&& && && \textit{Seen} & \textit{Novel} & \textit{Harm.} 
&& \textit{Seen} & \textit{Novel} & \textit{Harm.} 
&& \textit{Seen} & \textit{Novel} & \textit{Harm.} \\

\midrule

\multirow{17}{*}{I-1k} & &
\multirow{12}{*}{CNN} & &
RN101 & &
32.75 & 18.54 & 23.68 && 
28.41 & 13.75 & 18.53 && 
37.13 & 23.68 & 28.92  \\ 

&& &&RN101+FT &  &
34.03 & 19.93 & 25.14 && 
33.18 & 13.82 & 19.51 && 
37.95 & 22.43 & 28.19  \\ 

&& &&RN50 & &
29.84 & 17.43 & 22.01 && 
25.08 & 14.38 & 18.27 && 
33.91 & 23.54 & 27.79  \\ 

&& &&RN152 & &
30.39 & 17.29 & 22.04 && 
26.63 & 15.90 & 19.91 && 
35.70 & 23.40 & 28.27  \\ 

&& &&GoogleNet & &
18.02 & 10.83 & 13.53 && 
17.52 & 9.31 & 12.15 && 
24.88 & 15.76 & 19.30  \\ 

&& &&VGG16 & &
28.91 & 13.06 & 17.99 && 
25.85 & 9.93 & 14.35 && 
31.78 & 20.21 & 24.71  \\ 

&& &&Alexnet & &
19.61 & 9.31 & 12.62 && 
19.03 & 6.88 & 10.10 && 
23.60 & 13.82 & 17.43  \\ 

&& &&Shufflenet & &
0.23 & 0.00 & 0.00 && 
0.74 & 1.74 & 1.03 && 
29.53 & 17.78 & 22.20  \\ 

&& &&Inceptionv3 & &
31.01 & 14.03 & 19.32 && 
25.08 & 11.25 & 15.53 && 
32.29 & 18.47 & 23.50  \\ 

&& &&Inceptionv3$_{\text{adv}}$ & &
27.91 & 15.00 & 19.51 && 
24.69 & 12.08 & 16.23 && 
33.88 & 21.39 & 26.22  \\ 

\cmidrule{5-17}
&& &&RN50-MOCO$^{\dag}$ & &
2.25 & 0.00 & 0.00 && 
3.14 & 3.33 & 3.23 && 
34.92 & 23.33 & 27.98  \\ 

&& &&RN50-DINO$^{\dag}$ & &
32.56 & 19.03 & 24.02 && 
22.48 & 14.65 & 17.74 && 
41.59 & 27.57 & 33.16  \\ 

\cmidrule{3-17}

&& MLP&&MLP-Mixer && 
6.32 & 3.33 & 4.36 & &
6.40 & 2.36 & 3.45 & &
8.29 & 3.96 & 5.36  \\

\cmidrule{3-17}

&& \multirow{3}{*}{ViT} &&\cellcolor{gray!18}ViT$_{\text{large}}$&\cellcolor{gray!18}&
\textbf{52.33} & 24.51 & 33.39 && 
\textbf{44.84} & 18.82 & 26.51 &\cellcolor{gray!18} &
\cellcolor{gray!18}\textbf{59.77} &\cellcolor{gray!18} \textbf{34.10} & \cellcolor{gray!18}\textbf{43.42}  \\

&& &&DeiT$_{\text{base}}$ && 
37.17 & 17.22 & 23.54 & &
28.49 & 12.57 & 17.44 & &
38.37 & 21.53 & 27.58  \\ 

&& &&ViTB16-DINO$^{\dag}$&& 
35.19 & 20.97 & 26.28 & &
32.48 & 16.11 & 21.54 & &
40.89 & 28.40 & 33.52  \\ 

\midrule

\multirow{5}{*}{I-21k}
&& MLP && 
MLP-Mixer$_{\text{L16}}$ & &
24.57 & 11.18 & 15.37 & &
20.74 & 9.24 & 12.78 & &
29.03 & 13.61 & 18.53  \\ 

\cmidrule{3-17}

&& \multirow{3}{*}{ViT} && ViT$_{\text{base}}$ & &
45.19 & 25.07 & 32.25 & &
40.43 & 19.51 & 26.32 & &
52.87 & 31.81 & 39.72  \\ 

&& && \cellcolor{gray!18}ViT$_{\text{large}}$ &\cellcolor{gray!18} &
\cellcolor{gray!18}49.15 &\cellcolor{gray!18} \textbf{25.56} &\cellcolor{gray!18} \textbf{33.63} &\cellcolor{gray!18} &
\cellcolor{gray!18}43.29 &\cellcolor{gray!18} \textbf{19.86} &\cellcolor{gray!18} \cellcolor{gray!18}\textbf{27.23} & &
55.23 & 33.26 & 41.52  \\ 

&& && ViT$_{\text{huge}}$ & &
38.22 & 19.24 & 25.59 & &
31.71 & 18.13 & 23.06 & &
51.55 & 30.07 & 37.98  \\ 

\bottomrule
\end{tabularx}
\caption{Results of Embedding Based Methods for the SUN dataset using different features extracted from a diverse set of architecture types pretrained on ImageNet-1k (I-1k) and ImageNet-21k (I-21k). These backbones were trained via: supervised and self-supervised (${\dag}$) learning. The bold numbers correspond to the highest scores per column, and the shaded rows correspond to the most performant image feature per method. +FT indicates the features were fine-tuned with the seen classes from the training set. Surprisingly, Shufflenet and RN50-MOCO features seem to be not suited for this dataset, and using ViT$_{\text{large}}$ pretrained on ImageNet-1k features with ALE beat all methods, including it's counterpart pretrained on ImageNet-21k by a reasonable margin (1.9\%). Moreover, using the ViT$_{\text{large}}$ pretrained on ImageNet-21k features beats all other methods when using DEVISE and ESZSL.
}
\label{tab:sun_embedding_CNN}
\end{table*}

\begin{table*}[!htbp]
\newcolumntype{Y}{>{\raggedright\arraybackslash}X}
\newcolumntype{Z}{>{\centering\arraybackslash}X}
\centering
\footnotesize
\setlength\tabcolsep{1pt}
\renewcommand{\arraystretch}{1.2}

\begin{tabularx}{\textwidth}{l c l c l c YYY c YYY c YYY}
\toprule

\multicolumn{17}{@{\hskip 0.11in}c}{\bf \shortstack{Embedding Based GZSL Methods\\ AWA2 Dataset}}  \\ 
\midrule

{\multirow{2}{*}{\bf \shortstack{Dataset\\Pret. on}}}~ &~~~~&
{\multirow{2}{*}{\bf \shortstack{Arch\\ Type}}}~ &~~~~&
{\multirow{2}{*}{\bf Backbone}}~ &~~~~&
\multicolumn{3}{@{\hskip 0.11in}c}{\bf DEVISE} &~~~~& 
\multicolumn{3}{@{\hskip 0.11in}c}{\bf ESZSL} &~~~~& 
\multicolumn{3}{@{\hskip 0.11in}c}{\bf ALE} \\

\cmidrule{7-9}\cmidrule{11-13}\cmidrule{15-17}

&& && && \textit{Seen} & \textit{Novel} & \textit{Harm.} 
&& \textit{Seen} & \textit{Novel} & \textit{Harm.} 
&& \textit{Seen} & \textit{Novel} & \textit{Harm.} \\

\midrule

\multirow{17}{*}{I-1k} & &
\multirow{12}{*}{CNN} & &
RN101 &&
71.78 & 17.30 & 27.88 & &
88.84 & 4.04 & 7.72 & &
77.59 & 12.15 & 21.01  \\ 

&& &&RN101+FT &&
87.34 & 18.83 & 30.99 & &
93.07 & 6.12 & 11.49 & &
92.64 & 8.25 & 15.16  \\ 

&& &&RN50 &&
86.02 & 19.49 & 31.78 & &
89.05 & 4.75 & 9.02 & &
84.37 & 10.48 & 18.65  \\ 

&& &&\cellcolor{gray!18}RN152 &\cellcolor{gray!18}&
\cellcolor{gray!18}88.30 & \cellcolor{gray!18}\textbf{21.18} & \cellcolor{gray!18}\textbf{34.17} & &
91.31 & 5.94 & 11.15 & &
85.46 & 12.91 & 22.43  \\ 

&& &&GoogleNet &&
68.56 & 17.40 & 27.76 & &
80.11 & 4.26 & 8.10 & &
82.82 & 5.13 & 9.66  \\ 

&& &&VGG16 &&
78.85 & 16.21 & 26.89 & &
90.06 & 3.11 & 6.00 & &
77.34 & 11.52 & 20.06  \\ 

&& &&Alexnet &&
72.91 & 12.17 & 20.86 & &
79.09 & 2.77 & 5.36 & &
79.17 & 6.11 & 11.34  \\ 

&& &&Shufflenet &&
74.74 & 20.04 & 31.61 & &
51.66 & 3.69 & 6.89 & &
80.75 & 8.84 & 15.93  \\ 

&& &&Inceptionv3 &&
74.54 & 8.08 & 14.58 & &
91.49 & 4.43 & 8.46 & &
78.24 & 10.32 & 18.23  \\ 

&& &&Inceptionv3$_{\text{adv}}$ &&
89.33 & 12.14 & 21.38 & &
91.69 & 3.53 & 6.79 & &
82.21 & 8.06 & 14.69  \\ 

\cmidrule{5-17}
&& &&RN50-MOCO$^{\dag}$ &&
78.23 & 11.07 & 19.39 & &
57.62 & 3.73 & 7.01 & &
81.51 & 4.60 & 8.71  \\ 

&& &&RN50-DINO$^{\dag}$ &&
78.97 & 18.11 & 29.46 & &
81.25 & 8.08 & 14.71 & &
82.78 & 7.62 & 13.96  \\

\cmidrule{3-17}

&& MLP&&MLP-Mixer && 
21.06 & 10.87 & 14.34 && 
38.51 & 2.67 & 4.99 && 
94.29 & 12.78 & 22.52  \\ 

\cmidrule{3-17}

&& \multirow{3}{*}{ViT} &&\cellcolor{gray!18}ViT$_{\text{large}}$&\cellcolor{gray!18}&
83.78 & 18.00 & 29.63 &\cellcolor{gray!18}& 
\cellcolor{gray!18}\textbf{97.07} & \cellcolor{gray!18}\textbf{18.62} & \cellcolor{gray!18}\textbf{31.25} & &
92.28 & 13.75 & 23.93  \\

&& &&\cellcolor{gray!18}DeiT$_{\text{base}}$ &\cellcolor{gray!18}& 
\textbf{91.41} & 9.51 & 17.22 & &
94.08 & 2.33 & 4.54 &\cellcolor{gray!18} &
\cellcolor{gray!18}87.37 & \cellcolor{gray!18}\textbf{14.67} & \cellcolor{gray!18}\textbf{25.12}  \\ 

&& &&ViTB16-DINO$^{\dag}$&& 
71.83 & 19.41 & 30.56 & &
92.34 & 5.08 & 9.63 & &
79.73 & 6.55 & 12.10  \\ 

\midrule

\multirow{5}{*}{I-21k}
&& MLP && 
MLP-Mixer$_{\text{L16}}$ & &
82.92 & 10.37 & 18.43 & &
85.48 & 1.47 & 2.90 & &
86.24 & 1.88 & 3.69  \\ 

\cmidrule{3-17}

&& \multirow{3}{*}{ViT} && ViT$_{\text{base}}$ & &
82.35 & 14.05 & 24.00 & &
96.18 & 7.77 & 14.38 & &
\textbf{95.62} & 11.07 & 19.85  \\ 

&& && ViT$_{\text{large}}$ & &
86.49 & 19.53 & 31.87 & &
96.47 & 10.43 & 18.83 & &
95.03 & 12.76 & 22.49  \\ 

&& && ViT$_{\text{huge}}$ & &
80.00 & 12.37 & 21.43 & &
89.57 & 2.55 & 4.95 & &
81.40 & 6.72 & 12.41  \\

\bottomrule
\end{tabularx}
\caption{Results of Embedding Based Methods for the AWA2 dataset using different features extracted from a diverse set of architecture types pretrained on ImageNet-1k (I-1k) and ImageNet-21k (I-21k). These backbones were trained via: supervised and self-supervised (${\dag}$) learning. The bold numbers correspond to the highest scores per column, and the shaded rows correspond to the most performant image feature per method. +FT indicates the features were fine-tuned with the seen classes from the training set. Surprisingly, the most performant visual features are extracted from a RN152 pretrained on ImageNet-1k, using the DEVISE method.
}
\label{tab:awa2_embedding_CNN}
\end{table*}
\begin{table*}[!htbp]
\newcolumntype{Y}{>{\raggedright\arraybackslash}X}
\newcolumntype{Z}{>{\centering\arraybackslash}X}
\centering
\footnotesize
\setlength\tabcolsep{1pt}
\renewcommand{\arraystretch}{1.2}

\begin{tabularx}{\textwidth}{l c l c l c YYY c YYY c YYY}
\toprule

\multicolumn{17}{@{\hskip 0.11in}c}{\bf \shortstack{Generative Based GZSL Methods\\ CUB Dataset}}  \\ 
\midrule

{\multirow{2}{*}{\bf \shortstack{Dataset\\Pret. on}}}~ &~~~~&
{\multirow{2}{*}{\bf \shortstack{Arch\\ Type}}}~ &~~~~&
{\multirow{2}{*}{\bf Backbone}}~ &~~~~&
\multicolumn{3}{@{\hskip 0.11in}c}{\bf tfVAEGAN} &~~~~& 
\multicolumn{3}{@{\hskip 0.11in}c}{\bf CADA-VAE} &~~~~& 
\multicolumn{3}{@{\hskip 0.11in}c}{\bf CE} \\

\cmidrule{7-9}\cmidrule{11-13}\cmidrule{15-17}

&& && && \textit{Seen} & \textit{Novel} & \textit{Harm.} 
&& \textit{Seen} & \textit{Novel} & \textit{Harm.} 
&& \textit{Seen} & \textit{Novel} & \textit{Harm.} \\

\midrule

\multirow{17}{*}{I-1k} & &
\multirow{12}{*}{CNN} & &

RN101 &&
57.08 & 42.88 & 48.97 &&
58.27 & 49.71 & 53.65  &&
60.09 & \textbf{49.05} & 54.01  \\ 

&& &&\cellcolor{gray!18}RN101+FT &\cellcolor{gray!18}&
72.44 & 53.66 & 61.65 &&
76.45 & 57.53 & 65.65  &&
\cellcolor{gray!18}\textbf{76.71} &\cellcolor{gray!18} 48.81 &\cellcolor{gray!18} \textbf{59.66}  \\  

&& &&RN50 &&
49.29 & 42.35 & 45.55 &&
45.88 & 38.95 & 42.13  &&
42.79 & 35.46 & 38.78  \\ 

&& &&RN152 &&
50.23 & 44.48 & 47.18 &&
47.58 & 41.64 & 44.41  &&
45.77 & 35.52 & 40.00  \\ 

&& &&GoogleNet &&
38.54 & 33.34 & 35.76 &&
33.84 & 30.20 & 31.92  &&
33.72 & 26.05 & 29.39  \\ 

&& &&VGG16 &&
36.67 & 38.46 & 37.54 &&
37.12 & 35.38 & 36.23  &&
35.00 & 37.84 & 36.36  \\  

&& &&Alexnet &&
21.48 & 32.52 & 25.87 &&
22.36 & 24.31 & 23.29  &&
23.73 & 28.62 & 25.95  \\  

&& &&Shufflenet &&
51.62 & 43.83 & 47.41 &&
43.14 & 38.56 & 40.72  &&
48.95 & 37.69 & 42.59  \\ 

&& &&Inceptionv3 &&
54.01 & 50.41 & 52.15 &&
50.32 & 39.87 & 44.49  &&
54.80 & 38.45 & 45.19  \\  

&& &&Inceptionv3$_{\text{adv}}$ &&
62.81 & 41.34 & 49.86 &&
50.28 & 37.90 & 43.22  &&
54.83 & 35.41 & 43.03  \\  

\cmidrule{5-17}
&& &&RN50-MOCO$^{\dag}$ &&
41.88 & 29.13 & 34.36 &&
27.40 & 22.39 & 24.64  &&
34.01 & 24.09 & 28.21  \\  

&& &&RN50-DINO$^{\dag}$ &&
64.11 & 53.84 & 58.53 &&
55.05 & 47.59 & 51.05  &&
62.45 & 45.13 & 52.39  \\

\cmidrule{3-17}

&& MLP&&MLP-Mixer && 
18.64 & 15.87 & 17.15 &&
11.74 & 18.45 & 14.35  &&
10.99 & 10.94 & 10.97  \\  

\cmidrule{3-17}

&& \multirow{3}{*}{ViT} &&ViT$_{\text{large}}$&&
\textbf{80.34} & 54.34 & 64.83 &&
61.23 & 53.31 & 56.99  &&
70.22 & 43.07 & 53.39  \\

&& &&DeiT$_{\text{base}}$ && 
73.29 & 49.44 & 59.05 &&
60.39 & 50.05 & 54.74  &&
55.68 & 38.68 & 45.65  \\  

&& &&ViTB16-DINO$^{\dag}$&& 
76.82 & 57.94 & 66.06 &&
71.95 & 55.37 & 62.58  &&
61.29 & 45.47 & 52.21  \\

\midrule

\multirow{5}{*}{I-21k}
&& MLP && 
MLP-Mixer$_{\text{L16}}$ & &
30.91 & 28.77 & 29.80 &&
28.68 & 25.19 & 26.82  &&
17.46 & 20.76 & 18.96  \\

\cmidrule{3-17}

&& \multirow{3}{*}{ViT} && ViT$_{\text{base}}$ & &
74.16 & 71.13 & 72.61 &&
74.46 & 60.77 & 66.93  &&
61.01 & 51.25 & 55.71  \\

&& && ViT$_{\text{large}}$ & &
76.95 & 61.56 & 68.40 &&
72.54 & 58.94 & 65.04  &&
67.16 & 46.94 & 55.26  \\ 

&& && ViT$_{\text{huge}}$ & &
75.15 & 62.76 & 68.40 &&
70.53 & 60.50 & 65.13  &&
49.37 & 43.76 & 46.40  \\

&& &&\cellcolor{gray!18} ViT$_{\text{huge}}$+FT &\cellcolor{gray!18} &
\cellcolor{gray!18}78.32 &\cellcolor{gray!18} \textbf{76.26} &\cellcolor{gray!18} \textbf{77.27} &\cellcolor{gray!18}&
\cellcolor{gray!18}\textbf{77.99} &\cellcolor{gray!18} \textbf{74.46} &\cellcolor{gray!18} \textbf{76.18}  &&
70.87&	44.66 &	54.79 \\

\bottomrule
\end{tabularx}
\caption{Results of Generative Based Methods for the CUB dataset using different features extracted from a diverse set of architecture types pretrained on ImageNet-1k (I-1k) and ImageNet-21k (I-21k). These backbones were trained via: supervised and self-supervised (${\dag}$) learning. The bold numbers correspond to the highest scores per column, and the shaded rows correspond to the most performant image feature per method. +FT indicates the features were fine-tuned with the seen classes from the training set. The most performant visual features are extracted from a ViT$_{\text{huge}}$ pretrained on ImageNet-21k and fine-tuned with the seen classes, using the tfVAEGAN method.
}
\label{tab:cub_generative_CNN}
\end{table*}

\begin{table*}[!htbp]
\newcolumntype{Y}{>{\raggedright\arraybackslash}X}
\newcolumntype{Z}{>{\centering\arraybackslash}X}
\centering
\footnotesize
\setlength\tabcolsep{1pt}
\renewcommand{\arraystretch}{1.2}

\begin{tabularx}{\textwidth}{l c l c l c YYY c YYY c YYY}
\toprule

\multicolumn{17}{@{\hskip 0.11in}c}{\bf \shortstack{Generative Based GZSL Methods\\ SUN Dataset}}  \\ 
\midrule

{\multirow{2}{*}{\bf \shortstack{Dataset\\Pret. on}}}~ &~~~~&
{\multirow{2}{*}{\bf \shortstack{Arch\\ Type}}}~ &~~~~&
{\multirow{2}{*}{\bf Backbone}}~ &~~~~&
\multicolumn{3}{@{\hskip 0.11in}c}{\bf tfVAEGAN} &~~~~& 
\multicolumn{3}{@{\hskip 0.11in}c}{\bf CADA-VAE} &~~~~& 
\multicolumn{3}{@{\hskip 0.11in}c}{\bf CE} \\

\cmidrule{7-9}\cmidrule{11-13}\cmidrule{15-17}

&& && && \textit{Seen} & \textit{Novel} & \textit{Harm.} 
&& \textit{Seen} & \textit{Novel} & \textit{Harm.} 
&& \textit{Seen} & \textit{Novel} & \textit{Harm.} \\

\midrule

\multirow{17}{*}{I-1k} & &
\multirow{12}{*}{CNN} & &

\cellcolor{gray!18}RN101 &\cellcolor{gray!18}&
38.95 & 45.62 & 42.03  & ~ &
34.15 & 48.96 & 40.23  &\cellcolor{gray!18} ~ &
\cellcolor{gray!18}51.24 &\cellcolor{gray!18} \textbf{55.83} &\cellcolor{gray!18} \textbf{53.44}  \\ 

&& &&RN101+FT &&
35.08 & 38.06 & 36.51  & ~ &
39.84 & 51.60 & 44.97  & ~ &
29.07 & 38.75 & 33.22  \\ 

&& &&RN50 &&
34.61 & 45.07 & 39.15  & ~ &
34.07 & 41.53 & 37.43  & ~ & 23.95 & 42.36 & 30.60  \\

&& &&RN152 &&
35.35 & 45.97 & 39.97  & ~ &
37.05 & 40.00 & 38.47  & ~ & 26.20 & 43.33 & 32.66  \\ 

&& &&GoogleNet &&
27.17 & 38.26 & 31.78  & ~ &
24.96 & 36.32 & 29.59  & ~ &
20.43 & 38.96 & 26.80  \\ 

&& &&VGG16 &&
25.04 & 28.61 & 26.71  & ~ &
31.32 & 37.85 & 34.27  & ~ &
24.11 & 36.81 & 29.13  \\ 

&& &&Alexnet &&
25.27 & 34.72 & 29.25  & ~ &
16.51 & 23.06 & 19.24  & ~ &
13.84 & 30.90 & 19.12  \\ 

&& &&Shufflenet &&
31.59 & 42.71 & 36.32  & ~ &
30.62 & 37.64 & 33.77  & ~ &
24.92 & 37.50 & 29.94  \\

&& &&Inceptionv3 &&
30.00 & 32.15 & 31.04  & ~ &
32.64 & 38.26 & 35.23  & ~ &
26.09 & 37.43 & 30.74  \\

&& &&Inceptionv3$_{\text{adv}}$ &&
33.37 & 45.42 & 38.47  & ~ &
32.02 & 40.83 & 35.89  & ~ &
27.09 & 39.44 & 32.12  \\ 

\cmidrule{5-17}
&& &&RN50-MOCO$^{\dag}$ &&
37.44 & 42.99 & 40.02  & ~ &
35.08 & 38.13 & 36.54  & ~ &
30.62 & 44.24 & 36.19  \\ 

&& &&RN50-DINO$^{\dag}$ &&
42.60 & 46.67 & 44.54  & ~ &
41.16 & 46.39 & 43.62  & ~ &
29.38 & 55.56 & 38.43  \\

\cmidrule{3-17}

&& MLP&&MLP-Mixer && 
24.96 & 32.22 & 28.13  & ~ &
6.82 & 12.78 & 8.89  & ~ &
8.64 & 8.33 & 8.49  \\ 

\cmidrule{3-17}

&& \multirow{3}{*}{ViT} &&\cellcolor{gray!18}ViT$_{\text{large}}$&\cellcolor{gray!18}&
\cellcolor{gray!18}55.12 &\cellcolor{gray!18} \textbf{64.93} &\cellcolor{gray!18} \textbf{59.62}  &\cellcolor{gray!18} ~ &
\cellcolor{gray!18}\textbf{52.64} &\cellcolor{gray!18} \textbf{61.32} &\cellcolor{gray!18} \textbf{56.65}  & ~ &
50.70 & 52.78 & 51.72  \\

&& &&DeiT$_{\text{base}}$ && 
34.34 & 44.72 & 38.85  & ~ &
37.44 & 44.51 & 40.67  & ~ &
36.36 & 37.57 & 36.95  \\

&& &&ViTB16-DINO$^{\dag}$&& 
42.64 & 52.71 & 47.14  & ~ &
42.52 & 51.18 & 46.45  & ~ &
40.39 & 45.56 & 42.82  \\

\midrule

\multirow{5}{*}{I-21k}
&& MLP && 
MLP-Mixer$_{\text{L16}}$ & &
24.22 & 43.03 & 28.30  & ~ &
24.77 & 29.65 & 26.99  & ~ &
23.84 & 20.00 & 21.75  \\

\cmidrule{3-17}

&& \multirow{3}{*}{ViT} && ViT$_{\text{base}}$ & &
54.19 & 58.75 & 56.38  & ~ &
52.02 & 60.14 & 55.78  & ~ &
\textbf{51.53} & 50.74 & 51.13  \\

&& && ViT$_{\text{large}}$ & &
\textbf{57.13} & 61.32 & 59.15  & ~ &
52.83 & 60.69 & 56.49  & ~ &
50.12 & 50.69 & 50.40  \\

&& && ViT$_{\text{huge}}$ & &
43.37 & 53.61 & 47.95  & ~ &
47.64 & 51.18 & 49.34  & ~ &
49.79 & 16.05 & 24.27  \\

&& && ViT$_{\text{huge}}$+FT & &
44.26 & 54.24 & 48.75  & ~ &
44.99 & 55.35 & 49.64  & ~ &
6.86 & 53.61 & 12.16  \\

\bottomrule
\end{tabularx}
\caption{Results of Generative Based Methods for the SUN dataset using different features extracted from a diverse set of architecture types pretrained on ImageNet-1k (I-1k) and ImageNet-21k (I-21k). These backbones were trained via: supervised and self-supervised (${\dag}$) learning. The bold numbers correspond to the highest scores per column, and the shaded rows correspond to the most performant image feature per method. +FT indicates the features were fine-tuned with the seen classes from the training set. Surprisingly, the CE method does not seem to get any significant advantage from any of the ViT features, and overall, the most performant visual features are extracted from a ViT$_{\text{large}}$ pretrained on ImageNet-1k using the tfVAEGAN method.
}
\label{tab:sun_generative_CNN}
\end{table*}

\begin{table*}[!htbp]
\newcolumntype{Y}{>{\raggedright\arraybackslash}X}
\newcolumntype{Z}{>{\centering\arraybackslash}X}
\centering
\footnotesize
\setlength\tabcolsep{1pt}
\renewcommand{\arraystretch}{1.2}

\begin{tabularx}{\textwidth}{l c l c l c YYY c YYY c YYY}
\toprule

\multicolumn{17}{@{\hskip 0.11in}c}{\bf \shortstack{Generative Based GZSL Methods\\ AWA2 Dataset}}  \\ 
\midrule

{\multirow{2}{*}{\bf \shortstack{Dataset\\Pret. on}}}~ &~~~~&
{\multirow{2}{*}{\bf \shortstack{Arch\\ Type}}}~ &~~~~&
{\multirow{2}{*}{\bf Backbone}}~ &~~~~&
\multicolumn{3}{@{\hskip 0.11in}c}{\bf tfVAEGAN} &~~~~& 
\multicolumn{3}{@{\hskip 0.11in}c}{\bf CADA-VAE} &~~~~& 
\multicolumn{3}{@{\hskip 0.11in}c}{\bf CE} \\

\cmidrule{7-9}\cmidrule{11-13}\cmidrule{15-17}

&& && && \textit{Seen} & \textit{Novel} & \textit{Harm.} 
&& \textit{Seen} & \textit{Novel} & \textit{Harm.} 
&& \textit{Seen} & \textit{Novel} & \textit{Harm.} \\

\midrule

\multirow{17}{*}{I-1k} & &
\multirow{12}{*}{CNN} & &

RN101 &&
75.48 & 59.56 & 66.58  & ~ &
75.95 & 54.76 & 63.87  & ~ &
69.26 & 56.03 & 61.95  \\

&& &&RN101+FT &&
84.81 & 58.44 & 69.20  & ~ &
77.74 & 59.95 & 69.14  & ~ &
83.43 & 47.66 & 60.66  \\ 

&& &&RN50 &&
79.91 & 55.83 & 65.73  & ~ &
74.56 & 59.48 & 67.80  & ~ &
71.72 & 46.65 & 56.53  \\

&& &&RN152 &&
84.48 & 60.01 & 70.17  & ~ &
\textbf{88.88} & 57.69 & 70.48  & ~ &
76.74 & 40.25 & 52.8  \\

&& &&GoogleNet &&
69.08 & 55.17 & 61.35  & ~ &
71.51 & 53.22 & 62.22  & ~ &
73.36 & 47.63 & 57.76  \\

&& &&VGG16 &&
78.95 & 52.54 & 63.09  & ~ &
74.17 & 58.92 & 67.26  & ~ &
62.37 & 46.85 & 53.51  \\ 

&& &&Alexnet &&
64.31 & 41.11 & 50.16  & ~ &
61.98 & 40.80 & 49.60  & ~ &
57.13 & 39.39 & 46.63  \\  

&& &&Shufflenet &&
69.53 & 55.38 & 61.66  & ~ &
68.07 & 54.09 & 61.84  & ~ &
71.87 & 50.11 & 59.05  \\

&& &&Inceptionv3 &&
84.00 & 58.45 & 68.94  & ~ &
78.97 & 61.77 & 70.86  & ~ &
74.69 & 52.09 & 61.38  \\ 

&& &&Inceptionv3$_{\text{adv}}$ &&
85.48 & 59.21 & 69.96  & ~ &
82.15 & 53.43 & 65.23  & ~ &
66.96 & 53.73 & 59.62  \\  

\cmidrule{5-17}
&& &&RN50-MOCO$^{\dag}$ &&
68.51 & 51.75 & 58.96  & ~ &
65.52 & 55.53 & 62.02  & ~ &
64.77 & 40.92 & 50.15  \\  

&& &&RN50-DINO$^{\dag}$ &&
74.58 & 60.00 & 66.50  & ~ &
73.36 & 55.23 & 64.30  & ~ &
74.22 & 45.46 & 56.38  \\

\cmidrule{3-17}

&& MLP&&MLP-Mixer && 
28.14 & 25.27 & 26.63  & ~ &
14.01 & 41.56 & 27.90  & ~ &
22.93 & 20.68 & 21.75  \\

\cmidrule{3-17}

&& \multirow{3}{*}{ViT} &&\cellcolor{gray!18}ViT$_{\text{large}}$&\cellcolor{gray!18}&
90.14 & 68.28 & 77.70  & ~ &
85.55 & 70.62 & 79.25  &\cellcolor{gray!18} ~ &
\cellcolor{gray!18}81.37 &\cellcolor{gray!18} \textbf{69.14} &\cellcolor{gray!18} \textbf{74.75}  \\

&& &&DeiT$_{\text{base}}$ && 
84.63 & 51.89 & 64.33  & ~ &
77.83 & 59.04 & 68.50  & ~ &
79.35 & 50.62 & 61.81  \\ 

&& &&ViTB16-DINO$^{\dag}$&& 
77.64 & 57.77 & 66.24  & ~ &
75.69 & 63.90 & 71.25  & ~ &
81.66 & 54.65 & 65.48  \\

\midrule

\multirow{5}{*}{I-21k}
&& MLP && 
MLP-Mixer$_{\text{L16}}$ & &
72.63 & 42.47 & 53.60  & ~ &
70.15 & 51.01 & 60.12  & ~ &
61.51 & 39.01 & 47.74  \\

\cmidrule{3-17}

&& \multirow{3}{*}{ViT} && ViT$_{\text{base}}$ & &
54.19 & 58.75 & 56.38  & ~ &
84.48 & 67.42 & 76.67  & ~ &
77.80 & 49.54 & 60.53  \\

&& &&\cellcolor{gray!18} ViT$_{\text{large}}$ &\cellcolor{gray!18} &
\cellcolor{gray!18}\textbf{91.05} &\cellcolor{gray!18} \textbf{63.58} &\cellcolor{gray!18} \textbf{74.87}  &\cellcolor{gray!18} ~ &
\cellcolor{gray!18}88.69 &\cellcolor{gray!18} \textbf{70.75} &\cellcolor{gray!18} \textbf{80.40}  & ~ &
\textbf{78.32} & 59.21 & 67.44  \\

&& && ViT$_{\text{huge}}$ & &
88.85 & 60.79 & 72.19  & ~ &
85.68 & 60.95 & 72.25  & ~ &
75.57 & 53.24 & 62.47  \\

&& && ViT$_{\text{huge}}$+FT & &
68.23 & 61.63 & 64.76  & ~ &
80.69 & 60.76 & 69.32  & ~ &
75.23 & 60.10 & 66.82  \\

\bottomrule
\end{tabularx}
\caption{Results of Generative Based Methods for the AWA2 dataset using different features extracted from a diverse set of architecture types pretrained on ImageNet-1k (I-1k) and ImageNet-21k (I-21k). These backbones were trained via: supervised and self-supervised (${\dag}$) learning. The bold numbers correspond to the highest scores per column, and the shaded rows correspond to the most performant image feature per method. +FT indicates the features were fine-tuned with the seen classes from the training set. The most performant visual features are extracted from a ViT$_{\text{huge}}$ pretrained on ImageNet-21k and fine-tuned with the seen classes, using the CADA-VAE method. More interestingly, the features from a ViT$_{\text{large}}$ pretrained on ImageNet-1k seem competitive with the features from a ViT$_{\text{large}}$ pretrained on ImageNet-21k for the CE and tfVAEGAN methods respectively.
}
\label{tab:awa2_generative_CNN}
\end{table*}
\begin{table*}[!htbp]
\newcolumntype{Y}{>{\raggedright\arraybackslash}X}
\newcolumntype{Z}{>{\centering\arraybackslash}X}
\centering
\footnotesize
\setlength\tabcolsep{1pt}
\renewcommand{\arraystretch}{1.2}

\begin{tabularx}{\textwidth}{l c l c l c YYY c YYY}
\toprule

\multicolumn{13}{@{\hskip 0.11in}c}{\bf \shortstack{Disentanglement Based GZSL Methods\\ CUB Dataset}}  \\ 
\midrule

{\multirow{2}{*}{\bf \shortstack{Dataset\\Pret. on}}}~ &~~~~&
{\multirow{2}{*}{\bf \shortstack{Arch\\ Type}}}~ &~~~~&
{\multirow{2}{*}{\bf Backbone}}~ &~~~~&
\multicolumn{3}{@{\hskip 0.11in}c}{\bf SDGZSL} &~~~~& 
\multicolumn{3}{@{\hskip 0.11in}c}{\bf FREE}  \\

\cmidrule{7-9}\cmidrule{11-13}

&& && && \textit{Seen} & \textit{Novel} & \textit{Harm.} 
&& \textit{Seen} & \textit{Novel} & \textit{Harm.} \\

\midrule

\multirow{17}{*}{I-1k} & &
\multirow{12}{*}{CNN} & &

RN101 &&
56.41 & 51.89 & 54.06  & ~ &
58.30 & 55.10 & 56.70  \\

&& &&RN101+FT &&
75.00 & 64.26 & 69.21  & ~ &
75.30 & 56.00 & 64.20  \\

&& &&RN50 &&
48.48 & 37.53 & 42.31  & ~ &
48.40 & 41.10 & 44.45  \\

&& &&RN152 &&
48.22 & 43.37 & 45.67  & ~ &
50.70 & 43.90 & 47.10  \\

&& &&GoogleNet &&
32.72 & 32.88 & 32.80  & ~ &
33.78 & 34.34 & 34.06  \\

&& &&VGG16 &&
37.20 & 34.41 & 35.75  & ~ &
30.02 & 40.14 & 34.35  \\

&& &&Alexnet &&
27.12 & 24.98 & 26.01  & ~ &
15.47 & 34.16 & 21.30  \\ 

&& &&Shufflenet &&
47.66 & 38.90 & 42.84  & ~ &
47.65 & 43.64 & 45.56  \\

&& &&Inceptionv3 &&
56.70 & 46.39 & 51.03  & ~ &
61.88 & 44.60 & 51.83  \\ 

&& &&Inceptionv3$_{\text{adv}}$ &&
51.38 & 46.86 & 49.01  & ~ &
60.82 & 42.02 & 49.70  \\

\cmidrule{5-13}
&& &&RN50-MOCO$^{\dag}$ &&
36.67 & 30.22 & 33.14  & ~ &
42.53 & 29.38 & 34.75  \\

&& &&RN50-DINO$^{\dag}$ &&
59.27 & 51.68 & 55.22  & ~ &
66.62 & 53.08 & 59.08  \\ 

\cmidrule{3-13}

&& MLP&&MLP-Mixer && 
17.66 & 16.74 & 17.19  & ~ &
17.59 & 14.43 & 15.85  \\ 

\cmidrule{3-13}

&& \multirow{3}{*}{ViT} &&ViT$_{\text{large}}$&&
74.17 & 63.17 & 68.23  & ~ &
69.53 & 56.38 & 62.27  \\

&& &&DeiT$_{\text{base}}$ && 
62.91 & 54.40 & 58.35  & ~ &
61.80 & 43.80 & 51.27  \\ 

&& &&ViTB16-DINO$^{\dag}$&& 
68.72 & 64.25 & 66.41  & ~ &
73.70 & 53.46 & 61.97  \\

\midrule

\multirow{5}{*}{I-21k}
&& MLP && 
MLP-Mixer$_{\text{L16}}$ & &
33.77 & 26.13 & 29.46  & ~ &
31.58 & 23.61 & 27.02  \\

\cmidrule{3-13}

&& \multirow{3}{*}{ViT} && ViT$_{\text{base}}$ & &
78.01 & 70.10 & 73.84  & ~ &
67.50 & 66.15 & 66.82  \\

&& && ViT$_{\text{large}}$ & &
79.55 & 64.37 & 71.16  & ~ &
73.07 & 52.91 & 61.37  \\

&& && ViT$_{\text{huge}}$ & &
75.05 & 70.58 & 72.75  & ~ &
66.17 & 67.57 & 66.87  \\

&& &&\cellcolor{gray!18} ViT$_{\text{huge}}$+FT &\cellcolor{gray!18} &
\cellcolor{gray!18}\textbf{79.68} &\cellcolor{gray!18} \textbf{73.27} &\cellcolor{gray!18} \textbf{76.34}  &\cellcolor{gray!18} ~ &
\cellcolor{gray!18}\textbf{80.57} &\cellcolor{gray!18} \textbf{69.71} &\cellcolor{gray!18} \textbf{74.75}  \\

\bottomrule
\end{tabularx}
\caption{Results of Disentanglement Based Methods for the CUB dataset using different features extracted from a diverse set of architecture types pretrained on ImageNet-1k (I-1k) and ImageNet-21k (I-21k). These backbones were trained via: supervised and self-supervised (${\dag}$) learning. The bold numbers correspond to the highest scores per column, and the shaded rows correspond to the most performant image feature per method. +FT indicates the features were fine-tuned with the seen classes from the training set. The most performant visual features are extracted from a ViT$_{\text{huge}}$ pretrained on ImageNet-21k and fine-tuned with the seen classes, using the SDGZSL method.
}
\label{tab:cub_disentanglement_CNN}
\end{table*}

\begin{table*}[!htbp]
\newcolumntype{Y}{>{\raggedright\arraybackslash}X}
\newcolumntype{Z}{>{\centering\arraybackslash}X}
\centering
\footnotesize
\setlength\tabcolsep{1pt}
\renewcommand{\arraystretch}{1.2}

\begin{tabularx}{\textwidth}{l c l c l c YYY c YYY}
\toprule

\multicolumn{13}{@{\hskip 0.11in}c}{\bf \shortstack{Disentanglement Based GZSL Methods\\ SUN Dataset}}  \\ 
\midrule

{\multirow{2}{*}{\bf \shortstack{Dataset\\Pret. on}}}~ &~~~~&
{\multirow{2}{*}{\bf \shortstack{Arch\\ Type}}}~ &~~~~&
{\multirow{2}{*}{\bf Backbone}}~ &~~~~&
\multicolumn{3}{@{\hskip 0.11in}c}{\bf SDGZSL} &~~~~& 
\multicolumn{3}{@{\hskip 0.11in}c}{\bf FREE}  \\

\cmidrule{7-9}\cmidrule{11-13}

&& && && \textit{Seen} & \textit{Novel} & \textit{Harm.} 
&& \textit{Seen} & \textit{Novel} & \textit{Harm.} \\

\midrule

\multirow{17}{*}{I-1k} & &
\multirow{12}{*}{CNN} & &

RN101 &&
36.67 & 44.44 & 40.18  & ~ &
37.20 & 47.40 & 41.70  \\

&& &&RN101+FT &&
38.57 & 49.44 & 43.33  & ~ &
41.12 & 47.01 & 43.87  \\

&& &&RN50 &&
33.33 & 41.87 & 37.12  & ~ &
33.99 & 42.71 & 37.86  \\

&& &&RN152 &&
34.77 & 44.44 & 39.01  & ~ &
35.12 & 48.19 & 40.63  \\

&& &&GoogleNet &&
26.24 & 31.94 & 28.81  & ~ &
26.32 & 35.69 & 30.30  \\

&& &&VGG16 &&
30.81 & 33.82 & 32.25  & ~ &
27.75 & 40.00 & 32.77  \\

&& &&Alexnet &&
22.64 & 25.76 & 24.10  & ~ &
16.09 & 38.75 & 22.73  \\

&& &&Shufflenet &&
31.55 & 36.39 & 33.80  & ~ &
29.03 & 39.44 & 33.45  \\

&& &&Inceptionv3 &&
31.05 & 43.26 & 36.15  & ~ &
28.91 & 46.11 & 35.54  \\ 

&& &&Inceptionv3$_{\text{adv}}$ &&
31.43 & 44.37 & 36.80  & ~ &
35.12 & 38.54 & 36.75  \\

\cmidrule{5-13}
&& &&RN50-MOCO$^{\dag}$ &&
36.16 & 37.85 & 36.99  & ~ &
36.12 & 41.94 & 38.82  \\

&& &&RN50-DINO$^{\dag}$ &&
40.54 & 49.24 & 44.47  & ~ &
42.95 & 50.07 & 46.23  \\

\cmidrule{3-13}

&& MLP&&MLP-Mixer && 
6.94 & 8.40 & 7.60  & ~ &
6.98 & 13.06 & 9.09  \\

\cmidrule{3-13}

&& \multirow{3}{*}{ViT} &&\cellcolor{gray!18}ViT$_{\text{large}}$&\cellcolor{gray!18}&
\cellcolor{gray!18}51.59 &\cellcolor{gray!18} 63.75 &\cellcolor{gray!18} \textbf{57.03}  &\cellcolor{gray!18} ~ &
\cellcolor{gray!18}\textbf{54.15} &\cellcolor{gray!18} 57.22 &\cellcolor{gray!18} \textbf{55.64}  \\

&& &&DeiT$_{\text{base}}$ && 
31.94 & 45.69 & 37.60  & ~ &
36.98 & 43.54 & 39.99  \\

&& &&ViTB16-DINO$^{\dag}$&& 
40.04 & 51.53 & 45.06  & ~ &
39.61 & 53.19 & 45.41  \\

\midrule

\multirow{5}{*}{I-21k}
&& MLP && 
MLP-Mixer$_{\text{L16}}$ & &
19.92 & 32.99 & 24.84  & ~ &
25.23 & 30.63 & 27.67  \\

\cmidrule{3-13}

&& \multirow{3}{*}{ViT} && ViT$_{\text{base}}$ & &
\textbf{51.63} & 55.56 & 53.52  & ~ &
52.29 & 55.90 & 54.03  \\

&& && ViT$_{\text{large}}$ & &
32.13 & \textbf{68.96} & 43.84  & ~ &
49.34 & \textbf{59.44} & 53.92  \\

&& && ViT$_{\text{huge}}$ & &
30.43 & 62.01 & 40.82  & ~ &
44.42 & 50.00 & 47.04  \\

&& && ViT$_{\text{huge}}$+FT & &
39.53 & 47.99 & 43.35  & ~ &
44.57 & 52.22 & 48.10  \\

\bottomrule
\end{tabularx}
\caption{Results of Disentanglement Based Methods for the SUN dataset using different features extracted from a diverse set of architecture types pretrained on ImageNet-1k (I-1k) and ImageNet-21k (I-21k). These backbones were trained via: supervised and self-supervised (${\dag}$) learning. The bold numbers correspond to the highest scores per column, and the shaded rows correspond to the most performant image feature per method. +FT indicates the features were fine-tuned with the seen classes from the training set. The most performant visual features are extracted from a ViT$_{\text{large}}$ pretrained on ImageNet-1k using the SDGZSL method.
}
\label{tab:sun_disentanglement_CNN}
\end{table*}

\begin{table*}[!htbp]
\newcolumntype{Y}{>{\raggedright\arraybackslash}X}
\newcolumntype{Z}{>{\centering\arraybackslash}X}
\centering
\footnotesize
\setlength\tabcolsep{1pt}
\renewcommand{\arraystretch}{1.2}

\begin{tabularx}{\textwidth}{l c l c l c YYY c YYY}
\toprule

\multicolumn{13}{@{\hskip 0.11in}c}{\bf \shortstack{Disentanglement Based GZSL Methods\\ AWA2 Dataset}}  \\ 
\midrule

{\multirow{2}{*}{\bf \shortstack{Dataset\\Pret. on}}}~ &~~~~&
{\multirow{2}{*}{\bf \shortstack{Arch\\ Type}}}~ &~~~~&
{\multirow{2}{*}{\bf Backbone}}~ &~~~~&
\multicolumn{3}{@{\hskip 0.11in}c}{\bf SDGZSL} &~~~~& 
\multicolumn{3}{@{\hskip 0.11in}c}{\bf FREE}  \\

\cmidrule{7-9}\cmidrule{11-13}

&& && && \textit{Seen} & \textit{Novel} & \textit{Harm.} 
&& \textit{Seen} & \textit{Novel} & \textit{Harm.} \\

\midrule

\multirow{17}{*}{I-1k} & &
\multirow{12}{*}{CNN} & &

RN101 &&
75.27 & 58.65 & 65.93  & ~ &
75.22 & \textbf{56.00} & 64.20  \\

&& &&RN101+FT &&
83.99 & 60.35 & 70.24  & ~ &
\textbf{87.66} & 47.56 & 61.67  \\

&& &&RN50 &&
75.25 & 64.99 & 69.75  & ~ &
84.75 & 41.93 & 56.10  \\

&& &&RN152 &&
79.08 & 68.38 & 73.34  & ~ &
86.78 & 50.13 & 63.55  \\

&& &&GoogleNet &&
71.62 & 54.53 & 61.91  & ~ &
72.03 & 49.45 & 58.64  \\

&& &&VGG16 &&
78.12 & 57.02 & 65.93  & ~ &
76.39 & 54.86 & 63.86  \\

&& &&Alexnet &&
61.06 & 49.09 & 54.43  & ~ &
60.53 & 48.34 & 53.76  \\

&& &&Shufflenet &&
74.12 & 52.98 & 61.79  & ~ &
68.45 & 50.69 & 58.25  \\

&& &&Inceptionv3 &&
84.46 & 57.82 & 68.65  & ~ &
84.40 & 43.32 & 57.26  \\

&& &&\cellcolor{gray!18}Inceptionv3$_{\text{adv}}$ &\cellcolor{gray!18}&
80.22 & 57.68 & 67.11  &\cellcolor{gray!18} ~ &
\cellcolor{gray!18}85.38 &\cellcolor{gray!18} 51.71 &\cellcolor{gray!18} \textbf{64.41}  \\

\cmidrule{5-13}
&& &&RN50-MOCO$^{\dag}$ &&
66.26 & 48.95 & 56.31  & ~ &
72.37 & 48.88 & 58.35  \\

&& &&RN50-DINO$^{\dag}$ &&
68.56 & 64.99 & 66.73  & ~ &
74.89 & 52.41 & 61.67  \\

\cmidrule{3-13}

&& MLP&&MLP-Mixer && 
31.69 & 25.2 & 28.07  & ~ &
28.29 & 22.66 & 25.16  \\ 

\cmidrule{3-13}

&& \multirow{3}{*}{ViT} &&\cellcolor{gray!18}ViT$_{\text{large}}$&\cellcolor{gray!18}&
\cellcolor{gray!18}86.87 &\cellcolor{gray!18} \textbf{70.37} &\cellcolor{gray!18} \textbf{77.75}  & ~ &
87.29 & 48.14 & 62.06  \\

&& &&DeiT$_{\text{base}}$ && 
80.13 & 56.06 & 65.96  & ~ &
72.55 & 41.13 & 52.50  \\

&& &&ViTB16-DINO$^{\dag}$&& 
77.10 & 56.00 & 64.88  & ~ &
80.40 & 44.79 & 57.53  \\

\midrule

\multirow{5}{*}{I-21k}
&& MLP && 
MLP-Mixer$_{\text{L16}}$ & &
52.94 & 37.09 & 43.62  & ~ &
72.82 & 37.4 & 49.42  \\

\cmidrule{3-13}

&& \multirow{3}{*}{ViT} && ViT$_{\text{base}}$ & &
\textbf{88.00} & 60.76 & 71.88  & ~ &
77.14 & 48.44 & 59.51  \\

&& && ViT$_{\text{large}}$ & &
86.04 & 64.65 & 73.83  & ~ &
80.58 & 43.4 & 56.42  \\

&& && ViT$_{\text{huge}}$ & &
83.33 & 62.74 & 71.58  & ~ &
87.43 & 43.28 & 57.90  \\

&& && ViT$_{\text{huge}}$+FT & &
76.94 & 58.43 & 66.42  & ~ &
82.36 & 43.10 & 56.59  \\

\bottomrule
\end{tabularx}
\caption{Results of Disentanglement Based Methods for the AWA2 dataset using different features extracted from a diverse set of architecture types pretrained on ImageNet-1k (I-1k) and ImageNet-21k (I-21k). These backbones were trained via: supervised and self-supervised (${\dag}$) learning. The bold numbers correspond to the highest scores per column, and the shaded rows correspond to the most performant image feature per method. +FT indicates the features were fine-tuned with the seen classes from the training set. The most performant visual features are extracted from a ViT$_{\text{large}}$ pretrained on ImageNet-1k using the SDGZSL method.
}
\label{tab:awa2_disentanglement_CNN}
\end{table*}

Finally, in Table~\ref{tab:gzsl_using_clip_features_full}, we show full results of generative and disentanglement based methods for the CUB, SUN and AWA2 datasets when using different features extracted from different size and architectures of the visual encoder from all available OpenAI CLIP~\cite{CLIP}~\footnote{\href{https://github.com/openai/CLIP}{https://github.com/openai/CLIP}} models.

\begin{table*}[!htbp]
\newcolumntype{Y}{>{\raggedright\arraybackslash}X}
\newcolumntype{Z}{>{\centering\arraybackslash}X}
\centering
\footnotesize
\setlength\tabcolsep{1pt}
\renewcommand{\arraystretch}{1.2}

\begin{tabularx}{\textwidth}{cl c YYY c YYY c YYY c YYY c YYY}
\toprule

\multicolumn{22}{@{\hskip 0.11in}c}{\bf GZSL Methods using CLIP Visual Features} \\
\midrule

{\multirow{2}{*}{\bf \shortstack{Data \\set }}} &
{\multirow{2}{*}{\bf \shortstack{Back \\bone }}} &~&
\multicolumn{3}{@{\hskip 0.11in}c}{\bf tfVAEGAN} &~~& 
\multicolumn{3}{@{\hskip 0.11in}c}{\bf CADA-VAE} &~~& 
\multicolumn{3}{@{\hskip 0.11in}c}{\bf SDGZSL} &~~& 
\multicolumn{3}{@{\hskip 0.11in}c}{\bf FREE} &~~& 
\multicolumn{3}{@{\hskip 0.11in}c}{\bf CE} \\

\cmidrule{4-6}\cmidrule{8-10}\cmidrule{12-14}\cmidrule{16-18}\cmidrule{20-22}

&&& \textit{Seen} & \textit{Novel} & \textit{Harm.} 
&& \textit{Seen} & \textit{Novel} & \textit{Harm.} 
&& \textit{Seen} & \textit{Novel} & \textit{Harm.} 
&& \textit{Seen} & \textit{Novel} & \textit{Harm.} 
&& \textit{Seen} & \textit{Novel} & \textit{Harm.} \\

\midrule

\parbox[t]{1mm}{\multirow{11}{*}{\rotatebox[origin=c]{90}{CUB}}}
&{R50} &  & 
66.91 & 49.96 & 57.20  & ~ &
61.67 & 56.67 & 59.06  & ~ &
61.24 & 56.89 & 58.98  & ~ &
62.95 & 47.70 & 54.28  & ~ &
40.72 & 37.32 & 38.95  \\

&{R101} &  & 
65.83 & 60.13 & 62.85  & ~ &
62.31 & 61.97 & 62.14  & ~ &
65.03 & 59.53 & 62.16  & ~ &
59.41 & 60.05 & 59.73  & ~ &
60.00 & 52.52 & 56.01  \\

&{R50$_{\text{x4}}$} &  & 
70.48 & 65.41 & 67.85  & ~ &
67.90 & 68.84 & 68.37  & ~ &
66.38 & 67.77 & 67.07  & ~ &
68.55 & 59.75 & 63.85  & ~ &
62.57 & 44.50 & 52.01 \\

&{R50$_{\text{x16}}$} &  & 
73.94 & 69.22 & 71.50  & ~ &
76.40 & 67.15 & 71.47  & ~ &
72.65 & 68.92 & 70.74  & ~ &
71.02 & 63.27 & 66.92  & ~ &
61.04 & 46.26 & 52.63  \\

&{R50$_{\text{x64}}$} &  & 
82.09 & 67.93 & 74.34  & ~ &
75.91 & 70.84 & 73.29  & ~ & 
74.77 & 73.70 & 74.23  & ~ &
68.53 & \underline{73.47} & 70.91  & ~ &
59.06 & 49.52 & 53.87  \\

&{ViT$_{\text{B32}}$} &  & 
64.37 & 62.08 & 63.20  & ~ &
64.34 & 57.52 & 60.74  & ~ &
64.20 & 59.66 & 61.84  & ~ &
61.51 & 54.87 & 58.00  & ~ &
60.96 & 50.55 & 55.27  \\

&{ViT$_{\text{B16}}$} &  & 
76.48 & 66.32 & 71.04  & ~ &
71.13 & 67.79 & 69.42  & ~ &
64.81 & \underline{74.25} & 69.21  & ~ &
68.46 & 65.99 & 67.20  & ~ &
58.30 & \underline{62.17} & \underline{60.17}  \\

&{ViT$_{\text{L14}}$} &  & 
\underline{77.67} & \underline{72.36} & \underline{74.92}  & ~ &
\underline{77.90} & \underline{72.98} & \underline{75.36}  & ~ &
\underline{79.50} & 73.49 & \underline{76.38}  & ~ &
\underline{79.05} & 65.38 & \underline{71.56}  & ~ &
\underline{74.04} & 46.38 & 57.03  \\

\arrayrulecolor{gray!48} \cmidrule{2-22} 
\arrayrulecolor{black}

&{ViT$_{\text{L14}}^\dag$} &  & 
80.39 & 72.86 & 76.44  & ~ &
\textbf{81.96} & 71.24 & 76.22  & ~ & 
79.19 & 74.70 & 76.88  & ~ &
78.87 & 65.89 & 71.80  & ~ &
71.49 & 48.62 & 57.88  \\

&{ViT$_{\text{L14}}^\ddag$} &  & 
80.73 & \textbf{73.59} & 76.99  & ~ &
79.51 & 74.68 & 77.01  & ~ &
\textbf{80.40} & 74.33 & 77.25  & ~ &
79.47 & \textbf{68.25} & 73.44  & ~ &
\textbf{75.12} & 51.14 & 60.86 \\

&\cellcolor{gray!18} {ViT$_{\text{L14}}^\S$} & \cellcolor{gray!18}  & 
\cellcolor{gray!18} \textbf{82.82} & \cellcolor{gray!18} 72.27 & \cellcolor{gray!18} \textbf{77.18}  & \cellcolor{gray!18} ~ &
\cellcolor{gray!18} 81.47 & \cellcolor{gray!18} \textbf{75.05} & \cellcolor{gray!18} \textbf{78.13}  & \cellcolor{gray!18} ~ &
\cellcolor{gray!18} 80.38 & \cellcolor{gray!18} \textbf{75.67} & \cellcolor{gray!18} \textbf{77.96}  & \cellcolor{gray!18} ~ &
\cellcolor{gray!18} \textbf{80.48} & \cellcolor{gray!18} 67.54 & \cellcolor{gray!18} \textbf{73.45}  & \cellcolor{gray!18} ~ &
\cellcolor{gray!18} 71.15 & \cellcolor{gray!18} \textbf{56.82} & \cellcolor{gray!18} \textbf{63.18} \\


\midrule

\parbox[t]{1mm}{\multirow{11}{*}{\rotatebox[origin=c]{90}{SUN}}}
&{R50} &  & 
51.94 & 53.40 & 52.66  & ~ &
47.87 & 61.88 & 53.98  & ~ &
43.29 & 54.10 & 48.10  & ~ &
53.10 & 54.24 & 53.66  & ~ &
38.84 & 52.85 & 44.77  \\

&{R101} &  & 
54.53 & 62.99 & 58.46  & ~ &
53.02 & 61.53 & 56.96  & ~ &
53.88 & 63.96 & 58.49  & ~ &
53.18 & 58.19 & 55.57  & ~ &
51.24 & 55.83 & 53.44  \\

&{R50$_{\text{x4}}$} &  & 
54.38 & 65.00 & 59.22  & ~ &
55.16 & 60.56 & 57.73  & ~ &
55.00 & 65.62 & 59.84  & ~ &
53.37 & 62.57 & 57.61  & ~ &
54.92 & 55.07 & 55.00  \\

&{R50$_{\text{x16}}$} &  & 
57.87 & 68.33 & 62.67  & ~ &
57.64 & 62.92 & 60.16  & ~ &
46.71 & 61.18 & 52.97  & ~ &
54.22 & 62.22 & 57.95  & ~ &
54.26 & \underline{60.21} & 57.08  \\

&{R50$_{\text{x64}}$} &  & 
57.21 & \underline{69.79} & 62.88  & ~ &
59.46 & 65.63 & 62.39  & ~ &
57.17 & 66.60 & 61.52  & ~ &
\underline{60.04} & 57.78 & 58.89  & ~ &
50.04 & 59.10 & 54.19  \\

&{ViT$_{\text{B32}}$} &  & 
58.14 & 63.40 & 60.66  & ~ &
54.50 & 61.94 & 57.98  & ~ &
56.43 & 61.67 & 58.93  & ~ &
51.67 & 61.11 & 55.99  & ~ &
53.95 & 52.92 & 53.43  \\

&{ViT$_{\text{B16}}$} &  & 
57.98 & 67.01 & 62.17  & ~ &
54.69 & 62.08 & 58.15  & ~ &
57.40 & \underline{66.94} & 61.81  & ~ &
54.38 & 59.93 & 57.02  & ~ &
55.12 & 56.94 & 56.02  \\

&{ViT$_{\text{L14}}$} &  & 
\underline{59.84} & 68.89 & \underline{64.05}  & ~ &
\underline{62.11} & \underline{63.40} & \underline{63.18}  & ~ &
\underline{63.57} & 62.71 & \underline{63.13}  & ~ &
58.18 & \underline{62.71} & \underline{60.36}  & ~ &
\underline{60.16} & 57.85 & \underline{58.98}  \\

\arrayrulecolor{gray!48} \cmidrule{2-22} 
\arrayrulecolor{black}

&{ViT$_{\text{L14}}^\dag$} &  & 
61.75 & \textbf{70.49} & 65.83  &\cellcolor{gray!18}  ~ &
\cellcolor{gray!18} \textbf{61.47} &\cellcolor{gray!18}  65.76 &\cellcolor{gray!18}  \textbf{63.54}  & ~ &
\textbf{62.56} & 64.44 & 63.49  & ~ &
\textbf{58.95} & 62.99 & 60.90  & ~ &
52.79 & \textbf{64.37} & 58.01  \\

&{ViT$_{\text{L14}}^\ddag$} &  & 
60.93 & 67.08 & 63.86  & ~ &
57.95 & \textbf{68.19} & 62.65  & ~ &
60.08 & 67.15 & 63.42  & ~ &
58.37 & 62.43 & 60.33  & ~ &
56.78 & 64.03 & 60.19  \\

&{ViT$_{\text{L14}}^\S$} &\cellcolor{gray!18}  & 
\cellcolor{gray!18} \textbf{62.25} &\cellcolor{gray!18}  70.03 &\cellcolor{gray!18}  \textbf{65.91}  & ~ &
58.91 & 66.74 & 62.58  &\cellcolor{gray!18}  ~ &
\cellcolor{gray!18} 60.70 &\cellcolor{gray!18}  \textbf{67.50} &\cellcolor{gray!18}  \textbf{63.92}  &\cellcolor{gray!18}  ~ &
\cellcolor{gray!18} 58.72 &\cellcolor{gray!18}  \textbf{63.61} &\cellcolor{gray!18}  \textbf{61.07}  &\cellcolor{gray!18}  ~ &
\cellcolor{gray!18} \textbf{59.65} &\cellcolor{gray!18}  60.90 &\cellcolor{gray!18}  \textbf{60.27}  \\


\midrule

\parbox[t]{1mm}{\multirow{11}{*}{\rotatebox[origin=c]{90}{AWA2}}}
&{R50} &  & 
72.02 & 60.31 & 65.65  & ~ &
80.99 & 61.30 & 71.14  & ~ &
77.72 & 57.66 & 66.21  & ~ &
80.03 & 55.17 & 65.31  & ~ &
78.84 & 45.19 & 57.45  \\

&{R101} &  & 
71.94 & 64.79 & 68.18  & ~ &
85.25 & 59.34 & 70.87  & ~ &
80.29 & 68.78 & 74.09  & ~ &
78.64 & 59.11 & 67.49  & ~ &
76.61 & 62.34 & 68.74  \\

&{R50$_{\text{x4}}$} &  & 
83.93 & \underline{68.02} & 75.15  & ~ &
80.71 & \underline{69.87} & 77.02  & ~ &
82.23 & 68.39 & 74.67  & ~ &
83.30 & 57.29 & 67.89  & ~ &
81.46 & 61.03 & 69.78  \\

&{R50$_{\text{x16}}$} &  & 
86.48 & 66.49 & 75.18  & ~ &
89.35 & 65.88 & 77.30  & ~ &
84.24 & \underline{73.18} & \underline{78.32}  & ~ &
83.09 & 60.13 & 69.77  & ~ &
83.13 & 66.83 & 74.09  \\

&{R50$_{\text{x64}}$} &  & 
89.69 & 68.21 & 77.49  & ~ &
90.54 & 63.75 & 75.81  & ~ &
78.67 & 69.07 & 73.56  & ~ &
\underline{87.60} & 61.04 & 71.95  & ~ &
89.95 & 63.33 & 74.33  \\

&{ViT$_{\text{B32}}$} &  & 
84.24 & 62.41 & 71.70  & ~ &
80.53 & 65.35 & 73.90  & ~ &
84.33 & 62.43 & 71.75  & ~ &
74.01 & 57.13 & 64.48  & ~ &
78.91 & 62.86 & 69.98  \\

&{ViT$_{\text{B16}}$} &  & 
87.80 & 66.56 & 75.72  & ~ &
90.56 & 65.44 & 75.80  & ~ &
86.87 & 66.59 & 75.39  & ~ &
77.05 & 61.58 & 68.45  & ~ &
84.02 & 66.92 & 74.50  \\

&{ViT$_{\text{L14}}$} &  & 
\underline{90.95} & 69.50 & \underline{78.70}  & ~ &
\underline{92.68} & 68.14 & \underline{77.99}  & ~ &
\underline{89.99} & 68.76 & 77.91  & ~ &
81.10 & \underline{65.99} & \underline{72.77}  & ~ &
\underline{85.63} & \underline{71.96} & \underline{78.20}  \\

\arrayrulecolor{gray!48} \cmidrule{2-22} 
\arrayrulecolor{black}

&{ViT$_{\text{L14}}^\dag$} &  & 
91.25 & 67.77 & 77.77  & ~ &
90.48 & 70.07 & 78.98  & ~ &
89.18 & 71.40 & 79.30  &\cellcolor{gray!18} ~ &
\cellcolor{gray!18}81.27 &\cellcolor{gray!18} \textbf{71.56} &\cellcolor{gray!18} \textbf{75.94}  &\cellcolor{gray!18} ~ &
\cellcolor{gray!18}85.12 &\cellcolor{gray!18} \textbf{78.02} &\cellcolor{gray!18} \textbf{81.41}  \\

&{ViT$_{\text{L14}}^\ddag$} &\cellcolor{gray!18}  & 
\cellcolor{gray!18}\textbf{93.40} &\cellcolor{gray!18} 72.61 &\cellcolor{gray!18} \textbf{81.70}  &\cellcolor{gray!18} ~ &
\cellcolor{gray!18}\textbf{93.51} &\cellcolor{gray!18} \textbf{73.85} &\cellcolor{gray!18} \textbf{82.53}  &\cellcolor{gray!18} ~ &
\cellcolor{gray!18}85.77 &\cellcolor{gray!18} \textbf{74.16} &\cellcolor{gray!18} \textbf{79.54}  & ~ &
81.29 & 65.62 & 72.62  & ~ &
89.85 & 64.12 & 74.84  \\

&{ViT$_{\text{L14}}^\S$} &  & 
89.62 & \textbf{73.56} & 80.80  & ~ &
93.35 & 68.14 & 78.78  &\cellcolor{gray!18} ~ &
\cellcolor{gray!18}\textbf{91.50} &\cellcolor{gray!18} 70.33 &\cellcolor{gray!18} \textbf{79.54}  & ~ &
\textbf{81.53} & 69.84 & 75.23  & ~ &
\textbf{90.78} & 67.96 & 77.73  \\

\bottomrule
\end{tabularx}
\caption{Results of Generative and Disentanglement Based Methods for the CUB, SUN and AWA2 datasets using different features extracted from different size and architecture of the visual head from diverse CLIP models (i.e., Resnet50 (R50) and Vision Transformer (ViT)). The three bottom rows per section correspond to fine-tuned features using sentences with: $\dag$ the class names, $\ddag$ the attributes, and $\S$ both class names and attributes. The bold numbers correspond to the highest scores per column, the underline numbers correspond to the to the highest scores using features not fine-tuned, and the shaded rows correspond to the most performant image feature per method over all. Surprisingly, the most performant method for all datasets correspond to a \textit{generative} based method.
}
\label{tab:gzsl_using_clip_features_full}
\end{table*}



\section{Ethical Considerations}
\label{ethical}

Machine learning models still require collecting large amounts of annotated data. In the case of fine-grained recognition, these annotations often require specialized human knowledge. Zero-shot learning offers a way for bypassing the need to collect extensive amounts of data for training models for new classes of objects. 
We show that large-scale pre-trained models along with Generalized Zero-Shot Learning methods can obtain results that are competitive with the specialized knowledge from experts on classes that a trained model has never seen. We hope that the key insights and analysis provided in this paper will be useful in expanding and leveraging zero-shot research along with current progress in multi-modal learning. Allowing for the creation of models that do not depend on large amounts of data could be useful for practitioners without access to large scale resources or in domains where data is scarce such as the medical domain.

\end{document}